\documentclass[11pt,letterpaper]{article}
\usepackage[T1]{fontenc}
\usepackage[utf8]{inputenc}
\usepackage{lmodern,microtype}
\usepackage[margin=1in]{geometry}
\usepackage{booktabs,amsmath,amssymb,amsthm,mathtools}
\usepackage{graphicx,algorithm,algpseudocode}
\usepackage[font=small,labelfont=bf,skip=8pt]{caption}
\usepackage[font=small,skip=5pt]{subcaption}
\usepackage[table]{xcolor}
\usepackage{array,longtable,tabularx,multirow}
\usepackage{enumitem,titlesec,mdframed}
\usepackage{placeins}
\usepackage{tikz}
\usetikzlibrary{arrows.meta,positioning,shapes.geometric,calc}
\usepackage[numbers,sort&compress]{natbib}
\usepackage{xurl}
\definecolor{inkblue}{HTML}{17466B}
\definecolor{auditteal}{HTML}{087D85}
\definecolor{softteal}{HTML}{EFF8F8}
\definecolor{softblue}{HTML}{F0F4F8}
\usepackage[colorlinks=true,linkcolor=inkblue,citecolor=inkblue,urlcolor=auditteal]{hyperref}
\usepackage{bookmark}
\hypersetup{pdftitle={When the Winner's Curse Protects: Exact Update Audits After LLM Model Selection},pdfauthor={Qianyi Chen and Bo Li}}
\newtheorem{theorem}{Theorem}[section]
\newtheorem{proposition}[theorem]{Proposition}

\theoremstyle{definition}
\newtheorem{definition}[theorem]{Definition}

\theoremstyle{remark}
\newtheorem{remark}[theorem]{Remark}
\newcommand{\E}{\mathbb{E}}

\newcommand{\ind}{\mathbf{1}}
\newcommand{\Hybrid}{\textsc{HybridCertified}}
\newcommand{\FreshU}{\textsc{FreshUnconditional}}
\newcommand{\FreshE}{\textsc{FreshExact}}
\newcommand{\pp}{\,\mathrm{pp}}
\titlespacing*{\section}{0pt}{10pt plus 2pt minus 2pt}{5pt plus 1pt}
\titlespacing*{\subsection}{0pt}{7pt plus 2pt minus 2pt}{3pt plus 1pt}
\titlespacing*{\paragraph}{0pt}{3pt plus 1pt}{0.6em}
\titleformat{\section}{\large\bfseries}{\thesection}{0.65em}{}
\titleformat{\subsection}{\normalsize\bfseries}{\thesubsection}{0.65em}{}
\setlist{topsep=3pt,itemsep=1pt,parsep=0pt,leftmargin=1.5em}

\AtBeginDocument{%
  \setlength{\abovedisplayskip}{8pt plus 2pt minus 2pt}%
  \setlength{\belowdisplayskip}{8pt plus 2pt minus 2pt}%
  \setlength{\abovedisplayshortskip}{3pt plus 1pt}%
  \setlength{\belowdisplayshortskip}{6pt plus 1pt minus 1pt}%
}
\title{\LARGE\bfseries When the Winner's Curse Protects:\\[2pt] Exact Update Audits After LLM Model Selection}
\author{Qianyi Chen \qquad Bo Li\thanks{Corresponding author.}\\[4pt]
\small School of Economics and Management, Tsinghua University\\[2pt]
\small\texttt{\{cqy22@mails,libo@sem\}.tsinghua.edu.cn}}
\date{\vspace{-10pt}\small
\href{https://github.com/Cqyiiii/when-winner-curse-protects}{\raisebox{-0.12em}{\includegraphics[height=1em]{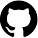}}\,GitHub}
\qquad
\href{https://cqyiiii.github.io/winner-curse/}{\raisebox{-0.12em}{\includegraphics[height=1em]{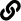}}\,Project Page}\vspace{-2\baselineskip}}
\renewenvironment{abstract}{%
  \par\small\setlength{\parindent}{0pt}%
  \setlength{\parskip}{0.45\baselineskip}%
  {\centering\normalsize\bfseries Abstract\par}\vspace{2pt}%
}{\par\vspace{3pt}}
\begin{document}
\maketitle
\begin{abstract}
Modern large language model (LLM) production relies on continual checkpointing and model selection across fine-tuning runs, prompts, and adaptation recipes. Because decoding and verification are costly, selection typically rests on a single pass over a finite validation panel. Under the classical \emph{winner's curse}, picking the top scorer without adjustment incurs upward bias from favorable random error realizations. Standard practice therefore discards these cached scores when auditing a follow-up update, forfeiting up to half the paired sample under a fixed evaluation budget. We invert this bias into a provable safeguard for upgrade certification: because parent optimism enters the child-minus-parent gain negatively, panel reuse suppresses false upgrades while supplying up to twice as many pairs. Building on this asymmetry, \textsc{HybridCertified} pairs an exact one-sided McNemar gate that guarantees conditional finite-sample false-upgrade control under arbitrary within-item candidate correlation with deterministic completion certificates for zero-distortion early stopping. Across 224 LLM cohorts spanning Qwen3-8B and Mistral-7B, \textsc{HybridCertified} reliably blocks forgetting and overfitting regressions while preserving beneficial updates, eliminates normal-test Type-I inflation, improves accuracy over a calibrated fresh audit on power-limited updates, and saves up to twice as many virtual queries, skipping over 70\% of the budget on degrading updates.
\end{abstract}
\begin{figure}[t]
\centering
\begin{tikzpicture}[font=\footnotesize,>=Latex,
 box/.style={draw=inkblue!65,rounded corners=2pt,fill=softblue,align=center,inner sep=5pt},
 route/.style={draw=inkblue!75,line width=.65pt,-{Latex[length=2mm]}},
 label text/.style={font=\scriptsize,fill=white,inner sep=2pt,align=center}]
\node[box,text width=3.0cm,minimum height=1.05cm] (parents) at (-5.7,0) {$K$ candidate parents\\$P_1,\ldots,P_K$};
\node[box,text width=3.7cm,minimum height=1.05cm] (winner) at (0,0) {Selected parent $J$\\optimistic $\widehat p_J(S)$\\parent outcomes are sunk};
\node[box,text width=3.0cm,minimum height=1.05cm] (child) at (5.7,0) {Candidate child $C_J$\\preassigned update};
\draw[route] (parents) -- node[label text,above]{select on $S$} (winner);
\draw[route] (winner) -- node[label text,above]{train on\\disjoint $D$} (child);
\node[box,text width=6.2cm,minimum height=1.55cm] (fresh) at (-3.9,-2.35) {\textbf{Fresh audit}\\Score both models on new items\\2 calls per pair: $n_F=B/2$};
\node[box,draw=auditteal,fill=softteal,text width=6.2cm,minimum height=1.55cm] (hybrid) at (3.9,-2.35) {\textbf{HybridCertified}\\Reuse $S$: 1 child call per item\\Add fresh pairs: $n_H=(B+n_S)/2$\\Conservative under $H_0:\Delta_J\leq0$};
\draw[draw=inkblue!75,line width=.65pt] (child.south) -- ++(0,-.48) -| (0,-1.18);
\draw[route] (0,-1.18) -| (fresh.north);
\draw[route] (0,-1.18) -| (hybrid.north);
\node[label text] at (0,-1.18) {fixed incremental budget $B$};
\node[box,fill=white,text width=12.4cm,minimum height=.8cm] (gate) at (0,-4.0) {\textbf{Exact McNemar gate: deploy the child or retain the parent}\\Completion certificates: early stopping with zero decision distortion.};
\draw[route] (fresh.south) -- (fresh.south |- gate.north);
\draw[route] (hybrid.south) -- (hybrid.south |- gate.north);
\end{tikzpicture}
\caption{Post-selection auditing. Reused parent outcomes buy more pairs. Selected-winner validity controls false upgrades; completion certificates preserve every endpoint action.}\label{fig:overview}
\end{figure}
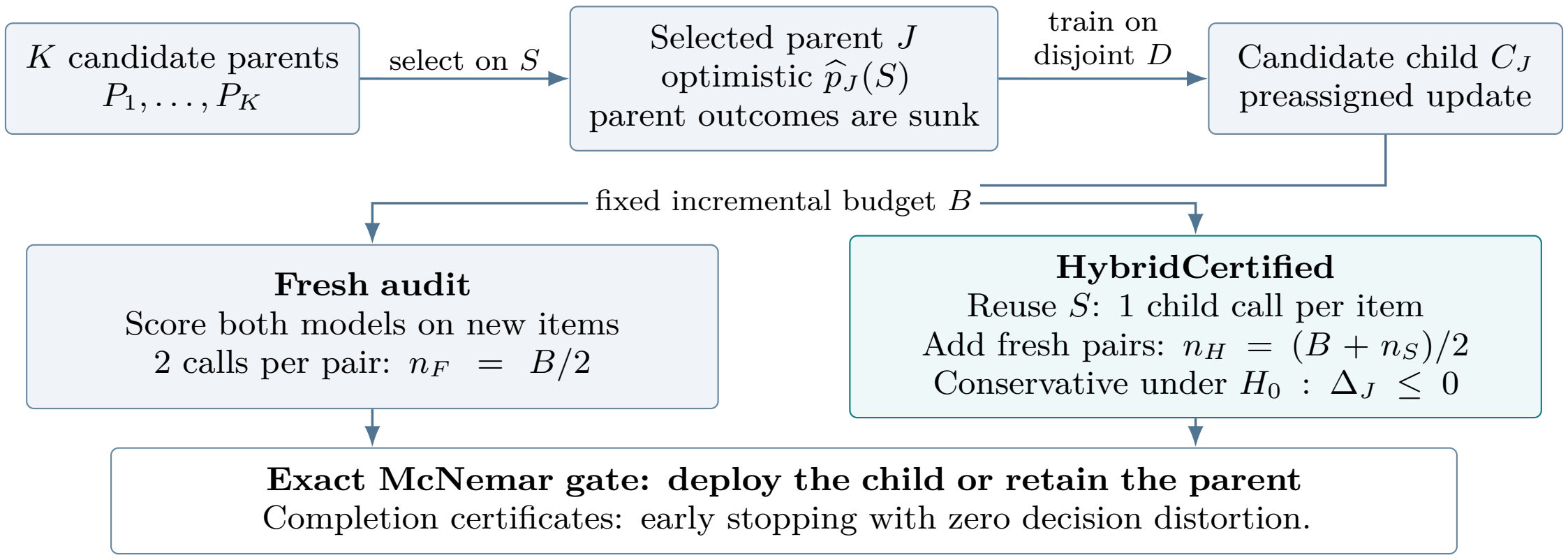
\section{Introduction}
Modern large language model (LLM) development unfolds as a continual cycle of model selection and iterative refinement. Post-training pipelines using supervised fine-tuning, low-rank adaptation, reinforcement learning from human feedback, and direct preference optimization routinely produce competing checkpoints, prompt variants, and training recipes \citep{ouyang2022,rafailov2023,hu2022lora}. Choosing which incumbent $P_J$ to retain typically relies on a single pass over a finite validation panel $S$, because scoring each candidate demands expensive autoregressive multi-token generation, strict exact-match or format verification, or judge evaluations \citep{polo2024,perlitz2024}. Once a follow-up candidate update $C_J$ is trained, deployment teams face a recurring statistical decision: does $C_J$ genuinely improve upon the selected incumbent $P_J$ under a limited incremental evaluation budget?

Reusing the validation panel $S$ for this audit raises an immediate statistical concern. Selecting the highest-scoring candidate induces the classical \emph{winner's curse}: the incumbent's empirical score $\widehat p_J(S)$ is biased upward by favorable random error realizations, invalidating standard inference that treats $P_J$ as fixed in advance \citep{fithian2014,dwork2015holdout,zrnic2025}. The usual remedy is to discard $S$ and evaluate both $P_J$ and $C_J$ on an independent fresh panel. Although statistically safe, this approach is operationally expensive. Every fresh paired comparison must decode and score two models, whereas reusing a cached item in $S$ requires scoring only the child $C_J$. Discarding $S$ therefore forfeits up to half of the paired observations obtainable from a fixed incremental budget $B$. For example, with $n_S=256$ cached parent outcomes, $B=256$ new evaluations buy 256 pairs under reuse versus 128 under a fresh audit; at $B=334$, pooling reused and fresh items supplies 295 pairs rather than 167.

The directional nature of an upgrade audit, however, changes the role of the winner's curse. Rather than estimating the incumbent's absolute accuracy, upgrade certification tests whether the child's population gain $\Delta_J$ is positive against the null $H_{0,J}:\Delta_J\leq0$. In the reused gain estimate $\widehat\Delta(S)=\widehat p_{C_J}(S)-\widehat p_J(S)$, the incumbent's optimistic selection error enters with a minus sign. Whenever the child inherits less optimism than the selected parent, selection introduces \emph{conservative drag} that makes an upgrade harder to certify. As illustrated in Figure~\ref{fig:overview}, the winner's curse can thus act as a built-in safeguard against false upgrades while panel reuse expands the paired sample by up to twofold. Whether this extra sample size outweighs conservative drag determines where reuse yields higher deployment utility. Turning this directional intuition into a rigorous finite-sample guarantee requires resolving three obstacles rooted directly in LLM post-training.

First, downward average bias neither controls false-upgrade tail probabilities nor addresses cross-candidate dependence. Candidate checkpoints adapted from a shared backbone such as Qwen3 or Mistral across seeds, data orders, or hyperparameters share pretraining representations and often succeed or fail together on the same prompts. A useful guarantee must therefore accommodate arbitrary, unknown within-item correlation across candidate models without assuming independent candidates or relying on Gaussian approximations. Second, follow-up LLM updates face sharp asymmetric failure modes, including catastrophic forgetting under domain or support shift and small-batch overfitting, that can severely degrade deployment accuracy. An audit must block such regressions while avoiding wasted decoding once a stream becomes decisive. Here the cached labels offer additional structure: a known parent success can never become a repair under the child. Third, evaluating reuse fairly requires a competitive fresh comparator. Because binary paired tests are discrete, an uncalibrated conditional McNemar fresh test can be unnecessarily conservative, risking crediting avoidable baseline conservatism to selection reuse \citep{mcnemar1947,suissa1991}.

We address these challenges with \Hybrid, an exact one-sided full-tail McNemar audit over reused and fresh pairs coupled with deterministic \emph{completion certificates}. Under iid binary evaluation rows, maximum-score parent selection, and potential updates preassigned independently of $S$, Theorem~\ref{thm:selected-winner} proves false-upgrade control at the nominal level conditional on the selected winner, even under arbitrary within-item dependence among candidates. Preassignment concerns the update recipes and randomness; only the winning parent's child needs to be trained. The proof conditions on rowwise parent--child score sums and invokes product-measure association \citep{fkg1971} to show that winner selection can only suppress the upper rejection tail. Attached to this fixed endpoint, completion certificates track known parent labels on unread selection items alongside atomic two-call fresh pairs. Within the fixed query order, they halt at the earliest prefix where every admissible completion produces the same accept or reject decision, guaranteeing zero decision distortion. For fair benchmarking, we also construct \FreshU, which uses exact rational Bernstein subdivision to certify an enlarged fresh cutoff uniformly across all discordance rates.

We evaluate the framework through complementary evidence linking mechanism, deployment utility, and evaluation cost. Log-likelihood overlap studies isolate selection--gain covariance, and 112 historical morphological-inflection cohorts characterize power-limited beneficial updates. Two prospective, outcome-blinded mixed-regime benchmarks cover 64 dual-family inflection cohorts from Qwen3-8B and Mistral-7B and 48 Banking77 intent-classification cohorts. Across these prospective mixtures of beneficial, forgetting, and overfitting updates, \Hybrid\ outperforms always updating by 1.92 and 16.80 percentage points and never updating by 4.51 and 5.37 points. The inflection contrast against always updating passes its prespecified one-sided Holm adjustment, but not its two-sided counterpart. Hybrid captures 71.6\% and 97.1\% of the hindsight-oracle headroom over unconditional updating. Neither prospective study establishes accuracy superiority over \FreshU\ at equal budget; strong beneficial gains leave little room to improve fresh acceptance. The historical benchmark and an exact binomial-sum surrogate instead delineate where the sample-size dividend overcomes conservative drag. Finally, exact independent-pair null bounds expose finite-sample Type-I inflation in normal-approximation gates, while completion certificates save $1.77\times$ to $2.00\times$ as many virtual queries as \FreshU, skipping up to 71.4\% of the full budget on degrading updates.

In summary, our contributions are four-fold:
\begin{itemize}
\item \textbf{Exact post-selection validity with correlated candidates.} Theorem~\ref{thm:selected-winner} proves finite-sample false-upgrade control conditional on the winner using rowwise sum conditioning and product-measure association, without candidate independence or correlation estimation.
\item \textbf{Zero-distortion completion certificates and calibrated fresh benchmarking.} We derive sharp completion bounds for one-call reused and two-call fresh acquisition (Proposition~\ref{prop:completion}, Algorithm~\ref{alg:completion}), alongside rational Bernstein calibration for same-budget fresh audits (Proposition~\ref{prop:bernstein}).
\item \textbf{Prospective dual-family and dual-domain LLM evaluation.} Through outcome-blinded mixed-regime benchmarks across Qwen3-8B and Mistral-7B on morphological inflection and intent classification, we quantify deployment gains, regression blocking, and cohort-level drivers of reuse versus fresh disagreement.
\item \textbf{Unified analysis of power crossover, null boundaries, and audit efficiency.} Combining an exact iid-surrogate phase boundary, rational null enclosures, twelve million correlated-candidate simulations, and full-stream certificate replays, we separate statistical error control from power trade-offs and virtual query savings.
\end{itemize}

The paper formalizes selection drag (Section~\ref{sec:setup}), develops exact certification, completion certificates, and the power boundary (Section~\ref{sec:method}), presents the empirical evidence (Section~\ref{sec:experiments}), and discusses practical guidance and limitations (Section~\ref{sec:conclusion}).
\subsection{Related Work}\label{sec:related}

Selective inference addresses uncertainty after selection \citep{fithian2014,taylor2015}, including exact conditional lasso inference \citep{lee2016}. Related work controls false coverage in selected intervals \citep{benjamini2005}, examines Bayesian selection paradoxes \citep{dawid1994}, and tunes winner correction to selection strength \citep{zrnic2025}. Reusable holdouts control information release \citep{dwork2015holdout}, while information-usage bounds quantify dependence-induced bias \citep{russo2020}. Under iid binary rows and preassigned updates, our product-measure argument \citep{fkg1971} controls false upgrades conditional on the maximum-score winner, with arbitrary within-row dependence.

Recent work uses exact one-sided McNemar testing to detect LLM degradation \citep{kuebler2026}, while our guarantee addresses selection-induced dependence when the incumbent's selection panel is reused. The exact McNemar test conditions on discordances \citep{mcnemar1947,agresti2013}; unconditional procedures can reduce conservatism by retaining all pairs \citep{suissa1991}, while maximizing $p$-values over nuisance-parameter confidence sets provides another calibration \citep{berger1994}. Our Fresh comparator uses rational Bernstein bounds over the entire discordance-rate interval before increasing McNemar's full-tail cutoff, with calibration specific to iid fresh pairs. The sampling unit separates comparisons of fitted classifiers from comparisons incorporating training variation \citep{dietterich1998,dror2018}; Holm adjustment \citep{holm1979}, Hoeffding bounds \citep{hoeffding1963}, binomial intervals \citep{clopper1934}, and Welch--Satterthwaite sensitivity \citep{welch1947,satterthwaite1946} serve distinct inferential purposes.

Efficient benchmarking predicts scores from curated items \citep{polo2024}, selects informative subsets using model correlations \citep{vivek2024}, or cuts the cost of preserving conclusions \citep{perlitz2024}; evaluation design emphasizes pairing and uncertainty \citep{miller2024}. Using such item selection in our audit needs a separate argument preserving iid sampling. Sequential tests \citep{wald1945}, confidence sequences and betting \citep{howard2021,waudby2024}, and A/B monitoring \citep{johari2022} support anytime-valid inference \citep{ramdas2023}. Completion follows deterministic curtailment \citep{eisenberg1980,chi2008}: cached parent outcomes limit feasible continuations, so queries stop only when all continuations give the same fixed-endpoint action, preserving its decision, power, and error control.

Reliable improvement over an incumbent motivates high-confidence policy improvement \citep{thomas2015}, baseline bootstrapping \citep{laroche2019}, and multi-objective extensions \citep{satija2021}; these infer policy values from logs, where unobserved actions and state visitation complicate inference, whereas paired audits score both models directly. Instruction tuning with human feedback \citep{ouyang2022}, direct preference optimization \citep{rafailov2023}, and parameter-efficient adaptation \citep{hu2022lora} generate updates whose deployment merits require separate evaluation. Our experiments isolate this audit through controlled LoRA updates on Qwen3 and Mistral \citep{qwen32025,mistral2023}. SIGMORPHON--UniMorph provides morphological inflection tasks \citep{goldman2023}, Banking77 provides fine-grained intent labels \citep{casanueva2020}, and AG News supports the classification mechanism study \citep{zhang2015}, with distinct error and discordance structures.
\section{Problem Formulation and Asymmetric Selection Drag}\label{sec:setup}
Selection ranks candidate parents by accuracy, whereas an update audit targets the child's gain over the selected parent. We distinguish these objectives through their data roles and incremental query costs before examining how selection affects the reused gain estimate.

\subsection{Selected incumbents and a fixed incremental budget}
For candidate checkpoint $P_k$, let $C_k$ be its refinement on disjoint training data $D$. The selection panel $S$ chooses the incumbent, $F$ supplies independent audit pairs, and $T$ evaluates deployment. Conditional on training information $\mathcal M$, all $K$ potential pairs $(P_k,C_k)$ have update recipes, data, and seeds fixed before observing $S$; only the winner's child needs training. Under a common scoring rule, $P_k(z),C_k(z)\in\{0,1\}$ indicate correctness. Complete potential-outcome rows are iid across items, with arbitrary within-row dependence. Write $p_k=\E[P_k(z)\mid\mathcal M]$ and $\Delta_k=\E[C_k(z)-P_k(z)\mid\mathcal M]$. The selected incumbent is
\begin{equation}
 J=\min\mathop{\arg\max}_{1\le k\le K}\widehat p_k(S),
 \qquad \widehat p_k(S)=\frac1{n_S}\sum_{z\in S}P_k(z).
 \label{eq:selection}
\end{equation}
The upgrade null is $H_{0,J}:\Delta_J\le0$. Training data $D$, fresh audit stream $F$, and deployment-test stream $T$ are independent of $S$, conditional on $\mathcal M$. Fresh audit and deployment-test rows follow the same distribution as selection items.

Each endpoint call scores one model on one item, so the incremental budget $B$ pays once per reused item and twice per fresh pair after excluding sunk parent-selection calls. For $B\ge n_S$ and integer pair counts, the resulting audit sizes are
\begin{equation}
 n_H=n_S+\frac{B-n_S}{2}=\frac{B+n_S}{2},\qquad n_F=\frac B2.
 \label{eq:budget}
\end{equation}
\subsection{The direction of the selection effect}
The gain decomposition makes the direction of selection dependence explicit. For $\widehat\Delta(A)=\widehat p_{C_J}(A)-\widehat p_J(A)$, parent optimism $O_P$ and child error $O_C$ enter with opposite signs:
\begin{equation}
\begin{gathered}
  O_P = \hat{p}_J(S) - \hat{p}_J(T), \qquad O_C = \hat{p}_{C_J}(S) - \hat{p}_{C_J}(T), \\[3pt]
  \hat{\Delta}(S) - \hat{\Delta}(T) = O_C - O_P.
\end{gathered}
\label{eq:drag-decomp}
\end{equation}
Reuse depresses the gain when parent optimism exceeds inherited child optimism. To isolate this drag, the mechanism studies fix both selection and audit panels at 32 items, turning off reuse's potential $2\times$ sample-size dividend. They independently vary selection ($K=4$ versus an unselected $K=1$ control) and audit overlap $\alpha\in\{0,0.25,0.50,0.75,1\}$. Let $Q_{\alpha,K}$ denote the audited update's contribution to deployment log utility; the difference-in-differences $D(\alpha)=(Q_{\alpha,4}-Q_{\alpha,1})-(Q_{0,4}-Q_{0,1})$ removes the zero-overlap selection contrast (Appendix~\ref{app:mechanism-construction}).

\suppressfloats[t]
\begin{figure}[t]\centering
\begin{subfigure}[t]{.48\linewidth}\centering\includegraphics[width=\linewidth]{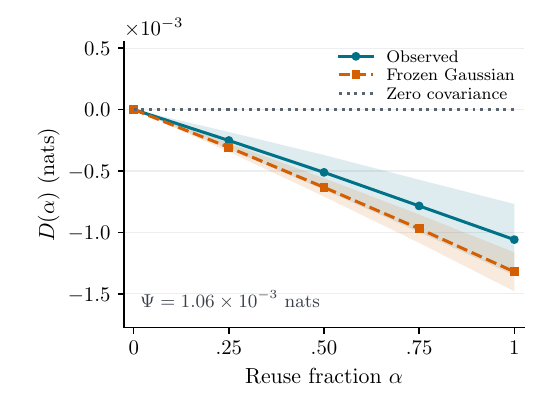}\caption{Synthetic Linear Systems}\label{fig:mech-synthetic}\end{subfigure}\hfill
\begin{subfigure}[t]{.48\linewidth}\centering\includegraphics[width=\linewidth]{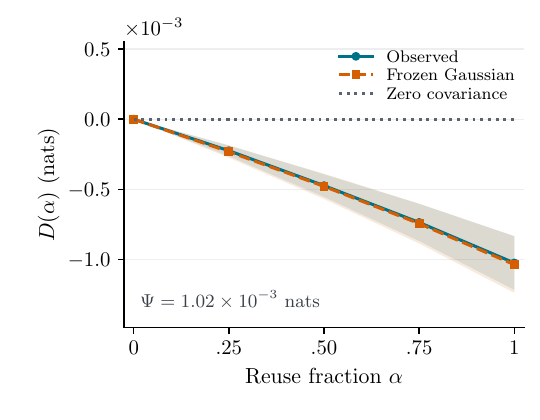}\caption{AG News}\label{fig:mech-agnews}\end{subfigure}
\caption{Controlled mechanism study of the selection--reuse interaction at fixed audit size. Holding the audit panel fixed at 32 items (suppressing the sample-size benefit of reuse), we vary the fraction $\alpha \in [0, 1]$ of audit items shared with the 32-item selection panel and measure the difference-in-differences update utility $D(\alpha) = (Q_{\alpha,4} - Q_{\alpha,1}) - (Q_{0,4} - Q_{0,1})$ in nats of mean answer-token log probability, contrasting top-of-four selection ($K=4$) against an unselected control ($K=1$) relative to zero overlap ($\alpha=0$). Shading shows pointwise 95\% bank $t$ intervals across 32 banks; dashed curves show frozen Gaussian predictions fitted on an independent 1,024-item panel; dotted lines ablate the selection-score--update-gain cross-covariance to zero. Here $\Psi = -D(1)$ is the full-overlap interaction magnitude.}\label{fig:overlap}
\end{figure}

Figure~\ref{fig:overlap} shows a nearly linear decline to $-\Psi$, with $\Psi=1.06\times10^{-3}$ nats on Synthetic Linear Systems and $1.02\times10^{-3}$ on AG News (32 banks each); an independent 80-bank Qwen3-8B AG News study gives $1.55\times10^{-3}$. Frozen Gaussian predictions fitted on 1,024 independent items track these curves. Zeroing only the selection-score--update-gain cross-covariance block, while retaining means and within-block covariances, makes the predicted interaction exactly zero. This model ablation isolates the covariance mechanism; the binary validity result in Section~\ref{sec:method} instead uses repair-sign ordering, without assuming negative mean reuse bias.

A $10^{-3}$-nat shift corresponds to about $0.1\%$ in geometric mean answer-token probability, yet is material relative to these small updates. From Table~\ref{tab:mechanism}, $\Psi$ is $24.4\%$ of AG News's net post-update selection advantage ($\Delta_{\mathrm{final}}=4.20\times10^{-3}$ nats), versus $6.5\%$ on Synthetic Linear Systems, and roughly two-thirds of the unconditional selection--gain contrast $|H|$ ($67.3\%$ and $69.5\%$, respectively). These log-utility effects motivate the power--drag trade-off in Sections~\ref{sec:method}--\ref{sec:experiments} and the $1.07$--$1.75$ pp accuracy drag observed there; nats and accuracy points are distinct scales.

\section{Exact Hybrid Certification and Completion Certificates}
\label{sec:method}

The audit combines an exact selected-winner guarantee with a deterministic acquisition rule. We establish the endpoint gate, calibrate a strengthened Fresh benchmark, and use completion certificates to avoid unnecessary queries. An iid surrogate isolates the power tradeoff.

\subsection{An exact gate that survives winner selection}
For paired binary outcomes, write $X_i=C_J(Z_i)-P_J(Z_i)$,
$b=\sum_i\ind\{X_i=1\}$ and $c=\sum_i\ind\{X_i=-1\}$ for repairs and
regressions, and $M=b+c$ for discordances. Fix horizons before child evaluation:
Hybrid uses $n_H=n_S+n_{H,F}$ pairs, where $n_{H,F}=(B-n_S)/2$;
Fresh uses $n_F=B/2$.

\begin{definition}[Full-tail endpoint gate]\label{def:endpoint}
For a cutoff $\tau\in(0,1)$, let
\begin{equation}
 p(b,c)=2^{-(b+c)}\sum_{r=b}^{b+c}\binom{b+c}{r},\qquad
 A_\tau(b,c)=\ind\{p(b,c)\leq\tau\},
 \label{eq:mcnemar}
\end{equation}
Set $p(0,0)=1$. \Hybrid\ and \FreshE\ use $\tau=\alpha=1/20$.
\end{definition}

\begin{theorem}[Independent-pair validity]\label{thm:independent-pairs}
For a fixed parent--child pair on iid rows, let
$q_+=\Pr(X_i=1)$ and $q_-=\Pr(X_i=-1)$. If $q_+\leq q_-$, then
$\Pr(A_\alpha=1\mid M=m)\leq\alpha$ for every possible $m$,
and consequently $\Pr(A_\alpha=1)\leq\alpha$.
\end{theorem}

The complete proof is in Appendix~\ref{app:indep-proof}.
Conditioning on $M$ removes the unknown discordance rate $d=q_++q_-$.
The null repair probability is at most $1/2$, so the fair-binomial tail
controls rejection for every $d\in[0,1]$, including no discordance.
Since $F$ is independent of $S$ conditional on training, this certifies
\FreshE\ given $J=j$. On reused rows, selection still affects repair signs.

Condition on training information $\mathcal M$ fixing parents and potential
children independently of $S$. Binary rows
$(P_1(Z),\ldots,P_K(Z),C_1(Z),\ldots,C_K(Z))$ are iid, with independent
Fresh rows from the same law. Select $J$ by maximizing the binary parent
score on $S$, using a fixed lowest-index tie rule. Within-row dependence is arbitrary.

\begin{theorem}[Exact selected-winner validity]\label{thm:selected-winner}
For each selectable $j$ with $\Delta_j\leq0$,
\begin{equation}
 \Pr(A^H_\alpha=1\mid J=j,\mathcal M)\leq\alpha.
 \label{eq:selected-validity}
\end{equation}
Hence $\Pr(A^H_\alpha=1,\Delta_J\leq0\mid\mathcal M)
\leq\alpha\Pr(\Delta_J\leq0\mid\mathcal M)$.
\end{theorem}

Appendix~\ref{app:selected-winner} gives the complete proof using
Proposition~\ref{prop:product-covariance}.
Checkpoints from a shared base model often succeed or fail together,
making within-item dependence essential. Conditioning
on $Q_i=P_j(Z_i)+C_j(Z_i)$ isolates independent repair signs; turning a
regression into a repair weakly lowers every parent-selection margin.
Winner probability decreases in these signs while rejection increases,
giving an association bound without estimating correlations. Mean drag
alone cannot bound this tail, and negative reuse bias is unnecessary:
Proposition~\ref{prop:selection-drag} and Remark~\ref{rem:drag-counterexample}
allow positive expected reuse-minus-test gain under a strict null.
Preassignment is sufficient for this theorem; Appendix~\ref{app:assumption-failures}
exhibits failure for one child-selection rule using $S$, without ruling
out all adaptive methods. The product law concerns conditioned signs
across items; no product law across distinct candidate checkpoints is required.

\subsection{Unconditional calibration for Fresh}
Discrete binary outcomes can make the conditional test conservative. Unconditional calibration strengthens Fresh while preserving its error guarantee.
Let $k_\tau(m)$ be the smallest repair count with full tail at most $\tau$,
or $m+1$ if none exists, and let
$a_m(\tau)=\Pr\{\operatorname{Bin}(m,1/2)\geq k_\tau(m)\}$.
At gain zero and discordance rate $d$, the iid null size is
\begin{equation}
 R_n(d;\tau)=\sum_{m=0}^{n}a_m(\tau)\binom{n}{m}d^m(1-d)^{n-m}.
 \label{eq:bernstein-size}
\end{equation}

\begin{proposition}[Rational Fresh calibration]\label{prop:bernstein}
If rational dyadic de Casteljau subdivision over $[0,1]$ yields Bernstein
coefficients bounded by $\alpha$ on every subinterval, then
$\sup_{d\in[0,1]}R_n(d;c_n)\leq\alpha$. The same cutoff controls every
nonpositive gain.
\end{proposition}

The de Casteljau proof is in Appendix~\ref{app:calib-cert}, and the exact
routine is Algorithm~\ref{alg:calibration} in Appendix~\ref{app:algo-calib};
Table~\ref{tab:fresh-certificates} records the five frozen certificates.
For iid fresh pairs, $M\sim\operatorname{Bin}(n,d)$ averages the discrete
conditional sizes; rational Bernstein bounds certify the maximum over
$d$. Thus \FreshU\ can raise its cutoff above $\alpha$: some conditional
sizes may exceed $\alpha$ while the unconditional size does not.
Banking77-Mixed uses $c_{128}=c_{167}=57/1000$. Selection changes the
mixture weights on reused rows, preventing this transfer to Hybrid,
which retains the conditional cutoff $\alpha$.
Certification covers the full nuisance-parameter interval, so the cutoff
is fixed independently of the observed repair and regression counts on the audit panel.

\subsection{Completion certificates}
Some decisions become inevitable before the endpoint, making the remaining queries unnecessary. At a partial audit, let $u_0,u_1$ count unread selection items with known
parent scores zero and one, and let $f$ count unread fresh pairs.

\begin{proposition}[Exact completion equivalence]\label{prop:completion}
The attainable extrema are
\begin{equation}
 p_{\min}=p(b+u_0+f,c),\qquad p_{\max}=p(b,c+u_1+f).
 \label{eq:completion-bounds}
\end{equation}
Accept if $p_{\max}\leq\tau$ and reject if $p_{\min}>\tau$.
Otherwise both actions remain possible. Every returned action equals
the fixed-endpoint action on the same complete stream.
\end{proposition}

Appendix~\ref{app:calib-cert} proves monotonicity and attainability;
Appendix~\ref{app:stopping-order} gives sharp order-dependent cost bounds.
Unread parent-one items cannot repair errors, so early regressions can
make even the favorable completion fail, rapidly rejecting degrading
updates. Parent-zero items can repair errors; fresh pairs allow either
sign and cost two calls. Algorithm~\ref{alg:completion} identifies an
inevitable endpoint action, preserving its size and power. Its stopping
rule preserves the original error guarantee without changing the testing
boundary or spending an additional error budget.

\begin{algorithm}[t]
\caption{Decision-preserving completion certificate}\label{alg:completion}
\small
\begin{algorithmic}[1]
\Require Ordered parent scores on $S$, fresh horizon $n_{H,F}$, cutoff $\tau$
\State $b,c,q\gets0$; $f\gets n_{H,F}$; $(u_0,u_1)\gets$ unread parent counts
\While{true}
  \State \textbf{if} $p(b,c+u_1+f)\leq\tau$ \textbf{then return} $(\mathrm{accept},q)$
  \State \textbf{if} $p(b+u_0+f,c)>\tau$ \textbf{then return} $(\mathrm{reject},q)$
  \If{$u_0+u_1>0$}
    \State Read next parent score $y$ and evaluate child score $z$
    \State $u_y\gets u_y-1$; $q\gets q+1$
  \Else
    \State Evaluate the next fresh pair $(y,z)$; $f\gets f-1$; $q\gets q+2$
  \EndIf
  \State $b\gets b+\ind\{y=0,z=1\}$; $c\gets c+\ind\{y=1,z=0\}$
\EndWhile
\end{algorithmic}
\end{algorithm}

\subsection{A tractable power boundary}\label{sec:power-boundary}
Power balances extra pairs against selection drag. For an iid three-point law,
take gain $|\Delta|\leq d$, discordance $d>0$, and conditional repair
probability $\theta=(d+\Delta)/(2d)$:
\begin{equation}
 \pi_n(\Delta,d;\tau)=\sum_{m=0}^{n}\binom{n}{m}d^m(1-d)^{n-m}
 \Pr\{\operatorname{Bin}(m,\theta)\geq k_\tau(m)\}.
 \label{eq:surrogate-power}
\end{equation}
The mean-shift surrogate represents Hybrid by an iid law with
$\Delta_{\mathrm{eff}}=\Delta-(n_S/n_H)\delta_S$, for positive drag
$\delta_S=\Delta-\Delta_S$. At fixed deployment gain,
\begin{equation}
 U_H-U_F=\Delta\{\pi_{n_H}(\Delta_{\mathrm{eff}},d;\alpha)
                  -\pi_{n_F}(\Delta,d;c_{n_F})\}.
 \label{eq:surrogate-utility}
\end{equation}
For common cutoffs, fixed $d>0$, and local gains, the normal approximation
gives the crossover
\begin{equation}
 \delta_S^*(\Delta)\simeq
 \frac{n_H}{n_S}\left(1-\sqrt{\frac{n_F}{n_H}}\right)\Delta
 =\frac{\Delta}{2\{1+\sqrt{r/(r+1)}\}},\qquad r=\frac{B}{n_S}\geq1.
 \label{eq:crossover-main}
\end{equation}
Within this approximation, reuse has higher power at $r=1$ for drag
below $(1-1/\sqrt2)\Delta\approx0.293\Delta$; the slope tends to $0.25$
as $r$ grows. Utility differences compress as both powers approach one.
Appendix~\ref{app:surrogate} derives the boundary and downward correction
for calibrated Fresh. These iid calculations do not establish dominance for selected audits.
The comparison holds discordance and deployment gain fixed; changing
either also changes how rapidly the policies reach the region in which
both audits are near power one.
\begin{table}[t]
\centering\small
\setlength{\tabcolsep}{2.4pt}
\caption{Prospective deployment: safety-constrained accuracy and virtual audit savings.}\label{tab:main-outcomes}
\begin{tabular*}{\linewidth}{@{\extracolsep{\fill}}lcrrrrrrrrrr@{}}
\toprule
\multicolumn{12}{@{}l}{\textbf{Panel A. Policy outcomes, grouped by finite-sample validity}} \\
\addlinespace[2pt]
 & & \multicolumn{5}{c}{Banking77-Mixed ($N=48$, $B=256$)} & \multicolumn{5}{c}{SIGMORPHON-Mixed ($N=64$, $B=334$)} \\
\cmidrule(lr){3-7}\cmidrule(l){8-12}
Policy & \shortstack{5\%\\Valid} & Acc.\% & Gain (pp) & G1 & G2 & Saved & Acc.\% & Gain (pp) & G1 & G2 & Saved \\
\midrule
\multicolumn{12}{@{}l}{\itshape Certified 5\% finite-sample audits} \\
Hybrid & $\checkmark$ & 79.01 & 5.37 & 22/24 & 0/24 & \textbf{96.1} & 62.04 & 4.51 & 26/32 & 1/32$\dagger$ & \textbf{45.5} \\
Fresh-U & $\checkmark$ & 78.79 & 5.14 & 21/24 & 0/24 & 48.2 & 62.27 & 4.74 & 28/32 & 0/32 & 25.6 \\
Fresh-E & $\checkmark$ & 78.79 & 5.14 & 21/24 & 0/24 & 48.6 & 62.12 & 4.59 & 27/32 & 0/32 & 26.1 \\
\midrule
\multicolumn{12}{@{}l}{\itshape Uncertified reuse heuristics} \\
Naive-Z & $\times$ & 79.01 & 5.37 & 22/24 & 0/24 & 0.0 & 62.29 & 4.76 & 28/32 & 1/32 & 0.0 \\
Naive-Mean & $\times$ & 79.51 & 5.87 & 24/24 & 0/24 & 0.0 & 62.76 & 5.23 & 32/32 & 4/32 & 0.0 \\
\midrule
\multicolumn{12}{@{}l}{\itshape No-audit baselines} \\
AU & $\times$ & 62.21 & $-11.43$ & 24/24 & 24/24 & -- & 60.12 & 2.59 & 32/32 & 32/32 & -- \\
NU & -- & 73.64 & 0.00 & 0/24 & 0/24 & -- & 57.53 & 0.00 & 0/32 & 0/32 & -- \\
\bottomrule
\end{tabular*}
\par\medskip

\begin{tabular*}{\linewidth}{@{\extracolsep{\fill}}llrrrr@{}}
\toprule
\multicolumn{6}{@{}l}{\textbf{Panel B. Headline prospective contrasts (Hybrid minus comparator)}} \\
\addlinespace[2pt]
Study / family & Comparator & $\widehat\Gamma$ (pp) & Cohort 95\% CI (pp) & $p_+$ (Holm) & $p_2$ (Holm) \\
\midrule
\multirow{3}{*}{\shortstack[l]{Banking77-Mixed\\Primary, $B=256$ (H4)}} & AU & 16.80 & $[11.21, 22.40]$ & $3.52\!\times\!10^{-7}$ & $7.03\!\times\!10^{-7}$ \\
 & NU & 5.37 & $[3.62, 7.13]$ & $3.10\!\times\!10^{-7}$ & $6.20\!\times\!10^{-7}$ \\
 & Fresh-U & 0.23 & $[-0.90, 1.36]$ & $0.6877$ & $1$ \\
\addlinespace[4pt]
\multirow{3}{*}{\shortstack[l]{SIGMORPHON-Mixed\\Pooled, $B=334$ (H12)}} & AU & 1.92 & $[0.52, 3.32]$ & $0.0353$ & $0.0707$ \\
 & NU & 4.51 & $[3.09, 5.94]$ & $1.83\!\times\!10^{-7}$ & $3.66\!\times\!10^{-7}$ \\
 & Fresh-U & $-0.22$ & $[-0.72, 0.27]$ & $1$ & $1$ \\
\bottomrule
\end{tabular*}

\par\vspace{4pt}\noindent\begin{minipage}{\linewidth}\footnotesize
Hybrid denotes HybridCertified; $\checkmark$: finite-sample false-upgrade probability $\leq5\%$ under the stated assumptions; $\times$: no such guarantee. G2 pools G2a/G2b; counts are accepted/eligible cohorts. Saved counts mean virtual incremental endpoint calls; heuristics use the full panel (0.0 saved); AU/NU require no audit. Fresh-U/E denote FreshUnconditional/Exact.
$\dagger$The accepted G2b cohort c030 has positive independent test gain ($+1.86\pp$). Among observed $\widehat\Delta_T\leq0$ cohorts, Hybrid accepts $0/29$ in inflection and $0/24$ in Banking77; Naive-Mean accepts $3/29$ in inflection. These are post-hoc diagnostics, not population Type-I proofs.
Cohort $t$ intervals are approximate. H12 adjusts each direction over twelve inflection contrasts; H4 uses the separate four-contrast Banking77 primary family. Neither study establishes accuracy superiority or equivalence over calibrated Fresh. Full families, secondary budgets, Fresh-E, Welch sensitivity, and conditional intervals: Appendix~\ref{app:precision}, Tables~\ref{tab:mixed}--\ref{tab:precision-all}.
\end{minipage}
\end{table}
\section{Experiments}\label{sec:experiments}
\subsection{Experimental Protocol and Research Questions}\label{sec:protocol}
We ask three questions. \textbf{Q1:} Can Hybrid block realistic LLM regressions while preserving beneficial updates under a strict 5\% false-upgrade guarantee (Section~\ref{sec:results})? \textbf{Q2:} How large is selection drag, and when does reuse improve accuracy over calibrated Fresh (Section~\ref{sec:empirical-crossover})? \textbf{Q3:} Why use exact gates, and how much evaluation can completion certificates save (Section~\ref{sec:stress})?

SIGMORPHON-Mixed comprises 40 Qwen3-8B and 24 Mistral-7B morphological-inflection cohorts at primary budget $B=334$. Banking77-Mixed comprises 48 Qwen3-8B intent-classification cohorts at primary $B=256$ and secondary $B=334$. Both fix a 50/25/25 mixture of balanced beneficial training (G1), support-shift forgetting (G2a), and repeated small-support training (G2b). Policies are blind to recipe assignments. Disjoint pools supply parent training, child updates, independent calibration, and evaluation. Each cohort selects among three parents on $n_S=256$ items; selection, fresh-audit, and 2,048-draw test streams are independent draws with replacement from a fixed 2,000-item evaluation frame. Calibration fixes recipes before the main study; all audit actions are frozen before test analysis.

Training uses rank-32 LoRA, scaling 64, zero adapter dropout, fresh AdamW, batch size 32, and greedy decoding; malformed and truncated outputs count as errors. Deployment utility averages paired cohort accuracy gains relative to NoUpdate. The 112 historical inflection cohorts and log-likelihood studies supply complementary evidence on power-limited updates and selection drag.

Primary inference uses ordinary paired cohort $t$ intervals and separate one- and two-sided Holm adjustments: one twelve-contrast inflection family, and separate four-contrast Banking77 families at each budget. Fixed-allocation Welch sensitivity and conditional simultaneous Hoeffding intervals separate cohort uncertainty from independent test-item sampling given fitted models and frozen actions (Appendices~\ref{app:protocol} and~\ref{sec:proof-inference}).

\begin{figure}[t]\centering
\begin{subfigure}[t]{.32\linewidth}\centering\includegraphics[height=4.25cm]{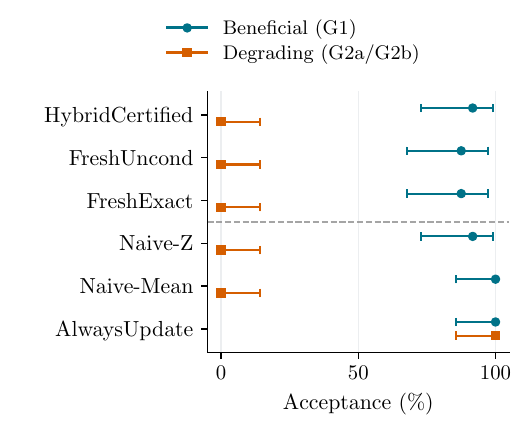}\caption{Primary-budget acceptance}\label{fig:bank-accept}\end{subfigure}\hfill
\begin{subfigure}[t]{.32\linewidth}\centering\includegraphics[height=4.25cm]{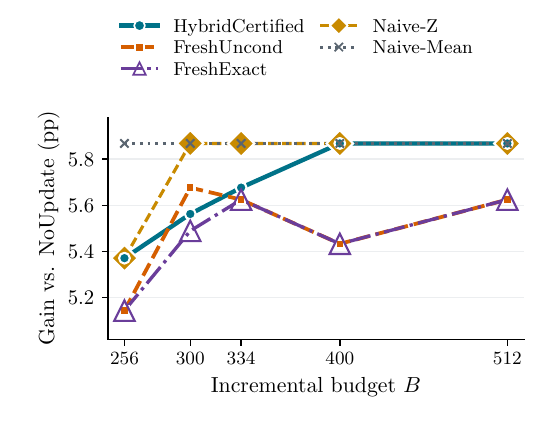}\caption{Deployment gain by budget}\label{fig:bank-budget}\end{subfigure}\hfill
\begin{subfigure}[t]{.32\linewidth}\centering\includegraphics[height=4.25cm]{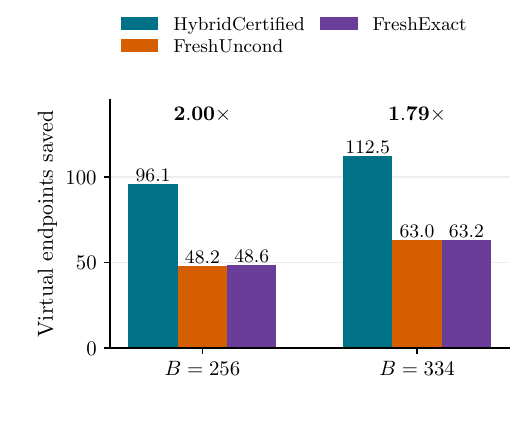}\caption{Virtual evaluations saved}\label{fig:bank-saved}\end{subfigure}
\caption{Banking77-Mixed: (a) 24 G1 and 24 pooled G2a/G2b cohorts; (b) nested-budget policies; (c) virtual completion savings at $B=256,334$. Dashed dividers separate certified 5\% audits (top) from uncertified heuristics and AlwaysUpdate (bottom).}\label{fig:bank}
\par\addvspace{\floatsep}
\setcounter{subfigure}{0}
\centering
\begin{subfigure}[t]{.32\linewidth}\centering\includegraphics[height=4.8cm]{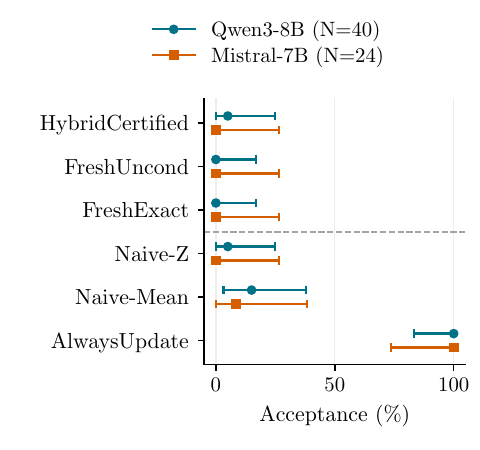}\caption{G2 acceptance}\label{fig:inflect-g2}\end{subfigure}\hfill
\begin{subfigure}[t]{.32\linewidth}\centering\includegraphics[height=4.8cm]{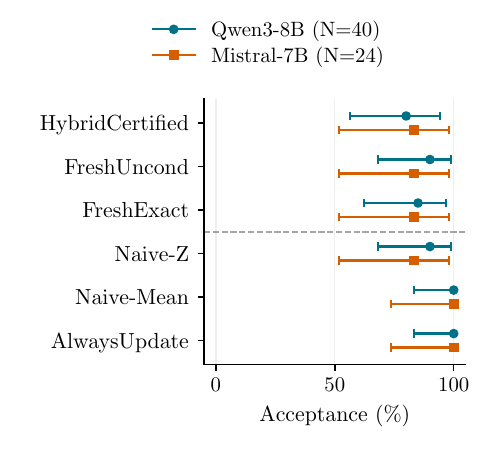}\caption{G1 acceptance}\label{fig:inflect-g1}\end{subfigure}\hfill
\begin{subfigure}[t]{.34\linewidth}\centering\includegraphics[height=4.8cm]{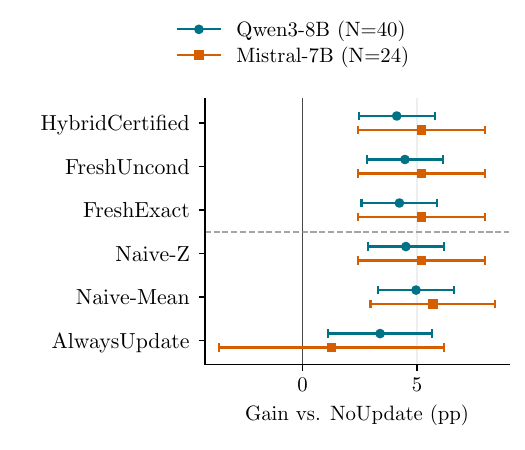}\caption{Deployment gain}\label{fig:inflect-gain}\end{subfigure}
\caption{SIGMORPHON-Mixed ($B=334$). Acceptance: pointwise binomial references within assigned recipe groups. Gains: ordinary cohort 95\% $t$ intervals. Dashed dividers mark the same validity tiers as Figure~\ref{fig:bank}.}\label{fig:mixed}
\end{figure}
\begin{table}[t]
\centering\small
\setlength{\tabcolsep}{3.5pt}
\caption{Selection drag and accuracy gains in power-limited benchmarks. All gains, optimism, and drag are in pp.}\label{tab:main-drag}
\begin{tabular*}{\linewidth}{@{\extracolsep{\fill}}lrrrrrrr@{}}
\toprule
\multicolumn{8}{@{}l}{\textbf{Panel A. Cross-study winner's curse and selection drag}} \\
\addlinespace[2pt]
Study / subset & $N$ & $\widehat\Delta_T$ & $O_P$ (SE) & $p_2$ & $O_C$ & Drag & $d_S$ \\
\midrule
SIGMORPHON-I & 48 & 9.15 & 1.36 (0.41) & $0.0016$ & 0.22 & $-1.14$ & 0.212 \\
SIGMORPHON-II & 64 & 9.07 & 1.30 (0.36) & $7.03\!\times\!10^{-4}$ & 0.23 & $-1.07$ & 0.210 \\
Mixed: Qwen & 40 & 3.39 & 1.51 (0.40) & $5.31\!\times\!10^{-4}$ & $-0.07$ & $-1.59$ & 0.214 \\
Mixed: Mistral & 24 & 1.26 & 3.27 (0.62) & $2.44\!\times\!10^{-5}$ & 1.52 & $-1.75$ & 0.303 \\
Mixed: pooled & 64 & 2.59 & 2.17 (0.36) & $6.96\!\times\!10^{-8}$ & 0.52 & $-1.65$ & 0.248 \\
Mixed: pooled G1 & 32 & 10.45 & 1.96 (0.49) & $3.40\!\times\!10^{-4}$ & 0.54 & $-1.42$ & 0.245 \\
Banking77-Mixed & 48 & $-11.43$ & 1.69 (0.36) & $2.47\!\times\!10^{-5}$ & 0.53 & $-1.16$ & 0.339 \\
Banking77 G1 & 24 & 11.74 & 1.53 (0.47) & $0.0034$ & 0.68 & $-0.85$ & 0.197 \\
\bottomrule
\end{tabular*}

\par\medskip
\setlength{\tabcolsep}{2.2pt}
\begin{tabular*}{\linewidth}{@{\extracolsep{\fill}}lrrrrrr@{}}
\toprule
\multicolumn{7}{@{}l}{\textbf{Panel B. Power-limited beneficial updates: Hybrid $-$ Fresh-U ($B=334$, retrospective)}} \\
\addlinespace[2pt]
Study / synthesis & $N$ & Hybrid accept & Fresh-U accept & $\widehat\Gamma$ & Cohort 95\% CI & $p_2$ \\
\midrule
SIGMORPHON-I & 48 & 44/48 (91.7\%) & 36/48 (75.0\%) & $+1.57$ & $[+0.37,+2.78]$ & 0.0118 \\
SIGMORPHON-II & 64 & 55/64 (85.9\%) & 50/64 (78.1\%) & $+0.62$ & $[-0.26,+1.51]$ & 0.1659 \\
SIGMORPHON-Hist & 112 & 99/112 (88.4\%) & 86/112 (76.8\%) & $\mathbf{+1.03}$ & $[+0.31,+1.75]$ & 0.0053 \\
Beneficial-168 & 168 & 148/168 (88.1\%) & 137/168 (81.5\%) & $+0.60$ & $[+0.06,+1.15]$ & 0.0305 \\
\bottomrule
\end{tabular*}
\par\vspace{4pt}\noindent\begin{minipage}{\linewidth}\footnotesize $O_P$ and $O_C$ are selected-parent and child optimism; Drag is $\widehat\Delta_S-\widehat\Delta_T=O_C-O_P$. The $p_2$ column reports unadjusted paired $t$ tests of parent optimism. Mixed denotes SIGMORPHON-Mixed; G1 is the balanced recipe. Population and G1 rows overlap. Table~\ref{tab:drag} retains the full ten-row audit, including both family-specific G1 rows. Panel B retains historical token/recovery caps; SIGMORPHON-I's Fresh comparator was strengthened post-hoc. Beneficial-168 retrospectively appends assigned G1 cohorts, without test-sign filtering. Panel B intervals and unadjusted cohort $t$ tests are descriptive and separate from Table~\ref{tab:main-outcomes}'s prospective Holm families.\end{minipage}

\end{table}
\begin{figure}[t]
\centering
\begin{subfigure}[t]{.455\linewidth}\centering\includegraphics[height=5.5cm]{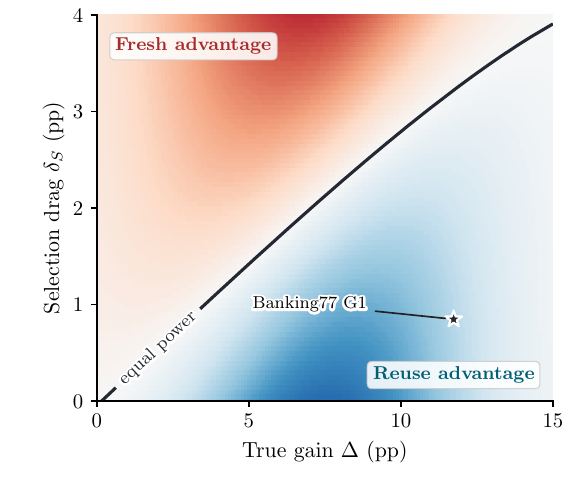}\caption{$B=256$}\end{subfigure}\hfill
\begin{subfigure}[t]{.535\linewidth}\centering\includegraphics[height=5.5cm]{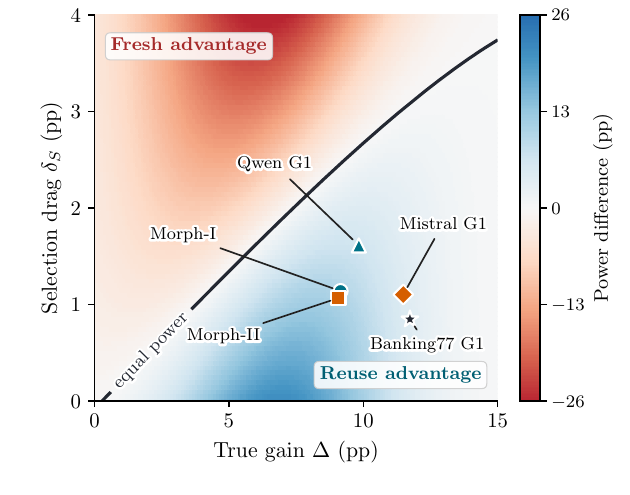}\caption{$B=334$}\end{subfigure}
\caption{Exact iid mean-shift surrogate power difference at $d=0.22$: black denotes equal power; markers are retrospective means (Banking77 G1: $d_S=0.197$). Morph-I/II denote historical inflection.}\label{fig:phase}
\par\addvspace{\floatsep}
\setcounter{subfigure}{0}
\centering
\begin{subfigure}[t]{.48\linewidth}\centering\includegraphics[width=\linewidth]{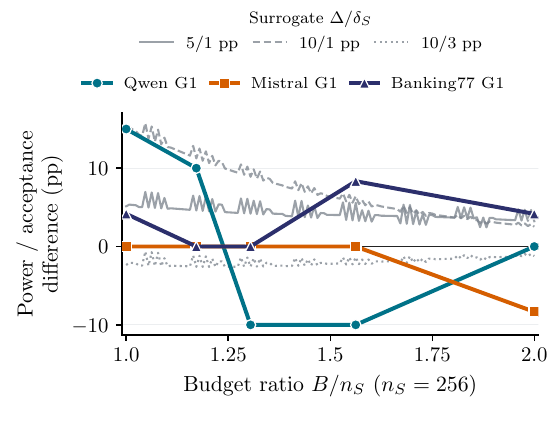}\caption{Acceptance difference on G1}\label{fig:crossover-accept}\end{subfigure}\hfill
\begin{subfigure}[t]{.48\linewidth}\centering\includegraphics[width=\linewidth]{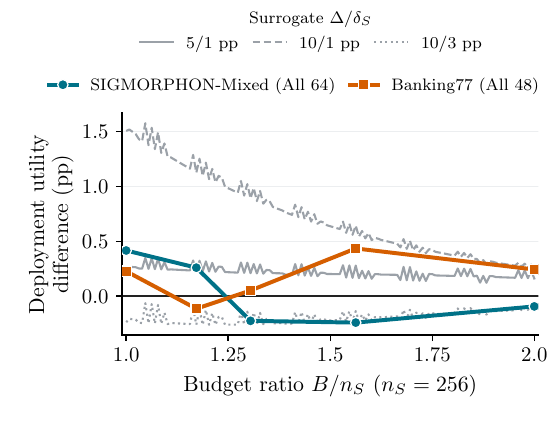}\caption{Utility difference on all cohorts}\label{fig:crossover-utility}\end{subfigure}
\caption{Budget-dependent reuse advantage. Gray: iid surrogates at stated gain/drag; color: fixed-policy replays. Acceptance uses G1 subsets; utility uses all 64 inflection and 48 intent cohorts.}\label{fig:crossover}

\end{figure}
\subsection{Selective Deployment Utility Across Families and Domains}\label{sec:results}
\paragraph{Safety-constrained comparison and why heuristics fail.}
Deployment auditing must preserve accuracy while controlling false upgrades. Table~\ref{tab:main-outcomes} therefore separates certified 5\% audits from uncertified heuristics and no-audit baselines, with the tight-budget Banking77 comparison first. Naive-Mean's higher raw inflection accuracy comes with four accepted G2 updates, three with negative independent test gain, and null false-upgrade rates of roughly 33\%--50\%. Naive-Z also lacks a finite-sample 5\% guarantee: its small-discordance inflation is quantified in Section~\ref{sec:stress}. Both heuristics consume the full panel and save zero calls. Their raw means do not solve the same safety-constrained problem.

\paragraph{Strong protection and oracle headroom among certified 5\% audits.}
At Banking77's canonical $B=n_S=256$ budget, reuse buys twice as many pairs. Hybrid attains the highest observed accuracy and G1 acceptance among certified audits, ties their zero G2 acceptance, and saves the most calls (Table~\ref{tab:main-outcomes}; Figure~\ref{fig:bank}). It gains $16.80\pp$ over AlwaysUpdate, averting that baseline's $11.43\pp$ loss, and $5.37\pp$ over NoUpdate. Its $79.01\%$ accuracy exceeds Fresh-U's $78.79\%$ in this sample while saving $2.00\times$ as many virtual calls.

In SIGMORPHON-Mixed, Hybrid gains $1.92\pp$ over AlwaysUpdate and $4.51\pp$ over NoUpdate (Figure~\ref{fig:mixed}). The AlwaysUpdate contrast passes one-sided Holm-12 ($p_+=0.0353$), but not two-sided Holm-12 ($p_2=0.0707$). Hybrid captures $71.6\%$ of inflection's $2.69\pp$ hindsight-oracle headroom over AlwaysUpdate, and $97.1\%$ in Banking77. Relative to NoUpdate, the oracle ceilings are $5.27$ and $5.87\pp$, of which Hybrid captures $85.6\%$ and $91.5\%$. These ceilings use realized independent-test accuracy.

Hybrid accepts only $1/32$ assigned G2 inflection updates and $0/24$ in Banking77. The single accepted G2b cohort, c030, has \emph{positive} test gain ($+1.86\pp$). Among observed nonpositive-gain updates, Hybrid accepts $0/29$ and $0/24$: post-hoc diagnostics supporting its practical protection, with population validity supplied separately by Theorem~\ref{thm:selected-winner}.

\paragraph{Why prospective accuracy is close to calibrated Fresh---and where reuse still wins.}
Large beneficial gains leave limited acceptance headroom. Mean G1 gains are $9.83\pp$ for Qwen inflection, $11.49\pp$ for Mistral inflection, and $11.74\pp$ for Banking77. Fresh-U already accepts $18/20$, $10/12$, and $21/24$ ($90.0\%$, $83.3\%$, and $87.5\%$). All three exact policies agree on all 24 Mistral cohorts. Hybrid recovers three Banking77 G1 updates missed by Fresh128, but misses two under strong reuse drag; Qwen disagreements likewise locate the budget boundary (Appendix~\ref{app:cohort-diagnostics}).

Neither prospective study establishes deployment-accuracy superiority or equivalence over calibrated Fresh: Hybrid-minus-Fresh-U gains are $-0.22\pp$ in inflection and $+0.23\pp$ in Banking77. The efficiency advantage remains clear. Hybrid saves $45.5$ versus $25.6$ calls in inflection and $96.1$ versus $48.2$ in Banking77: $1.774\times$ and $1.996\times$ as many saved virtual incremental endpoint calls. Banking77 needs $23.1\%$ fewer remaining audit calls, with every endpoint decision preserved.

\begin{figure}[t]\centering
\begin{subfigure}[t]{.45\linewidth}\centering\includegraphics[height=5.1cm]{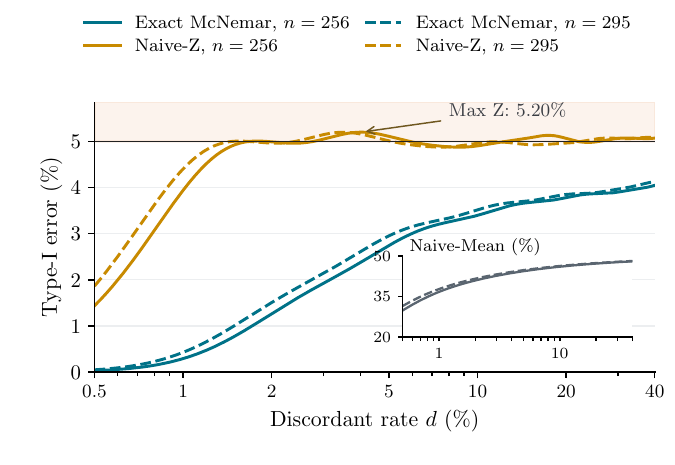}\caption{Exact known-law null rejection}\end{subfigure}\hfill
\begin{subfigure}[t]{.54\linewidth}\centering\includegraphics[height=5.1cm]{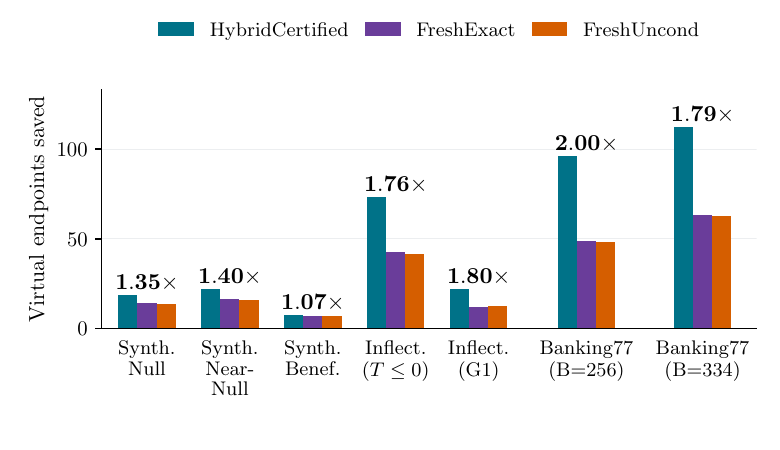}\caption{Completion-certificate savings}\end{subfigure}
\caption{(a) Exact known-law rejection at two fixed horizons (Naive-Mean inset). (b) Mean virtual endpoints saved. The nonpositive inflection test-gain subset is a post-hoc diagnostic.}\label{fig:null}
\end{figure}
\begin{table}[t]
\centering\small
\setlength{\tabcolsep}{6pt}
\caption{Exact null boundaries and virtual savings from decision-preserving completion.}\label{tab:main-safety}\label{tab:null}\label{tab:savings}
\begin{tabular*}{\linewidth}{@{\extracolsep{\fill}}rlrr@{}}
\toprule
\multicolumn{4}{@{}l}{\textbf{Panel A. Independent-pair null maxima, $d\in[0.005,0.40]$}} \\
\addlinespace[2pt]
Pairs & Endpoint rule & Maximum size (\%) & Witness $d$ \\
\midrule
256 & Exact McNemar & 4.0488 & 0.40000 \\
256 & Naive-Z & 5.2025 & 0.04031 \\
295 & Exact McNemar & 4.1272 & 0.40000 \\
295 & Naive-Z & 5.2013 & 0.03499 \\
\bottomrule
\end{tabular*}
\par\medskip
\setlength{\tabcolsep}{3.4pt}
\begin{tabular*}{\linewidth}{@{\extracolsep{\fill}}lrrrrrrr@{}}
\toprule
\multicolumn{8}{@{}l}{\textbf{Panel B. Mean virtual endpoint calls saved}} \\
\addlinespace[2pt]
Population & $B$ & Hybrid & Fresh-E & Fresh-U & H/$B$ (\%) & H/FU & $p_2$ \\
\midrule
Synthetic: null & 334 & 18.6 & 14.2 & 13.7 & 5.6 & 1.351 & -- \\
Synthetic: near null & 334 & 22.0 & 16.2 & 15.7 & 6.6 & 1.403 & -- \\
Synthetic: beneficial & 334 & 7.4 & 7.2 & 6.9 & 2.2 & 1.068 & -- \\
\addlinespace[2pt]
Mixed: all & 334 & 45.5 & 26.1 & 25.6 & 13.6 & 1.774 & $3.28\!\times\!10^{-11}$ \\
Mixed: $\widehat\Delta_T\leq0$ & 334 & 73.4 & 42.6 & 41.7 & 22.0 & 1.762 & $2.72\!\times\!10^{-10}$ \\
Mixed: G1 & 334 & 22.2 & 12.2 & 12.3 & 6.6 & 1.802 & $7.64\!\times\!10^{-4}$ \\
\addlinespace[2pt]
Banking77-Mixed & 256 & 96.1 & 48.6 & 48.2 & 37.6 & 1.996 & $2.55\!\times\!10^{-10}$ \\
Banking77: G2a & 256 & 166.8 & 77.7 & 76.8 & \textbf{65.2} & 2.171 & -- \\
Banking77: G2b & 256 & 182.7 & 99.0 & 97.8 & \textbf{71.4} & 1.867 & -- \\
Banking77-Mixed & 334 & 112.5 & 63.2 & 63.0 & 33.7 & 1.787 & $6.24\!\times\!10^{-12}$ \\
\bottomrule
\end{tabular*}
\par\vspace{4pt}\noindent\begin{minipage}{\linewidth}\footnotesize Panel A uses exact rational supremum enclosures of width at most $10^{-10}$; reported maxima retain the certified upper-bound rounding. Selected-winner validity is established separately at the nominal 5\% level. Panel B: H denotes Hybrid; H/$B$ is its fraction of the full budget saved, and H/FU compares saved calls, not total costs. Synthetic rows average 80 cells per role; empirical paired $p_2$ values are unadjusted descriptive tests. Dashes in $p_2$ mark synthetic cell averages or descriptive G2a/G2b rows. The observed nonpositive-test-gain subset is post-hoc. Savings preserve the frozen endpoint decisions and original query order; they are virtual endpoint savings, not measured runtime or monetary savings.\end{minipage}
\end{table}
\subsection{Winner's Curse and the Budget--Drag Crossover}\label{sec:empirical-crossover}
\paragraph{Measured drag locates the accuracy--power tradeoff.}
Across full-study and family rows, selected-parent optimism is $1.30$--$3.27\pp$ and reuse-minus-test drag is $-1.75$ to $-1.07\pp$ (Table~\ref{tab:main-drag}, Panel A). Child optimism is smaller, so reuse understates the update's gain on average. Figure~\ref{fig:phase} places the plotted study-level means in the blue reuse-advantage region of an exact iid mean-shift surrogate. Prospective G1 means lie far to the right, around $10$--$12\pp$, where both audits approach saturation; larger power gains occur for harder updates around $4$--$8\pp$. The surrogate makes the sample-size dividend and drag directly comparable; its approximately linear boundary is derived in Section~\ref{sec:power-boundary}. Study means do not imply a universal bias sign or locate every individual cohort.

\paragraph{Power-limited regime: historical accuracy gains.}
Table~\ref{tab:main-drag}, Panel B makes the complementary empirical advantage explicit. Across 112 historical inflection cohorts, Hybrid accepts $99/112$ updates ($88.4\%$) versus Fresh-U's $86/112$ ($76.8\%$), delivering $+1.03\pp$ accuracy (95\% cohort interval $[0.31,1.75]\pp$). This closes over half of Fresh's remaining gap to the $9.11\pp$ hindsight ceiling. Every historical cohort has positive observed test gain, so this oracle coincides with AlwaysUpdate.

Appending prospectively assigned G1 cohorts gives retrospective 144- and 168-cohort gains of $0.69$ and $0.60\pp$; inclusion uses the recipe, not test-gain signs. Table~\ref{tab:main-drag}'s footnote and Appendix~\ref{app:retrospective-synthesis} give the historical caps, post-hoc comparator, and descriptive-inference scope.

\paragraph{Budget replays reveal the crossover.}
Figure~\ref{fig:crossover} tracks the same prospective cohorts across nested budgets. SIGMORPHON-Mixed gives Hybrid-minus-Fresh-U gains of $+0.42\pp$ at counterfactual $B=256$, $-0.22\pp$ at primary $B=334$, and $-0.09\pp$ at $B=512$; Banking77 gives $+0.23$, $+0.05$, and $+0.24\pp$. Discrete cutoffs and heterogeneous gains make utility nonmonotone. These descriptive sweeps use five frozen fresh certificates and 129 offline calibrations, distinguished in Appendix~\ref{app:offline-calibration-milestones}.

\subsection{Exact Null Boundaries and Completion-Certificate Efficiency}\label{sec:stress}
\paragraph{Exact gates avoid sparse-discordance failures.}
Normal approximation can fail precisely when parent and child predictions agree closely. With only $m=3$ discordant pairs, Naive-Z has a $12.50\%$ conditional null false-upgrade rate, whereas exact McNemar rejects $0\%$ (Table~\ref{tab:small-discordance}). Mixing over discordance counts on independent pairs with $d\in[0.005,0.40]$ still gives Naive-Z exact null maxima of $5.2025\%$ at $n=256$ and $5.2013\%$ at $n=295$, exceeding 5\%; exact McNemar maxima are $4.0488\%$ and $4.1272\%$ (Figure~\ref{fig:null}; Table~\ref{tab:main-safety}). Theorem~\ref{thm:selected-winner} supplies validity after winner selection.

Across 240 stress-test cells and 12 million replicates, observed Naive-Z null and near-null acceptance stays below 5\% for $K\ge2$, even at within-row dependence $\rho=.95$. Selection can mask normal-test inflation; this protection disappears at $K=1$ or identical candidates (Appendix~\ref{app:mc-grid}). Exact McNemar avoids relying on that masking and stays valid when discordance is sparse.

\paragraph{Exact monotonicity turns decisive evidence into saved calls.}
Completion certificates exploit the gate's monotonicity: known parent successes cannot become repairs, so a degrading update can be rejected as soon as even its best feasible completion fails. Reuse needs only one child call per pair; fresh acquisition needs two. In Banking77, Hybrid skips \textbf{65.2\%} of the full $B=256$ budget on G2a and \textbf{71.4\%} on G2b (Table~\ref{tab:main-safety}, Panel B). At $B=334$, its overall saved-call advantage remains $1.787\times$.

Every endpoint action is preserved across 6,534 exhaustive cases, 12 million simulated paths, and 912 empirical replays. Synthetic cell savings ratios span $0.895$--$1.852$, so efficiency depends on the stream. All savings are \emph{virtual incremental endpoint calls} under the frozen query order: complete streams were acquired for common-policy comparisons, and runtime, token, and monetary savings were not measured.

\section{Discussion, Limitations, and Conclusion}\label{sec:conclusion}

Choose between \Hybrid\ and \FreshU\ according to whether extra pairs compensate for selection drag. Reuse is attractive when cached parent labels and preassigned updates expand a power-limited fresh budget; Fresh can be preferable under strong drag, saturated power, or selection-adaptive recipes. Calibrate gates and budgets independently, and fix scoring, child recipes, and query order before observing $S$. Retain cached labels; only the winner's child needs training. Completion preserves either valid endpoint's action on every complete stream, stopping when all feasible continuations agree.

Conditional on training information, the exact selected-winner guarantee requires iid rows, identical binary scoring for selection and audit, maximum-score parent selection with a fixed lowest-index tie rule, and potential updates preassigned independently of the selection panel. Fresh rows must be independent draws from the same law; candidate outcomes may nevertheless be arbitrarily dependent within a row. This accommodates common checkpoint errors without making the selected stream iid or covering correlated prompts and clustered examples. Selection-adaptive recipes, dependent rows, and repeated deployment need further analysis: single-update control does not bound accumulated errors across deployments. The protective tail ordering does not require negative mean reuse bias, which can reverse even under a strict null. The iid power map represents the sample-size--drag tradeoff, not the selected joint law.

The accuracy evidence covers two model families and domains, fixed recipe proportions, and finite frames. Family and execution host are confounded, preventing architectural attribution. Cohort $t$ inference is approximate under fixed allocations; conditional intervals quantify only test-item uncertainty given fitted models and actions. The post-hoc $0/29$ and $0/24$ counts do not prove population validity; nonsignificance or identical actions do not establish equivalence, and prospective superiority over calibrated Fresh remains unresolved. Retrospective syntheses and replays remain separate from prospective testing families. Data separation cannot exclude semantic overlap or pretraining contamination. For judge-based and open-ended evaluation, exact-match success and judge approval define different binary targets. Extending the finite-frame guarantee to production requires scoring and sampling that meet the stated assumptions.

Under these conditions, winner selection provides exact finite-sample false-upgrade control conditional on the winner. Reuse buys more pairs; completion removes queries only when all possible completions force the same endpoint action. Savings are virtual, not measured runtime, tokens, or cost. Power and drag determine reuse's accuracy value while its error guarantee remains intact.
\FloatBarrier
\label{main:lastpage}
\clearpage
\linespread{1}\selectfont
\setlength{\parskip}{0.45\baselineskip}
\setlength{\textfloatsep}{16pt plus 3pt minus 2pt}
\setlength{\floatsep}{14pt plus 2pt minus 2pt}
\bibliographystyle{plainnat}
\begingroup\small
\setlength{\bibsep}{2pt}
\bibliography{references}
\endgroup
\clearpage
\appendix
\setcounter{topnumber}{4}
\setcounter{bottomnumber}{3}
\setcounter{totalnumber}{6}
\counterwithin{figure}{section}
\counterwithin{table}{section}
\renewcommand{\theHfigure}{appendix.\thesection.\arabic{figure}}
\renewcommand{\theHtable}{appendix.\thesection.\arabic{table}}
\section{Notation and Complete Theoretical Arguments}
\label{app:proofs}

This appendix proves conditional endpoint validity, iid Fresh calibration,
and deterministic completion equivalence. Table~\ref{tab:proof-map}
maps the arguments; the power boundary is an iid surrogate, and
Appendix~\ref{app:assumption-failures} separates assumption failures
from pathwise query costs.

\begin{table}[!htbp]
\centering\small
\caption{Theorem-to-proof map and scope of the supplementary arguments.}
\label{tab:proof-map}
\begin{tabularx}{\linewidth}{@{}p{.27\linewidth}p{.18\linewidth}X@{}}
\toprule
Result & Appendix & Main ingredient or qualification\\
\midrule
Theorem~\ref{thm:independent-pairs} & \ref{app:indep-proof} & Condition on discordance; binomial sign domination.\\
Theorem~\ref{thm:selected-winner} & \ref{app:selected-winner} & Rowwise-sum conditioning and Proposition~\ref{prop:product-covariance}; arbitrary within-item dependence.\\
Selection optimism and drag & \ref{sec:proof-drag} & Proposition~\ref{prop:selection-drag} and strict-null Remark~\ref{rem:drag-counterexample}.\\
Proposition~\ref{prop:bernstein} & \ref{app:calib-cert}, \ref{app:algo-calib} & Rational Bernstein bounds; Algorithm~\ref{alg:calibration} requires iid Fresh rows.\\
Proposition~\ref{prop:completion} & \ref{app:calib-cert}, \ref{app:stopping-order} & Attainable completions; action invariance and order-specific cost minima.\\
Section~\ref{sec:power-boundary} & \ref{app:surrogate} & Exact iid binomial sum; local-normal crossover and calibrated-cutoff correction.\\
Section~\ref{sec:stress} null boundary & \ref{app:surrogate} & Discrete conditional sizes and rational mixture enclosures; Table~\ref{tab:small-discordance}.\\
Conditional and stratified intervals & \ref{app:inference} & Finite-sample test-item bound; separate cohort-sampling approximation.\\
\bottomrule
\end{tabularx}
\end{table}

\begin{table}[!htbp]
\centering\small
\caption{Notation. Probabilities for potential models condition on training information $\mathcal M$; confidence statements additionally condition on frozen audit actions.}
\label{tab:notation}
\begin{tabular}{@{}p{0.27\linewidth}p{0.69\linewidth}@{}}
\toprule
Symbol & Meaning\\
\midrule
$K,j,J$ & Number of candidate parents, candidate index, and selected index.\\
$P_j,C_j;\ \mathcal M$ & Binary scores; training information fixing all potential models.\\
$U,D,Q,\mathcal C$ & Parent-training, child-training, calibration, and evaluation pools.\\
$S,F,T;\ Z_i$ & Selection, fresh-audit, and deployment-test samples; an evaluation row.\\
$B;\ n_S,n_F,n_H,n_{H,F}$ & Incremental budget; selection, Fresh, Hybrid, and Hybrid-fresh sizes.\\
$p_j=p_{P_j},p_{C_j},\Delta_j$ & Population accuracies and child-minus-parent gain.\\
$\widehat p,\widehat\Delta$ & Corresponding sample accuracies and gains.\\
$O_P,O_C,\delta_S$ & Parent/child selection optimism; positive conservative gain drag.\\
$\rho,\Psi,D(a)$ & Within-row candidate dependence, mechanism interaction, and overlap curve.\\
$A_i^\pi,\widehat\Gamma_{H,\pi}$ & Cohort upgrade action and empirical Hybrid-minus-policy utility contrast.\\
$X_i;\ b,c,M$ & Paired score difference; repair, regression, and discordance counts.\\
$q_+,q_-,d,\theta$ & Repair/regression probabilities; discordance and repair-sign rates.\\
$\alpha,\tau,c_n;\ A_\tau$ & Nominal level, endpoint and Fresh cutoffs; acceptance action.\\
$p(b,c),k_\tau(m),a_m$ & Full-tail $p$-value; critical repair count; conditional null size.\\
$R_n,\pi_n,\Delta_{\rm eff}$ & Unconditional null size, iid-surrogate power, and effective Hybrid mean.\\
$u_0,u_1,f;\ q$ & Unread selection-label counts and fresh pairs; calls used.\\
$Q_i,R_i,w(R)$ & Rowwise score sum, repair sign, and conditional winner probability.\\
$\Phi,z_\tau$ & Standard normal cdf and upper-tail critical value $\Phi^{-1}(1-\tau)$.\\
$N_\ell,n_{T,i},a_{\ell i}$ & Cohorts in contrast $\ell$, test rows in cohort $i$, and frozen action difference.\\
$g_\ell,\widehat g_\ell,L,\eta$ & Utility contrast, estimate, family size, and confidence error level.\\
$N_h,w_h,\bar Y_h,s_h^2$ & Stratum size, weight, mean utility difference, and variance.\\
$V_h,V,\nu$ & Stratum/total variance; Welch--Satterthwaite degrees of freedom.\\
\bottomrule
\end{tabular}
\end{table}

\subsection{Independent-pair validity}\label{sec:proof-independent}\label{app:indep-proof}
Condition on $\mathcal M$, which fixes all parents and potential children,
including independent training data and preassigned algorithmic randomness.
The horizons and lowest-index tie rule are fixed. A potential child can
depend on its parent and other training information, but its recipe cannot
adapt to the realized selection panel. Complete binary score rows are iid;
models within a row may be arbitrarily dependent.

For a fixed candidate, suppress $j$ and write
$q_+=\Pr(P=0,C=1)$, $q_-=\Pr(P=1,C=0)$,
$d=q_++q_-$ and $\Delta=q_+-q_-$. If $d>0$, let $\theta=q_+/d$.
When $d=0$, all paired outcomes agree and the gate never accepts.

\begin{proof}[Proof of Theorem~\ref{thm:independent-pairs}]
The repair, regression, and agreement counts are multinomial, so
$b\mid M=m\sim\operatorname{Bin}(m,\theta)$. More strongly, conditional
on the entire discordance pattern, the remaining repair indicators are
independent Bernoulli variables of parameter $\theta$. Define
\begin{equation}
 k_\tau(m)=\min\left\{b\in\{0,\ldots,m\}:
       2^{-m}\sum_{r=b}^{m}\binom{m}{r}\leq\tau\right\},
 \label{eq:critical-count}
\end{equation}
with $k_\tau(m)=m+1$ for an empty set. Under $\Delta\leq0$,
$\theta\leq1/2$. Coupling Bernoulli variables through common uniforms gives
\[
 \Pr\{b\geq k_\alpha(m)\mid M=m\}
 \leq\Pr\{\operatorname{Bin}(m,1/2)\geq k_\alpha(m)\}\leq\alpha.
\]
This includes $m=0$, for which rejection is impossible. Average over $M$.
Independent Fresh rows retain this law conditional on $J=j$.
\end{proof}

\subsection{Selected-winner validity}\label{sec:proof-selected}\label{app:selected-winner}

\begin{proposition}[Product-measure association]\label{prop:product-covariance}
If $R_1,\ldots,R_m$ are independent Bernoulli variables, $f$ is
coordinatewise increasing, and $w$ is coordinatewise decreasing, then
$\mathbb E[f(R)w(R)]\leq\mathbb E[f(R)]\mathbb E[w(R)]$.
\end{proposition}

\begin{proof}
For one Bernoulli variable of parameter $t$ and increasing $f,h$,
$\operatorname{Cov}(f(R),h(R))
=t(1-t)[f(1)-f(0)][h(1)-h(0)]\geq0$.
Induct on the coordinates, using the total-covariance identity:
the conditional covariance in the last coordinate is nonnegative,
and averaging that coordinate preserves monotonicity in the others.
Thus increasing functions are positively associated, the product-measure
case of the FKG inequality \citep{fkg1971}. Take $h=-w$.
Degenerate coordinates and $m=0$ are included.
\end{proof}

\begin{proof}[Proof of Theorem~\ref{thm:selected-winner}]
Fix $j$ with $\Delta_j\leq0$ and analyze its potential audit stream on all
$n_H$ rows, whether or not it wins. Condition on the complete vector
$Q_i=P_j(Z_i)+C_j(Z_i)\in\{0,1,2\}$. It fixes concordant outcomes and the
$m$ discordant positions. Conditional on $Q$, their repair indicators
$R_i=\ind\{P_j(Z_i)=0,C_j(Z_i)=1\}$ are independent
$\operatorname{Bernoulli}(\theta_j)$ variables with $\theta_j\leq1/2$:
conditioning factorizes row by row.

Let $w(r)=\Pr(J=j\mid Q,R=r)$. Fresh signs do not affect this probability.
For a discordant selection row and any competitor $k\ne j$, the parent
margin $V_{ik}=P_j(Z_i)-P_k(Z_i)$ satisfies
\[
 R_i=1\ \Longrightarrow\ V_{ik}\in\{-1,0\},\qquad
 R_i=0\ \Longrightarrow\ V_{ik}\in\{0,1\}.
\]
Every possible regression-row margin vector therefore dominates every
possible repair-row margin vector, even when competitors have different
conditional laws in the two cases. To compare configurations differing at
row $i$, couple every unchanged row identically; conditional row
independence permits this. At row $i$, any coupling of the two competitor
vectors respects the displayed support ordering. All total margins
decrease weakly on changing regression to repair. Since
\begin{equation}
 \{J=j\}=\bigcap_{k\ne j}
 \left\{\sum_{i\in S}V_{ik}\geq\ind\{k<j\}\right\},
 \label{eq:winner-margins}
\end{equation}
$w$ is decreasing in repair signs. Remove degenerate coordinates if
necessary; no conditional law at a zero-probability sign is needed.

At fixed $Q$, rejection is the increasing function
$f(R)=\ind\{\sum_iR_i\geq k_\alpha(m)\}$. Whenever
$\Pr(J=j\mid Q)>0$, Proposition~\ref{prop:product-covariance} yields
\[
 \Pr(A^H_\alpha=1\mid Q,J=j)
 =\frac{\mathbb E[f(R)w(R)\mid Q]}{\mathbb E[w(R)\mid Q]}
 \leq\mathbb E[f(R)\mid Q]\leq\alpha.
\]
Fresh repair signs are included; $w$ is constant in those coordinates.
Average over $Q\mid J=j$ and then sum over null indices:
\[
 \Pr(A^H_\alpha=1,\Delta_J\leq0\mid\mathcal M)
 \leq\alpha\sum_{j:\Delta_j\leq0}\Pr(J=j\mid\mathcal M).
\]
This proves both claims. Selection may change the distribution of $M$;
the argument works because it bounds every conditional sign tail.
\end{proof}

Association applies to the conditioned repair signs, so no
candidate-score independence is assumed. The separate roles of iid rows
and binary selection are examined in Appendix~\ref{app:assumption-failures}.

\subsection{Optimism and the scope of selection drag}\label{sec:proof-drag}

\begin{proposition}[Expected optimism and boundary-null drag]
\label{prop:selection-drag}
Under Theorem~\ref{thm:selected-winner}, each selectable $j$ satisfies
\[
 \mathbb E[\widehat p_{P_j}(S)\mid J=j,\mathcal M]\geq p_{P_j}.
\]
If $\Delta_j=0$, then
$\mathbb E[\widehat\Delta_j(S)\mid J=j,\mathcal M]\leq0$.
For an independent test sample $T$, expected parent optimism is therefore
nonnegative and expected reuse-minus-test gain is nonpositive at this
boundary null. The pathwise identity is
\[
 \widehat\Delta_j(S)-\widehat\Delta_j(T)
 =\{\widehat p_{C_j}(S)-\widehat p_{C_j}(T)\}
  -\{\widehat p_{P_j}(S)-\widehat p_{P_j}(T)\}.
\]
\end{proposition}

\begin{proof}
Before selection, candidate $j$'s parent scores are independent Bernoulli
variables. Changing one from zero to one changes its rowwise margins from
$\{-1,0\}$ to $\{0,1\}$. The preceding coupling shows that winner probability
is increasing in these scores. Positive association with their sample mean
proves parent optimism. At $\Delta_j=0$, condition on $Q$ on $S$.
If there are $m_S$ discordances, association with the decreasing winner
weight gives $\mathbb E[\sum R_i\mid Q,J=j]\leq m_S/2$.
The gain sum is $2\sum R_i-m_S$, so its expectation is nonpositive.
Independent test means remain unbiased conditional on $J=j$.
The final identity follows by expanding both gains.
\end{proof}

\begin{remark}[A strict-null counterexample]\label{rem:drag-counterexample}
Expected reuse-minus-population gain need not be negative away from the
boundary null.
Take two parents with lowest-index tie breaking, $C_1=P_1$, and iid rows
\[
\begin{array}{c|ccc}
 \text{probability}&P_1&P_2&C_2\\ \hline
 a&0&0&1\\ b&1&1&0\\ c&0&1&1
\end{array}
\qquad a,b,c>0,\quad a+b+c=1,\quad a<b.
\]
Here $\Delta_2=a-b<0$. The parent-score difference $P_2-P_1$ is one
only on the third row type and zero otherwise, so the tie rule gives
$\{J=2\}=\{H\geq1\}$ for $H\sim\operatorname{Bin}(n,c)$ and
$n=n_S$. Write $t=1-c$ and $w_n=1-t^n>0$. A specified repair or
regression row contributes to a winning sample only if at least one of
the other $n-1$ rows is of the third type. Consequently,
\begin{align}
 \mathbb E[(C_2-P_2)(Z_i)\ind\{J=2\}]
   &=(a-b)(1-t^{n-1}),\nonumber\\
 \mathbb E[\widehat\Delta_2(S)\mid J=2]
   &=(a-b)\frac{1-t^{n-1}}{w_n},\label{eq:strict-null-conditional-gain}\\
 \mathbb E[\widehat\Delta_2(S)\mid J=2]-\Delta_2
   &=(a-b)\frac{t^n-t^{n-1}}{w_n}
     =\frac{(b-a)c\,t^{n-1}}{w_n}>0.\label{eq:strict-null-positive-bias}
\end{align}
The conditional gain remains nonpositive, including zero when $n=1$.
The same calculation separates the two optimism terms:
\begin{equation}
 \mathbb E[O_P\mid J=2]=\frac{ac\,t^{n-1}}{w_n},\qquad
 \mathbb E[O_C\mid J=2]=\frac{bc\,t^{n-1}}{w_n}.
 \label{eq:strict-null-optimism}
\end{equation}
Thus the child's optimism exceeds the parent's even though the latter is
positive. The bias in Equation~\eqref{eq:strict-null-positive-bias} has
the opposite sign from the conservative-drag convention
$\delta_S=\Delta-\Delta_S$. Conditional sign domination still controls the
full-tail test: the sign probability on discordances is $a/(a+b)<1/2$,
and winning changes their prevalence rather than reversing this bound.
\end{remark}

Mean control is also insufficient by itself. For example, a five-row
stream that is all repairs with probability $1/10$ and all regressions
otherwise has mean gain $-4/5$, yet the full-tail gate at $1/20$ accepts
with probability $1/10$ because $p(5,0)=1/32$. This deliberately
dependent-row example is outside the theorem; it isolates why a mean
argument cannot replace its conditional tail domination.

\subsection{Rational calibration and exact completion}
\label{sec:proof-bernstein}\label{app:calib-cert}

\begin{proof}[Proof of Proposition~\ref{prop:bernstein}]
At fixed $d$, rejection probability is largest at $\theta=1/2$ among
$\Delta\leq0$. Under iid rows, $M\sim\operatorname{Bin}(n,d)$, giving
Equation~\eqref{eq:bernstein-size}. The coefficients $a_m(\tau)$ are rational
binomial sums, and rational cutoff comparisons use integer arithmetic.

For a degree-$n$ Bernstein polynomial with coefficients $a_i$, initialize
$a_i^{(0)}=a_i$ and subdivide at $t\in[0,1]$ by
\begin{equation}
 a_i^{(r+1)}=(1-t)a_i^{(r)}+t a_{i+1}^{(r)}
 \quad(0\leq i\leq n-r-1).
 \label{eq:de-casteljau}
\end{equation}
The left and right coefficient vectors are
$L_j=a_0^{(j)}$ and $H_j=a_j^{(n-j)}$.
For a direct verification, take $V\sim\operatorname{Bin}(n,u)$ and
$W\mid V=j\sim\operatorname{Bin}(j,t)$. Then
$W\sim\operatorname{Bin}(n,ut)$ and $\mathbb E[a_W\mid V=j]=L_j$,
so the left polynomial equals $R_n(tu;\tau)$.
Using $W=V+\operatorname{Bin}(n-V,t)$ instead gives
$W\sim\operatorname{Bin}(n,t+(1-t)u)$ and
$\mathbb E[a_W\mid V=j]=H_j$, proving the right formula.

The Bernstein basis is nonnegative and sums to one; a polynomial on each
subinterval lies between its minimum and maximum transformed coefficients.
Repeated $t=1/2$ subdivision preserves rational arithmetic.
A finite partition whose coefficient maxima are all at most $\alpha$
therefore certifies the stated supremum and, by sign domination, the
entire composite null.
\end{proof}

For supremum enclosures, let $U$ be the largest coefficient upper bound
over a covering partition and $L$ the largest exact evaluation at visited
rational points. Then $L\leq\sup_dR_n(d;\tau)\leq U$.
Bisecting intervals with the largest upper bounds refines the enclosure;
$U-L\leq10^{-10}$ certifies its width. This also applies to other fixed
rejection regions after replacing $a_m$ by their exact conditional sizes.
The null-boundary calculations restrict $d$ to $[0.005,0.40]$;
Fresh calibration covers $[0,1]$.

\label{sec:proof-completion}
\begin{proof}[Proof of Proposition~\ref{prop:completion}]
Let $Y_m\sim\operatorname{Bin}(m,1/2)$, where $m=b+c$.
Appending one fair Bernoulli variable gives
\begin{align*}
 p(b+1,c)&=\tfrac12\Pr(Y_m\geq b+1)+\tfrac12\Pr(Y_m\geq b)\leq p(b,c),\\
 p(b,c+1)&=\tfrac12\Pr(Y_m\geq b)+\tfrac12\Pr(Y_m\geq b-1)\geq p(b,c).
\end{align*}
Boundary tail conventions include $m=0$.
Every completion has
$b\leq b'\leq b+u_0+f$ and $c\leq c'\leq c+u_1+f$.
Monotonicity gives
$p(b+u_0+f,c)\leq p(b',c')\leq p(b,c+u_1+f)$.
Both extremes are attainable: for the minimum, make parent-zero items
and fresh pairs repairs and parent-one items ties; for the maximum,
make parent-one items and fresh pairs regressions and parent-zero items ties.

If the maximum is at most $\tau$, every completion accepts; if the minimum
exceeds $\tau$, every completion rejects. Otherwise the two extremes
produce different actions. Algorithm~\ref{alg:completion} thus stops at
the first decisive prefix in the specified query order, treating each
fresh pair atomically. At the full endpoint $u_0=u_1=f=0$, one condition
must hold. Its action agrees with the complete stream on every path,
hence acceptance probability and deployment utility also agree.
This deterministic equivalence inherits the endpoint's statistical validity.
\end{proof}

\subsection{Surrogate power, asymptotic crossover, and sparse discordances}
\label{sec:proof-surrogate}\label{sec:proof-crossover}\label{app:surrogate}

For an iid three-point law,
$\Pr(X=1)=(d+\Delta)/2$, $\Pr(X=-1)=(d-\Delta)/2$, and
$\Pr(X=0)=1-d$, where $|\Delta|\leq d\leq1$.
Conditioning on $M$ gives Equation~\eqref{eq:surrogate-power};
$d=0$ forces zero gain and zero acceptance probability.
If reused and fresh means are $\Delta-\delta_S$ and $\Delta$, their pooled
mean is exactly $\Delta_{\rm eff}=\Delta-(n_S/n_H)\delta_S$.
Approximating this selected, heterogeneous stream by a single iid law of
that mean and common $d$ is the surrogate step, defined where
$|\Delta_{\rm eff}|\leq d$.
Since deployment accuracy is $p_P+A\Delta$, fixed-gain utility is
$p_P+\Delta\Pr(A=1)$, proving Equation~\eqref{eq:surrogate-utility}.
For heterogeneous cohorts, the corresponding contrast is
$\mathbb E[(A_H-A_F)\Delta]$, retaining action--gain dependence.

\paragraph{The crossover slope.}
For fixed $d>0$ and weak gains, the iid central limit approximation is
\begin{equation}
 \pi_n(\Delta,d;\tau)
 \simeq \Phi\!\left(\frac{\sqrt n\,\Delta}{\sqrt d}-z_\tau\right),
 \qquad z_\tau=\Phi^{-1}(1-\tau).
 \label{eq:gaussian-power}
\end{equation}
It is a first-order local-alternative limit when $\Delta=O(n^{-1/2})$:
$\operatorname{Var}(X)=d-\Delta^2=d+o(1)$ and $M/n\to d$.
For the comparison, let the horizons grow at fixed positive ratios and
let both $\Delta$ and $\delta_S$ have this local scale. The fixed-$d$
assumption excludes sparse limits in which $nd$ remains bounded.
With a common critical value, equal powers require
$\sqrt{n_H}\{\Delta-(n_S/n_H)\delta_S\}=\sqrt{n_F}\Delta$.
Writing $r=B/n_S\geq1$ gives
\begin{align}
 \delta_S^*(\Delta)&\simeq
 \frac{n_H}{n_S}\left(1-\sqrt{\frac{n_F}{n_H}}\right)\Delta
 \label{eq:crossover-threshold}\\
 &=\frac{\Delta}{2\{1+\sqrt{r/(r+1)}\}}.
 \label{eq:crossover-slope}
\end{align}
The slope decreases from $1-1/\sqrt2$ at $r=1$ toward $1/4$.
For calibrated Fresh, the first-order critical-value correction is
\begin{equation}
 \delta_S^*(\Delta)\simeq\frac{n_H}{n_S}
 \left[\left(1-\sqrt{\frac{n_F}{n_H}}\right)\Delta
       -\sqrt{\frac d{n_H}}\{z_\alpha-z_{c_{n_F}}\}\right].
 \label{eq:crossover-calibrated}
\end{equation}
A larger Fresh cutoff lowers the boundary. These local-normal formulas
approximate the iid surrogate; the plotted finite-sample boundaries use
the exact sums. Saturation compresses utility differences when both powers approach one.

\paragraph{Why the normal rule peaks at small discordance.}
Given $M=m$ and signed gain sum $s=2b-m$, the paired normal statistic is
\[
 \widehat z=\frac{s\sqrt{n-1}}{\sqrt{nm-s^2}}.
\]
For positive $s$, the frozen rule rejects when $\widehat z\geq z_\alpha$,
equivalently $s\geq z_\alpha\sqrt{nm/(n-1+z_\alpha^2)}$, with positive
zero-variance streams accepted. Set $k_n^Z(0)=1$; for $m\geq1$ its
exact critical repair count is
\begin{equation}
 k_n^Z(m)=\min\!\left\{m+1,\left\lceil
 \frac{m+z_\alpha\sqrt{nm/(n-1+z_\alpha^2)}}{2}
 \right\rceil\right\}.
 \label{eq:naive-z-critical}
\end{equation}
The irregular conditional sizes in Table~\ref{tab:small-discordance}
are mixed by $M\sim\operatorname{Bin}(n,d)$. For any coefficient
sequence $a_m$, the mixture derivative is
\[
 \frac{\partial R_n(d)}{\partial d}
 =n\sum_{m=0}^{n-1}(a_{m+1}-a_m)
       \binom{n-1}{m}d^m(1-d)^{n-1-m}.
\]
Its extrema balance the rises and falls of the discrete conditional sizes.
At the certified normal-rule maxima, $nd$ is approximately $10.32$
for both horizons, locating $d$ near $0.040$ and $0.035$, respectively.
This explains the sparse-discordance peak; the rational supremum
calculation, rather than a Gaussian approximation, establishes its magnitude.

\paragraph{Explicit small-discordance comparison.}
\label{app:small-discordance}
Table~\ref{tab:small-discordance} gives exact conditional sizes at
$n=256$ and $\alpha=1/20$. At $m=3,4,9$, Naive-Z has sizes
$1/8$, $1/16$, and $46/512$. Mixing all conditional sizes, including
conservative regions, yields the approximately $0.20\pp$ breach in
Table~\ref{tab:null}. Rational cdf bounds certify
$1.644853<z_{1/20}<1.644854$; squared-statistic comparisons determine
all thresholds without rounding ambiguity.

\begin{table}[!htbp]
\centering\small
\caption{Exact conditional null thresholds at $n=256$, $\alpha=1/20$. The entry $k=m+1$ means no rejection. Fractions are exact; percentages are rounded. Naive-Z uses the sample-variance normal statistic and the frozen $z\geq z_{1/20}$ cutoff.}
\label{tab:small-discordance}
\begin{tabular}{@{}rrrrr@{}}
\toprule
$m$ & $k_{256}^Z(m)$ & $k_{1/20}(m)$ & $a_m^Z$ & $a_m(1/20)$\\
\midrule
1 & 2 & 2 & $0$ & $0$\\
2 & 3 & 3 & $0$ & $0$\\
3 & 3 & 4 & $1/8\ (12.50\%)$ & $0$\\
4 & 4 & 5 & $1/16\ (6.25\%)$ & $0$\\
5 & 5 & 5 & $1/32\ (3.12\%)$ & $1/32\ (3.12\%)$\\
6 & 6 & 6 & $1/64\ (1.56\%)$ & $1/64\ (1.56\%)$\\
7 & 6 & 7 & $8/128\ (6.25\%)$ & $1/128\ (0.78\%)$\\
8 & 7 & 7 & $9/256\ (3.52\%)$ & $9/256\ (3.52\%)$\\
9 & 7 & 8 & $46/512\ (8.98\%)$ & $10/512\ (1.95\%)$\\
10 & 8 & 9 & $56/1024\ (5.47\%)$ & $11/1024\ (1.07\%)$\\
12 & 9 & 10 & $299/4096\ (7.30\%)$ & $79/4096\ (1.93\%)$\\
15 & 11 & 12 & $1941/32768\ (5.92\%)$ & $576/32768\ (1.76\%)$\\
\bottomrule
\end{tabular}
\end{table}

\subsection{Conditional intervals and fixed-allocation sensitivity}
\label{app:inference}\label{sec:proof-inference}

\paragraph{Simultaneous conditional Hoeffding intervals.}
Let $\mathcal G$ fix every fitted model and frozen audit action.
For contrast $\ell$ over a cohort set $I_\ell$ of size $N_\ell$, define
$a_{\ell i}=A_i^H-A_i^{(\ell)}\in\{-1,0,1\}$ and
\[
 \widehat g_\ell=\frac1{N_\ell}\sum_{i\in I_\ell}
       \frac{a_{\ell i}}{n_{T,i}}\sum_{t=1}^{n_{T,i}}X_{it},
 \qquad
 g_\ell=\frac1{N_\ell}\sum_{i\in I_\ell}a_{\ell i}\Delta_i.
\]
Conditional on $\mathcal G$, the test draws are independent across
$i,t$, each $X_{it}\in[-1,1]$ has mean $\Delta_i$, and a weighted summand
has range width $2|a_{\ell i}|/(N_\ell n_{T,i})$.
Hoeffding's inequality \citep{hoeffding1963} therefore gives
\[
 \Pr\{|\widehat g_\ell-g_\ell|>\varepsilon\mid\mathcal G\}
 \leq2\exp\!\left\{-\frac{N_\ell^2\varepsilon^2}
                  {2\sum_{i\in I_\ell}a_{\ell i}^2/n_{T,i}}\right\}.
\]
For $L$ contrasts and family error $\eta$, set
\begin{equation}
 r_\ell=\sqrt{\frac{2\log(2L/\eta)}{N_\ell^2}
                  \sum_{i\in I_\ell}\frac{a_{\ell i}^2}{n_{T,i}}},
 \quad
 c_\ell=\frac1{N_\ell}\sum_{i\in I_\ell}|a_{\ell i}|,\quad
 I_\ell^{\rm H}=[\widehat g_\ell-r_\ell,\widehat g_\ell+r_\ell]
                    \cap[-c_\ell,c_\ell].
 \label{eq:conditional-hoeffding}
\end{equation}
A union bound gives
$\Pr\{\forall\ell:\ g_\ell\in I_\ell^{\rm H}\mid\mathcal G\}\geq1-\eta$;
contrasts need not be independent. When all action differences vanish,
the interval is exactly $\{0\}$. Here $n_{T,i}=2048$ and $\eta=0.05$.
SIGMORPHON-Mixed uses $L=12$ across pooled and family-specific comparisons;
Banking77-Mixed uses $L=4$ separately at each prespecified budget.
These intervals describe test-item uncertainty for the realized models
and actions.

\paragraph{Fixed-allocation stratified Welch inference.}
For a given comparison, let $Y_i=a_i\widehat\Delta_i(T)$ be its paired
cohort utility difference. Partition cohorts into fixed strata $h$ with
counts $N_h$, weights $w_h=N_h/N$, sample means $\bar Y_h$, and unbiased
sample variances $s_h^2$. Under independent cohort sampling within strata,
\begin{equation}
 \widehat g_{\rm strat}=\sum_h w_h\bar Y_h=\bar Y,\qquad
 V_h=\frac{w_h^2s_h^2}{N_h},\qquad
 V=\sum_hV_h,\qquad
 \nu=\frac{V^2}{\sum_hV_h^2/(N_h-1)}.
 \label{eq:welch-stratified}
\end{equation}
The variance formula follows by adding independent stratum-mean
variances; matching the first two moments of their estimated variance
gives the Welch--Satterthwaite degrees of freedom
\citep{satterthwaite1946,welch1947}. For $V>0$, write
$t=\widehat g_{\rm strat}/\sqrt V$. The approximate interval and unadjusted
$p$-values are
\begin{equation}
 \widehat g_{\rm strat}\pm t_{\nu,1-\eta/2}\sqrt V,\qquad
 p_+=1-F_{t_\nu}(t),\qquad p_2=2\{1-F_{t_\nu}(|t|)\}.
 \label{eq:welch-tests}
\end{equation}
Pooled SIGMORPHON-Mixed uses six family-by-recipe strata; each family
and Banking77-Mixed use three recipe strata with weights
$(1/2,1/4,1/4)$. Holm adjustment is applied separately to the specified
one- and two-sided families. An identically zero action contrast has
interval $\{0\}$ and $p=1$.

For comparison, the primary ordinary cohort procedure uses
$\bar Y\pm t_{N-1,1-\eta/2}s_Y/\sqrt N$. Fixed allocation motivates the
stratified sensitivity because random fluctuations in stratum proportions
are absent; the point estimate remains unchanged. Both $t$ procedures
are cohort-sampling approximations. Their target differs from the
conditional test-item target in Equation~\eqref{eq:conditional-hoeffding},
so their uncertainty radii are not added together.

\subsection{Exact calibration algorithm and the Fresh instantiation}
\label{app:calibration-algorithm}\label{app:algo-calib}

Algorithm~\ref{alg:calibration} searches a prespecified rational cutoff grid
and returns the largest candidate it can certify at the chosen subdivision
depth. A depth limit makes the search finite; failure to certify is not
proof that a cutoff is invalid. Every returned bound is exact.
The routine $\operatorname{SplitHalf}$ applies
Equation~\eqref{eq:de-casteljau} with $t=1/2$ and returns the two coefficient
vectors. Its midpoint is the midpoint of the current interval.

\begin{algorithm}[!htbp]
\caption{Rational Bernstein calibration of a Fresh cutoff}
\label{alg:calibration}
\small
\begin{algorithmic}[1]
\Require Horizon $n$, descending rational cutoff grid $\mathcal T$, level $\alpha$, depth limit $h_{\max}$
\For{$\tau\in\mathcal T$}
  \State Compute $a_m(\tau)=2^{-m}\sum_{r=k_\tau(m)}^m\binom mr$ exactly, $m=0,\ldots,n$
  \State $\mathcal Q\gets\{([0,1],a,0)\}$; $\mathcal L\gets\varnothing$; $\mathrm{unresolved}\gets\mathrm{false}$
  \While{$\mathcal Q\ne\varnothing$}
    \State Pop $(I,v,h)$ from $\mathcal Q$
    \If{$\max_jv_j\leq\alpha$}
      \State Add $(I,\max_jv_j)$ to $\mathcal L$
    \ElsIf{$h=h_{\max}$}
      \State $\mathrm{unresolved}\gets\mathrm{true}$; \textbf{break}
    \Else
      \State $(v_{\rm L},v_{\rm R})\gets\operatorname{SplitHalf}(v)$
      \State Add the two half-intervals with $(v_{\rm L},h+1)$ and $(v_{\rm R},h+1)$ to $\mathcal Q$
    \EndIf
  \EndWhile
  \If{not $\mathrm{unresolved}$}
    \State \Return $(c_n=\tau,\mathcal L)$
  \EndIf
\EndFor
\State \Return no cutoff certified on this grid at this depth
\end{algorithmic}
\end{algorithm}

For Fresh in Algorithm~\ref{alg:completion}, initialize
$u_0=u_1=0$, $f=n_F$, and use $\tau=\alpha$ for \FreshE\ or the certified
$c_{n_F}$ for \FreshU. Each fresh pair costs two endpoint evaluations.
Hybrid first queries children in the original $S$ order, then fresh pairs
in their original order. Both instantiations stop at the first
action-determining prefix within these atomic query units.

\subsection{Assumption boundaries and completion-certificate stopping geometry}
\label{app:assumption-failures}\label{app:stopping-order}

The sufficient assumptions of Theorem~\ref{thm:selected-winner} have
different functions: preassignment fixes a potential child before seeing
$S$, iid rows make the conditional signs independent, and binary parent
selection makes the winner weight decreasing in those signs. Removing
one of these conditions requires a new validity argument. It does not
logically imply that every method outside this class is invalid.

\begin{proposition}[A finite adaptive-child failure]\label{prop:adaptive-child-failure}
There is an iid binary evaluation problem with one fixed parent and two
fixed boundary-null children in which selecting the child with the higher
score on the reused panel makes the full-tail gate at $\alpha=1/20$
anticonservative. At panel size $19$, its exact false-upgrade probability
is $20\,531\,654/387\,420\,489>1/20$.
\end{proposition}

\begin{proof}
Let the item type be uniform on $\{a,b,h\}$, independently across rows,
and let each of the three models be correct on exactly one type:
\[
\begin{array}{c|ccc}
 \text{type}&P&C^{(1)}&C^{(2)}\\\hline
 a&0&1&0\\
 b&0&0&1\\
 h&1&0&0
\end{array}
\qquad
p_P=p_{C^{(1)}}=p_{C^{(2)}}=1/3.
\]
Both population gains are therefore zero. For a panel with type counts
$(a,b,h)$, choose $\widehat\ell=1$ if $a\geq b$ and $2$ otherwise.
The chosen child's repair count is $\max(a,b)$ and its regression count
is $h$. With $n_S=n_H=19$ and no fresh rows, the exact acceptance
probability is
\begin{align}
 \Pr(A_{\widehat\ell}=1)
 &=3^{-19}\!\sum_{a+b+h=19}
    \frac{19!}{a!b!h!}\,
    \ind\!\left\{20\sum_{r=\max(a,b)}^{\max(a,b)+h}
             \binom{\max(a,b)+h}{r}\leq2^{\max(a,b)+h}\right\}
    \nonumber\\
 &=\frac{61\,594\,962}{3^{19}}
  =\frac{20\,531\,654}{387\,420\,489}
  \approx0.0529958>0.05.
 \label{eq:adaptive-child-exact}
\end{align}
The sum has only $\binom{21}{2}=210$ integer triples. As a direct
arithmetic check, their accepting multinomial weights for
$h=0,1,2,3,4,5$ are respectively
$524\,288$, $4\,980\,736$, $22\,413\,312$, $28\,862\,634$,
$4\,465\,152$, and $348\,840$; all larger $h$ contribute zero.
Either preassigned child alone has acceptance probability
$3\,661\,393/129\,140\,163\approx0.0283521$.
Since the two children have the same regression count, selecting the
higher score accepts exactly when at least one fixed-child gate accepts;
the intersection probability is
$1\,436\,704/387\,420\,489$. These exact probabilities establish the
inflation rather than inferring it merely from the availability of two
null choices.
\end{proof}

Preassignment is sufficient for this theorem, not necessary for every
valid adaptive audit. Identical children, independent random choice,
valid multiplicity correction, or independent Fresh auditing can retain
control. In the last case, conditioning on child selection leaves the
fixed pair required by Theorem~\ref{thm:independent-pairs}, provided the
new rows remain iid from the same law.

\paragraph{A different selection metric changes the winner event.}
The binary implication
$R_i=1\Rightarrow P_j(Z_i)-P_k(Z_i)\in\{-1,0\}$ remains true under
any selection metric. What fails is the winner representation in
Equation~\eqref{eq:winner-margins}: margins $L_j-L_k$ of another
selection score need not decrease when regressions become repairs.

For an exact example, set $P_1=P_2=P$ and
$C_j=C^{(j)}$ in Proposition~\ref{prop:adaptive-child-failure}.
Use continuous-valued utilities $L_j(Z)=1+C_j(Z)+\varepsilon U$,
where $\varepsilon>0$ and $U\sim\operatorname{Uniform}(0,1)$ is
independent across rows and shared within each row. The common jitter
cancels, so selection reproduces Equation~\eqref{eq:adaptive-child-exact}
despite preassigned children. A type-$a$ repair has positive selection-score
margin while both binary parent margins are zero. This is an abstract
construction, not a generative claim about LLM log probabilities or a
claim that every metric mismatch causes inflation.

\paragraph{Which stochastic assumptions are doing work?}
A mean null under heterogeneous or dependent rows need not yield
independent, common-parameter signs after conditioning on discordance;
extensions require stronger conditional assumptions. Replacing the full
tail by a mid-$p$ value or normal approximation also changes the
conditional-size premise. Selected-winner validity bounds every
discordance pattern, whereas Fresh calibration uses their iid binomial
mixture. Completion equivalence is deterministic and needs neither
distributional argument.

\paragraph{Fixed streams, legal query units, and sharp prefix costs.}
Fix the complete target panel before acquisition. Reused items retain
their known parent labels and cost one child call each; fresh pairs have
neither score initially known and are acquired atomically at cost two.
Write $N_0,N_1$ for the parent-zero and parent-one counts in $S$,
$n_S=N_0+N_1$, and $N_F$ for the number of fresh pairs. A completed
stream is summarized by
\begin{equation}
 b_S,\ c_S,\ t_0=N_0-b_S,\ t_1=N_1-c_S;
 \qquad b_F,\ c_F,\ t_F=N_F-b_F-c_F.
 \label{eq:completion-stream-types}
\end{equation}
Here $t_0,t_1$ are reused $00$ and $11$ ties, and $t_F$ counts both
types of fresh tie. All counts are nonnegative integers. Let
$\mathcal O_{\rm all}$ allow arbitrary permutations of these labelled
atomic units, and let $\mathcal O_{S\to F}$ require every reused item
before any fresh pair, while allowing permutations within each part.
Algorithm~\ref{alg:completion} uses one fixed order in the latter class.

\begin{proposition}[Invariant actions and attainable cost lower bounds]
\label{prop:stopping-order}
For either order class, completion certification returns
$A_\tau(b_S+b_F,c_S+c_F)$ on every complete stream. The action is
invariant to any permitted permutation, but query cost can change.
For each class the formulas below give the exact minimum cost over its
permutations, with $+\infty$ assigned to the action opposite to the
endpoint. Thus they are sharp pathwise lower bounds for every allowed
order, and some allowed permutation attains each finite bound.
\end{proposition}

All following minimizations are over integers and an empty feasible set
has minimum $+\infty$. For an unrestricted labelled permutation, define
$\mathcal B(a,b,c,d)=\{(x,y,u,v)\in\mathbb Z_{\geq0}^4:
x\leq a,\ y\leq b,\ u\leq c,\ v\leq d\}$. The acceptance and rejection
minima are
\begin{align}
 L_A^{\rm all}
 &=\min_{\substack{(x,y,u,v)\in\mathcal B(b_S,t_1,b_F,t_F)\\
          p(x+u,N_1+N_F-y-u-v)\leq\tau}}
       \{x+y+2(u+v)\},\label{eq:minimum-accept-all}\\
 L_R^{\rm all}
 &=\min_{\substack{(x,y,u,v)\in\mathcal B(c_S,t_0,c_F,t_F)\\
          p(N_0+N_F-y-u-v,x+u)>\tau}}
       \{x+y+2(u+v)\}.\label{eq:minimum-reject-all}
\end{align}
For acceptance, $x,y$ count observed reused repairs and $11$ ties,
and $u,v$ count fresh repairs and ties. For rejection they instead count
reused regressions and $00$ ties, and fresh regressions and ties.
Thus the feasible sets encode both hidden outcomes and known parent
labels, and the objective retains the one-call/two-call distinction.

The restriction $S\to F$ changes these minima. A certificate obtained
within $S$ has optimal costs
\begin{align}
 L_{A,S}&=\min_{\substack{0\leq x\leq b_S,\ 0\leq y\leq t_1\\
                       p(x,N_1+N_F-y)\leq\tau}}(x+y),\nonumber\\
 L_{R,S}&=\min_{\substack{0\leq x\leq c_S,\ 0\leq y\leq t_0\\
                       p(N_0+N_F-y,x)>\tau}}(x+y).
 \label{eq:minimum-within-selection}
\end{align}
If acquisition reaches $F$, every reused item must already have been
queried, including those that do not help the certificate. The costs are
\begin{align}
 L_{A,F}&=n_S+2\min_{\substack{0\leq u\leq b_F,\ 0\leq v\leq t_F\\
                          p(b_S+u,c_S+N_F-u-v)\leq\tau}}(u+v),\nonumber\\
 L_{R,F}&=n_S+2\min_{\substack{0\leq u\leq c_F,\ 0\leq v\leq t_F\\
                          p(b_S+N_F-u-v,c_S+u)>\tau}}(u+v).
 \label{eq:minimum-within-fresh}
\end{align}
Consequently,
\begin{equation}
 L_A^{S\to F}=\min\{L_{A,S},L_{A,F}\},\qquad
 L_R^{S\to F}=\min\{L_{R,S},L_{R,F}\}.
 \label{eq:minimum-selection-first}
\end{equation}
These expressions also include $q=0$ certificates and the pure-Fresh
case $n_S=0$. The endpoint totals $(b,c)$ alone cannot determine the
minimum: at cutoff $1/20$, a stream of five reused repairs and two ties
needs at least five calls if both ties have parent label zero, but seven
if both have parent label one.

\begin{proof}[Proof of Proposition~\ref{prop:stopping-order}]
By Proposition~\ref{prop:completion}, every decisive prefix returns
the fixed endpoint action, and the full stream always decides. This
remains pathwise true under adaptive legal ordering when target items,
reuse status, horizon, and cutoff remain fixed.

To prove the minima, let a prefix contain $x_{S,+},x_{S,-}$ reused
repairs and regressions, $x_{S,0},x_{S,1}$ reused $00$ and $11$ ties,
and $x_{F,+},x_{F,-},x_{F,0}$ fresh repairs, regressions, and ties.
Substitution in the completion bounds gives the exact cancellations
\begin{align}
 p_{\max}
  &=p(x_{S,+}+x_{F,+},\,
       N_1+N_F-x_{S,1}-x_{F,+}-x_{F,0}),\nonumber\\
 p_{\min}
  &=p(N_0+N_F-x_{S,0}-x_{F,-}-x_{F,0},\,
       x_{S,-}+x_{F,-}).
 \label{eq:completion-prefix-cancellation}
\end{align}
Reading reused regressions, reused $00$ ties, or fresh regressions
leaves $p_{\max}$ unchanged. Under unrestricted permutations, deleting
these queries preserves acceptance and lowers cost, leaving exactly the
four types in Equation~\eqref{eq:minimum-accept-all}. For rejection,
delete reused repairs, reused $11$ ties, and fresh repairs to obtain
Equation~\eqref{eq:minimum-reject-all}. Hence every order is bounded
below by the corresponding minimum. Query the types of a minimizing
tuple first: they certify by its stated cost, and the lower bound
precludes an earlier stop, proving attainability.

Under $S\to F$, a decisive prefix is within $S$ or contains all of $S$
followed by part of $F$. The first case permits deletion of unhelpful
reused types, giving Equation~\eqref{eq:minimum-within-selection}.
In the second, all $n_S$ calls are compulsory and $b_S,c_S$ are fixed;
deleting only unhelpful fresh types gives
Equation~\eqref{eq:minimum-within-fresh}. Placing the selected useful
types first within the relevant part certifies by each candidate cost.
Choosing a minimizing case and applying the lower bound proves
attainability of Equation~\eqref{eq:minimum-selection-first}.
\end{proof}

These are oracle bounds for a completed stream: an online policy cannot
identify helpful hidden outcomes without querying them. A fixed order
stops at its first certificate; earlier prefixes admit both actions.
For eight reused repairs, two $11$ ties, and two $00$ ties, querying the
$11$ ties before five repairs accepts after seven calls. Querying the
$00$ ties, all eight repairs, then the $11$ ties requires eleven calls
for the same action. Separately stoppable single-model calls on fresh
pairs introduce additional information states and are outside these
atomic-pair minima.
\FloatBarrier
\section{Experimental Details and Calibration Protocol}
\label{app:protocol}

Table~\ref{tab:overview} organizes the five evidence pillars by task, model family, and inferential role. The four accuracy studies contain 224 training-and-audit cohorts, spanning historical beneficial updates and prospective mixtures of beneficial and degrading recipes. The protocol details below retain these distinct study settings.
\begin{table}[!htbp]
\centering\footnotesize
\setlength{\tabcolsep}{3pt}
\captionsetup{position=top}
\caption{Experimental overview of the five evidence pillars.}\label{tab:overview}
\begin{tabular}{@{}>{\raggedright\arraybackslash}p{.23\linewidth}>{\raggedright\arraybackslash}p{.10\linewidth}>{\raggedright\arraybackslash}p{.15\linewidth}>{\raggedright\arraybackslash}p{.17\linewidth}r>{\raggedright\arraybackslash}p{.16\linewidth}@{}}
\toprule
Study & Role & Task domain & Model family & \shortstack{Cohorts\\($N$)} & Budget $B$ / metric \\
\midrule
Synthetic Linear Systems & Mechanism & Linear-system prediction & Qwen3.6-35B-A3B & 32 & Overlap / nats \\
AG News & Mechanism & News classification & Qwen3.6-35B-A3B & 32 & Overlap / nats \\
AG News & Mechanism & News classification & Qwen3-8B & 80 & Replication / nats \\
\mbox{SIGMORPHON-Hist} & Historical & Morphological inflection & Qwen3-8B & 112 & $334$ / accuracy \\
\mbox{SIGMORPHON-Mixed} & Prospective & Morphological inflection & Qwen3-8B; Mistral-7B & 64 & $334$ / accuracy \\
Banking77-Mixed & Prospective & Intent classification & Qwen3-8B & 48 & $256,334$ / accuracy \\
\bottomrule
\end{tabular}
\par\vspace{4pt}\noindent\begin{minipage}{\linewidth}\footnotesize Hist pools 48 and 64 cohorts; Mixed comprises 40 Qwen and 24 Mistral cohorts. Accuracy is binary exact match.\end{minipage}
\end{table}

For cohort $i$ and policy $\pi$, let $A_i^\pi$ be its upgrade action. The deployment-accuracy contrast is
\begin{equation}
 \widehat\Gamma_{H,\pi}=\frac1N\sum_{i=1}^N (A_i^H-A_i^\pi)\widehat\Delta_i(T_i).
 \label{eq:utility}
\end{equation}

\subsection{Tasks, data partitions, and scoring}
\label{app:data}
\paragraph{Morphological inflection.}
The SIGMORPHON studies use the Finnish and Turkish part-1 datasets of SIGMORPHON--UniMorph 2023 \citep{goldman2023}. An input specifies a language, lemma, and morphological feature bundle; the output is the corresponding inflected form. In SIGMORPHON-Mixed, the parent-training pool $U$ contains 5,940 Finnish and 6,119 Turkish forms, and the update pool $D$ contains 4,060 and 3,881 forms. The calibration pool $Q$ and evaluation frame $\mathcal C$ each contain 1,000 forms per language. Lemmas are disjoint across $U,D,Q,\mathcal C$. Correctness requires a single nontruncated target string after Unicode NFC normalization and removal of outer whitespace; case, internal whitespace, and diacritics are retained.

\paragraph{Intent classification.}
Banking77 consists of English customer queries with 77 intent labels \citep{casanueva2020}. Identity normalization uses NFC, collapsed whitespace, and case folding; duplicate removal prioritizes the official test partition. The resulting training and test populations contain 9,992 and 3,079 items. Training items are split into $|U|=3{,}080$ and $|D|=6{,}912$; the test partition supplies $|Q|=1{,}079$ and $|\mathcal C|=2{,}000$. All four pools are disjoint. Class-balanced parent sampling reserves at least eight update examples per class: two small classes consequently have parent quotas of 27 and 33, while 55 classes contribute 40 and 20 contribute 41. The model returns one canonical intent string: lowercase ASCII with nonalphanumeric runs replaced by underscores.

\begin{table}[!htbp]
\centering\small
\caption{Task examples. The intent utterance is from the \href{https://github.com/PolyAI-LDN/task-specific-datasets/blob/master/banking_data/test.csv}{official Banking77 benchmark}.}\label{tab:task-examples}
\begin{tabular}{@{}p{.18\linewidth}p{.47\linewidth}p{.28\linewidth}@{}}
\toprule
Task & Input fields & Target / output vocabulary \\
\midrule
Finnish inflection & Lemma \texttt{-aistinen}; features \texttt{ADJ;ACC(PL)} & \texttt{-aistiset} \\
Turkish inflection & Lemma \texttt{alb\"um}; features \texttt{N;ABL(PL)} & \texttt{alb\"umlerden} \\
Banking intent & Utterance: ``My card has not arrived yet.'' & \texttt{card\_arrival} \\
\bottomrule
\end{tabular}
\end{table}

\paragraph{Mechanism tasks.}
Synthetic Linear Systems uses two-variable systems with coefficient magnitudes in $\{1,2,3\}$ and integer solutions in $[-9,9]$. AG News is four-class news classification \citep{zhang2015}; its prompt asks for the label 0 (World), 1 (Sports), 2 (Business), or 3 (Sci/Tech). Removing conflicting labels, training--test overlap, and same-label duplicates leaves 119,817 training articles; 1,024 are reserved for development, and 7,344 evaluation items remain after a separate 256-item reservation. Mechanism utility is mean semantic-answer-token log probability in nats, excluding EOS; it is distinct from the binary exact-match utility used in deployment studies.

\paragraph{Evaluation sampling.}
Each prospective cohort selects among three independently initialized parent runs on $S_{256}$. The main $S$, $F_{256}$, and $T_{2048}$ streams are independent uniform draws with replacement from the fixed frame $\mathcal C$; repeated items remain in the sample. Thus the iid sampling guarantee is conditional on the finite frame and fitted models. All potential-child training and order seeds are assigned before parent selection, with the lowest fixed index breaking ties. Only the selected parent's child is trained.

\subsection{Adaptation and decoding}
\label{app:training}
SIGMORPHON-Mixed uses Qwen3-8B and Mistral-7B-Instruct-v0.3; Banking77-Mixed uses the same Qwen3-8B revision \citep{qwen32025,mistral2023}. All prospective runs use rank-32 LoRA with scaling 64 and zero dropout \citep{hu2022lora}, bfloat16 base weights, and float32 adapters. Adapters target the query, key, value, attention-output, feed-forward gate/up/down, and language-model output projections. The effective batch is 32, implemented with microbatches of eight and gradient checkpointing. Every training operation starts a fresh AdamW optimizer with $(\beta_1,\beta_2)=(0.9,0.95)$, $\epsilon=10^{-12}$, zero weight decay, and no gradient clipping. The loss averages answer-plus-EOS token losses within examples and then across examples; prompt-token loss is zero.

Decoding uses deterministic greedy search with evaluation batches of 32, SDPA attention, and TF32 disabled. Inflection permits 64 new tokens; intent classification permits 20. The fixed prompt widths are 145 and 167 tokens for Qwen and Mistral inflection, and 134 for intent classification. EOS terminates generation. Malformed and truncated outputs remain in accuracy denominators. The software stack uses PyTorch 2.8.0, Transformers 4.57.1, PEFT 0.17.1, and CUDA 12.8.

\begin{table}[!htbp]
\centering\footnotesize
\setlength{\tabcolsep}{4pt}
\caption{Frozen prospective update recipes. The effective batch size is 32; support size and total training visits need not coincide.}\label{tab:recipes}
\begin{tabular}{@{}llrrrl@{}}
\toprule
Task / family & Recipe & Cohorts & Steps & Learning rate & Update support \\
\midrule
Inflection / Qwen & G1 & 20 & 128 & $10^{-4}$ & Balanced, 4,096 visits \\
 & G2a & 10 & 128 & $2\cdot10^{-4}$ & One language, 4,096 visits \\
 & G2b & 10 & 32 & $5\cdot10^{-5}$ & Repeated 64-example support \\
Inflection / Mistral & G1 & 12 & 128 & $10^{-4}$ & Balanced, 4,096 visits \\
 & G2a & 6 & 128 & $1.5\cdot10^{-4}$ & One language, 4,096 visits \\
 & G2b & 6 & 32 & $5\cdot10^{-5}$ & Repeated 64-example support \\
Banking77 / Qwen & G1 & 24 & 96 & $10^{-4}$ & All 77 classes, 3,072 visits \\
 & G2a & 12 & 96 & $2\cdot10^{-4}$ & Class-index half, 3,072 visits \\
 & G2b & 12 & 32 & $5\cdot10^{-5}$ & Eight examples from eight classes \\
\bottomrule
\end{tabular}
\end{table}

Parent training uses 2,048 assigned examples. The inflection parent runs take 128 optimizer steps; Banking77 parents take 64 steps at learning rate $10^{-4}$. Each model family allocates half its cohorts to balanced updates (G1), one quarter to support shift (G2a), and one quarter to repeated small-support training (G2b). Inflection G2a uses a shuffled complete language-specific update pool followed by a shuffled prefix to reach 4,096 visits; G2 recipes are balanced across languages. Banking77 G2a uses predefined halves of the class indices, while G2b repeats a 64-example support drawn from eight classes. Audit policies receive correctness outcomes rather than recipe labels.

\subsection{Independent calibration and outcome blinding}
\label{app:freeze}
SIGMORPHON-Mixed retains the three recipes qualified on an independent calibration pool before the main study; an additional recipe that failed calibration was excluded before the prospective design was fixed. Banking77-Mixed calibrates on $Q_{512}$ using two pilot parents with accuracies 73.63\% and 72.85\%. The first candidate dose for each recipe passed its calibration gate (Table~\ref{tab:phase0}). Recipe assignments, seeds, sample indices, audit cutoffs, and analysis families are fixed before evaluation. Primary policy actions are frozen before access to the main test outcomes for analysis; Banking77 freezes actions at every reported budget together. No main-test outcome is used for recipe choice, threshold tuning, cohort exclusion, or sample extension.

\begin{table}[!htbp]
\centering\small
\caption{Independent Banking77 calibration. Gains relative to each pilot parent are in pp.}\label{tab:phase0}
\begin{tabular}{@{}lrrrl@{}}
\toprule
Recipe & Pilot 1 & Pilot 2 & Mean & Frozen dose \\
\midrule
G1 & $+11.5234$ & $+13.2813$ & $+12.4023$ & 96 steps, $10^{-4}$ \\
G2a & $-27.1484$ & $-25.1953$ & $-26.1719$ & 96 steps, $2\cdot10^{-4}$ \\
G2b & $-51.5625$ & $-28.1250$ & $-39.8438$ & 32 steps, $5\cdot10^{-5}$ \\
\bottomrule
\end{tabular}
\end{table}

\subsection{Historical studies and compute}
\label{app:historical}
The 32-cohort mechanism experiments used Qwen3.6-35B-A3B with rank-32 adaptation. In each 32-cohort study, four parents shared a training multiset of 768 examples but differed in data order; each received 24 updates with batch 32 and learning rate $10^{-4}$. The endpoint update used one batch of 128 at $10^{-5}$. The later AG News experiment used Qwen3-8B in 80 fixed cohorts. SIGMORPHON-I and SIGMORPHON-II contain 48 and 64 distinct historical cohorts; the former retained 2,048 test pairs per cohort and the latter 4,096. Their protocols and generation configurations differ from the prospective implementation, so the pooled historical comparison retains study-specific variance in its sensitivity analysis. Cohort recovery did not create additional scientific replicates.

\begin{table}[!htbp]
\centering\small
\caption{Physical workload of the two prospective main studies.}\label{tab:physical-compute}
\begin{tabular}{@{}lrr@{}}
\toprule
Quantity & SIGMORPHON-Mixed & Banking77-Mixed \\
\midrule
Cohorts & 64 & 48 \\
Parent training runs & 192 & 144 \\
Selected-child training runs & 64 & 48 \\
Optimizer steps & 31,232 & 13,056 \\
Training-example visits & 999,424 & 417,792 \\
Physical endpoint evaluations & 360,448 & 270,336 \\
Test pairs & 131,072 & 98,304 \\
\bottomrule
\end{tabular}
\end{table}

Each main cohort acquires all three parent selection streams, the selected child stream, 256 fresh pairs, and 2,048 test pairs, totaling 5,632 endpoint evaluations. The Banking77 pilot additionally uses eight training operations, 576 optimizer steps, and 4,096 endpoints. The numerical supplement supplies fixed outcome arrays and assignments sufficient to reproduce policy comparisons, curves, and stopping costs. Full rendered prompts, adapters, and raw responses are not included in that numerical release; Table~\ref{tab:task-examples} illustrates benchmark inputs and targets rather than full chat templates.
\subsection{Formal policy definitions and evaluation contracts}
\label{app:policy-specification}
The comparison separates an audit's statistical rule, its available information, and its acquisition cost. Throughout, $\alpha=1/20$, $p(b,c)$ is the full binomial right-tail in Definition~\ref{def:endpoint}, and $b,c$ count repairs and regressions on the policy's own panel. For an incremental allowance $B\ge n_S$, define $n_{HF}=\lfloor(B-n_S)/2\rfloor$ and $n_F=\lfloor B/2\rfloor$. A reused item reveals only the child's outcome at cost one; a fresh item is acquired as an atomic parent--child pair at cost two. The recorded parent-selection evaluations are sunk. Table~\ref{tab:policy-specification} specifies all nine policy families under this endpoint accounting; historical monetary-budget variants are distinguished below.

\begin{table}[!htbp]
\centering\footnotesize
\setlength{\tabcolsep}{3pt}
\renewcommand{\arraystretch}{1.12}
\caption{Audit policies, evaluation panels, and validity contracts. Costs count incremental model--item evaluations.}\label{tab:policy-specification}
\begin{tabular}{@{}>{\raggedright\arraybackslash}p{.18\linewidth}>{\raggedright\arraybackslash}p{.27\linewidth}>{\raggedright\arraybackslash}p{.18\linewidth}>{\raggedright\arraybackslash}p{.12\linewidth}>{\raggedright\arraybackslash}p{.18\linewidth}@{}}
\toprule
Policy & Acceptance rule & Panel / full cost & Acquisition & Null-risk contract \\
\midrule
\Hybrid & $p(b,c)\le\alpha$ & $S$ and $n_{HF}$ fresh pairs; $n_S+2n_{HF}$ & Exact completion & Selected-winner level $\alpha$ (Theorem~\ref{thm:selected-winner}). \\
\addlinespace[3pt]
\textsc{Fresh}\newline\textsc{Unconditional} & $p(b,c)\le c_{n_F}$, with certified cutoff & First $n_F$ fresh pairs; $2n_F$ & Exact completion & iid Fresh level $\alpha$; horizon-specific certificate. \\
\addlinespace[3pt]
\FreshE & $p(b,c)\le\alpha$ & First $n_F$ fresh pairs; $2n_F$ & Exact completion & iid Fresh level $\alpha$ conditional on discordance. \\
\addlinespace[3pt]
Naive-Mean & $b>c$ & Same panel as Hybrid; $n_S+2n_{HF}$ & Full panel & No nominal-level guarantee. \\
\addlinespace[3pt]
Naive-Z & $Z\ge\Phi^{-1}(0.95)$; Eq.~\eqref{eq:policy-z} & Same panel as Hybrid; $n_S+2n_{HF}$ & Full panel & Normal approximation; size may exceed $\alpha$. \\
\addlinespace[3pt]
AlwaysUpdate & Always deploy the child & None; zero & No audit & A null child is always accepted. \\
\addlinespace[3pt]
NoUpdate & Always retain the parent & None; zero & No audit & Zero false upgrades and zero power. \\
\addlinespace[3pt]
FreshExact256 & $p(b,c)\le\alpha$ & All $256$ fresh pairs; $512$ & Full panel & iid Fresh level $\alpha$; exceeds any $B<512$. \\
\addlinespace[3pt]
FreshSeq & First $t$ with $E_t\ge 1/\alpha=20$; Eq.~\eqref{eq:policy-freshseq} & Fresh until crossing or cap; $2t$ in the endpoint frontier & Sequential crossing & Anytime level $\alpha$ under the iid Fresh betting contract. \\
\bottomrule
\end{tabular}
\end{table}

For the paired normal rule, write $D=b-c$, $m=b+c$, and let $n$ include both concordant and discordant pairs. The implemented statistic uses the sample variance with denominator $n-1$:
\begin{equation}
 Z=\frac{\overline X}{s_X/\sqrt n}
   =\frac{D\sqrt{n-1}}{\sqrt{nm-D^2}},\qquad
 s_X^2=\frac{m-D^2/n}{n-1}.
 \label{eq:policy-z}
\end{equation}
When $s_X=0$, this rule accepts exactly when $\overline X>0$. Naive-Mean and Naive-Z use the complete assigned panel: their reported costs do not include the stopping mechanism used for the exact endpoint gates. The strengthened Fresh cutoff is $57/1000$ for the primary $128$- and $167$-pair comparisons; the other frozen prospective cutoffs and their interval certificates appear in Table~\ref{tab:fresh-certificates}. The separately tabulated offline horizon calibrations support descriptive budget analysis and do not replace the frozen prospective rules.

The historical FreshSeq comparator places half its betting mass at $\lambda=1$ and half uniformly on $(0,1)$. With fresh gains $X_i\in\{-1,0,1\}$, its statistic is
\begin{equation}
 E_t=\frac12\prod_{i=1}^{t}(1+X_i)
       +\frac12\int_0^1\prod_{i=1}^{t}(1+\lambda X_i)\,d\lambda.
 \label{eq:policy-freshseq}
\end{equation}
Conditional on the fitted pair, iid fresh gains with nonpositive mean make every integrand a nonnegative supermartingale starting at one. Integration preserves this property, so the probability of ever crossing $20$ is at most $1/20$. FreshSeq retains the parent if no crossing occurs before its legal cap; its query count is the number of pairs consumed on either outcome. The historical monetary implementation reserves each pair's worst-case cost before observing its task or outputs, whereas the endpoint frontier caps the number of two-call pairs. Unlike completion of a fixed endpoint gate, a sequential crossing is part of the decision definition and can change when the same outcomes are reordered.

The historical fixed-fresh implementation chooses $N_b=\lfloor B_b/(\overline c_P+\overline c_{C,r})\rfloor$, using mean endpoint costs calibrated before selection and the realized cost $B_b$ of the complete child-on-selection panel. It completes all $N_b$ pairs even if realized spending differs from $B_b$: matching uses calibrated mean costs. Fixing the horizon before any fresh response preserves conditional fresh validity. The later 167-pair comparator instead has historical admission constraints; an unresolved monetary cap triggers retention. This conservative fallback differs from exact endpoint-action equivalence. Endpoint-budget replays make the complete target affordable.

All prospective completion costs use the original occurrence order on $S$, followed by the original fresh-stream order; a fresh comparator uses only its own fresh prefix. Outcomes already acquired for another policy remain hidden until the evaluated policy would have paid for them. Thus reported savings describe virtual policy execution on complete recorded streams. They do not reduce the physical workload needed to acquire all streams for the controlled comparison.
\subsection{Overlap experiments and the frozen covariance prediction}
\label{app:mechanism-construction}
The mechanism studies vary selection--audit dependence while holding trained models fixed. Each $32$-cohort study fits four parents per cohort on $768$ examples and two potential updates per parent on separate $128$-example datasets. Replays select on $32$ log-utility scores and apply a positive-mean-gain gate on $32$ audit items. Overlap is $\lambda\in\{0,1/4,1/2,3/4,1\}$, labelled $\alpha$ in the source figures. The recorded selection and reference panels $S,V$ each contain $256$ outcomes. Nested prefixes of independently permuted $32$-item panels supply $32\lambda$ reused draws and $32(1-\lambda)$ reference draws. Selection chooses the best of four parents, averaging exact ties uniformly; the unselected control averages all four. The $16{,}384$ replays also average equally over both update releases. This mechanism tie convention differs from the fixed lowest-index rule used for binary certification.

For a parent index $k$ and update release $r\in\{0,1\}$, let $L_k$ denote the parent's semantic-answer log utility and let $G_{kr}$ denote the child-minus-parent log-utility gain. The prediction model fits the mean and sample covariance of
\[
 W=(L_1,\ldots,L_4,G_{10},\ldots,G_{40},G_{11},\ldots,G_{41})
\]
on a separate $1{,}024$-item $F$ panel, using covariance denominator $1{,}023$. These moments are frozen before evaluating overlap curves. Four selection and four reference blocks, each representing eight draws, have independent Gaussian block means with covariance $\widehat\Sigma/8$. Shared selection blocks create overlap while retaining cross-candidate and cross-release covariance. The Gaussian prediction uses $131{,}072$ Gaussian draws with common inner random numbers across overlap settings.

Within this Gaussian construction, the shared selection blocks give
\[
 \operatorname{Cov}(\overline L_S,\overline G_{A_\lambda})
 =\frac{\lambda}{32}\widehat\Sigma_{LG}.
\]
Overlap therefore scales selection--gain covariance at fixed marginal panel sizes.

To isolate the contribution of selection--gain covariance, write $J_K$ for either the uniform-parent selector ($K=1$) or the maximum-score selector ($K=4$), and let $A_{\lambda,J_Kr}$ indicate a positive audit-panel mean gain. The update contribution to predicted utility is
\begin{align}
 Q_{\lambda,K}
 &=\frac12\sum_{r=0}^1
   \E_{\widehat\mu,\widehat\Sigma}
   [\widehat\mu_{G,J_Kr}\,A_{\lambda,J_Kr}],\nonumber\\
 D(\lambda)
 &=(Q_{\lambda,4}-Q_{\lambda,1})-(Q_{0,4}-Q_{0,1}).
 \label{eq:mechanism-construction}
\end{align}
The fixed parent-utility contribution cancels. Weighting acceptance by the moment-fitting panel's mean gain targets deployment utility. The observed interaction compares the same selectors and overlaps, with $\Psi=-D(1)$ under the plotted convention.

The covariance ablation replaces only the selection--gain cross block:
\[
 \widehat\Sigma^{(0)}=
 \begin{pmatrix}
   \widehat\Sigma_{LL}&0\\
   0&\widehat\Sigma_{GG}
 \end{pmatrix}.
\]
It retains means and both within-block covariances. Gaussian independence of selection-score and gain fluctuations then gives $D^{(0)}(\lambda)=0$: overlap cannot change the selector interaction. The zero ablation curve is exact, without Monte Carlo noise; neither candidate independence nor zero update gain is assumed.

Intervals are pointwise cohort $t_{31}$ intervals. Interior-overlap prediction errors are descriptive paired comparisons, without simultaneous coverage across tasks or overlaps. These log-utility studies isolate a covariance mechanism in nats; their Gaussian predictions neither replace the finite-sample binary proof nor give exact selected-stream power.
\FloatBarrier
\section{Supplementary Empirical Results and Diagnostics}
\label{app:tables}
\subsection{Mechanism and historical comparisons}
\label{app:historical-tables}
Table~\ref{tab:mechanism}, Panel~A, separates the log-probability mechanisms from accuracy-based selection effects. Across the three synthetic/32-cohort mechanism studies, the selected parent starts ahead but the additional unconditional update gain is lower; the remaining child advantage is the sum of these two components. The larger AG News study identifies optimism and the selection-by-audit interaction without identifying the full decomposition. Panel~B retains the complete cross-study selection audit, including pooled, family, and G1 summaries. Table~\ref{tab:merged} gives both historical study contrasts and their equal-cohort-weight synthesis. For SIGMORPHON-Hist, the fixed 48:64 mixture sensitivity gives a two-sided Welch $p=0.00523$.
\begin{table}[!htbp]
\centering
\captionsetup{position=top}
\caption{Mechanism decompositions and the complete cross-study selection audit. Panel A retains the original scales; accuracy entries are fractions, not pp. In Panel B, gain, parent/child optimism, and reuse-minus-test drag are in pp.}
\label{tab:mechanism}\label{tab:drag}
\begingroup
\centering\scriptsize
\setlength{\tabcolsep}{2.3pt}
\noindent\textbf{Panel A. Mechanism decompositions}\par\smallskip
\begin{tabular}{@{}lrlrrrrrrr@{}}
\toprule
Study & $N$ & Unit & $O$ & $B$ & $\Psi$ & $H$ & $\Delta_{\rm init}$ & $\Delta_{\rm gain}$ & $\Delta_{\rm final}$ \\
\midrule
Synthetic pilot & 16 & nats & -- & $-0.004413$ & 0.000748 & $-0.000951$ & 0.019284 & $-0.000951$ & 0.018333 \\
Synthetic Linear Systems & 32 & nats & -- & $-0.005073$ & 0.001059 & $-0.001523$ & 0.017917 & $-0.001523$ & 0.016394 \\
AG News & 32 & nats & -- & $-0.004626$ & 0.001024 & $-0.001522$ & 0.005722 & $-0.001522$ & 0.004200 \\
AG News & 80 & nats & 0.039017 & -- & 0.001546 & -- & -- & -- & -- \\
SIGMORPHON-II & 64 & accuracy & 0.013004 & $-0.010658$ & -- & $-0.004898$ & 0.006149 & $-0.004898$ & 0.001251 \\
\bottomrule
\end{tabular}
\par\vspace{4pt}\noindent\begin{minipage}{\linewidth}\footnotesize $B$ is reused-minus-test gain; $H$ is the selected-minus-unselected unconditional update-gain contrast. The synthetic and 32-cohort AG News $\Psi$ is the selection-by-audit interaction. The 80-cohort AG News $O$ contrasts selected and uniformly chosen parents on paired score panels; unavailable decompositions are left blank.\end{minipage}
\par
\endgroup
\par\medskip
\begingroup
\centering\small
\setlength{\tabcolsep}{3.5pt}
\noindent\textbf{Panel B. Cross-study selection audit}\par\smallskip
\begin{tabular}{@{}lrrrrrrr@{}}
\toprule
Study / subset & $N$ & $\widehat\Delta_T$ & $O_P$ (SE) & $p_2$ & $O_C$ & Drag & $d_S$ \\
\midrule
SIGMORPHON-I & 48 & 9.15 & 1.36 (0.41) & $0.0016$ & 0.22 & $-1.14$ & 0.212 \\
SIGMORPHON-II & 64 & 9.07 & 1.30 (0.36) & $7.03\!\times\!10^{-4}$ & 0.23 & $-1.07$ & 0.210 \\
Mixed: Qwen & 40 & 3.39 & 1.51 (0.40) & $5.31\!\times\!10^{-4}$ & $-0.07$ & $-1.59$ & 0.214 \\
Mixed: Qwen G1 & 20 & 9.83 & 1.86 (0.57) & $0.0044$ & 0.25 & $-1.61$ & 0.218 \\
Mixed: Mistral & 24 & 1.26 & 3.27 (0.62) & $2.44\!\times\!10^{-5}$ & 1.52 & $-1.75$ & 0.303 \\
Mixed: Mistral G1 & 12 & 11.49 & 2.14 (0.91) & $0.0390$ & 1.03 & $-1.10$ & 0.291 \\
Mixed: pooled & 64 & 2.59 & 2.17 (0.36) & $6.96\!\times\!10^{-8}$ & 0.52 & $-1.65$ & 0.248 \\
Mixed: pooled G1 & 32 & 10.45 & 1.96 (0.49) & $3.40\!\times\!10^{-4}$ & 0.54 & $-1.42$ & 0.245 \\
Banking77-Mixed & 48 & $-11.43$ & 1.69 (0.36) & $2.47\!\times\!10^{-5}$ & 0.53 & $-1.16$ & 0.339 \\
Banking77 G1 & 24 & 11.74 & 1.53 (0.47) & $0.0034$ & 0.68 & $-0.85$ & 0.197 \\
\bottomrule
\end{tabular}
\par\vspace{4pt}\noindent\begin{minipage}{\linewidth}\footnotesize Unadjusted paired $t$ tests assess parent optimism; Mixed is SIGMORPHON-Mixed and G1 is the balanced recipe.\end{minipage}
\par
\endgroup
\end{table}
\begin{table}[!htbp]
\centering\footnotesize
\setlength{\tabcolsep}{3pt}
\caption{Retrospective historical synthesis at $B=334$: Hybrid minus the strengthened Fresh comparator, in pp.}\label{tab:merged}
\begin{tabular}{@{}lrrrrr@{}}
\toprule
Study & $N$ & $\widehat\Gamma$ & Cohort 95\% CI & Conditional 95\% CI & $p_2$ \\
\midrule
SIGMORPHON-I & 48 & 1.5727 & $[0.3652, 2.7802]$ & $[1.1395, 2.0058]$ & $0.0118$ \\
SIGMORPHON-II & 64 & 0.6207 & $[-0.2642, 1.5055]$ & $[0.4007, 0.8406]$ & $0.1659$ \\
SIGMORPHON-Hist & 112 & 1.0287 & $[0.3115, 1.7458]$ & $[0.8045, 1.2528]$ & $0.0053$ \\
\bottomrule
\end{tabular}
\par\vspace{4pt}\noindent\begin{minipage}{\linewidth}\footnotesize The comparator was strengthened after historical outcomes. Conditional intervals are nominal fixed-rule sensitivities.\end{minipage}
\end{table}

\subsection{Complete prospective contrasts and policies}
\label{app:precision}
Tables~\ref{tab:mixed} and~\ref{tab:bank} give all twelve SIGMORPHON-Mixed and all eight Banking77-Mixed Hybrid-minus-comparator contrasts. Table~\ref{tab:precision-all} complements these contrasts with higher-precision effects and the two additional inferential views derived in Appendix~\ref{app:proofs}; Panels~A and~B report SIGMORPHON-Mixed and Banking77-Mixed, respectively. Tables~\ref{tab:mixedpolicies} and~\ref{tab:bankpolicies} list all eight SIGMORPHON-Mixed and all seven Banking77-Mixed policies. Naive-Mean and Naive-Z can attain higher realized utility by accepting more updates, while AU and NU use no audit queries.
\begin{table}[!htbp]
\centering\small
\setlength{\tabcolsep}{4.5pt}
\captionsetup{position=top}
\caption{SIGMORPHON-Mixed: twelve Hybrid-minus-comparator accuracy contrasts at $B=334$. Effects and ordinary cohort 95\% confidence intervals are in pp.}\label{tab:mixed}
\begin{tabular}{@{}llrrrr@{}}
\toprule
Population & Comparator & $\widehat\Gamma$ & Cohort 95\% CI & $p_+$ (H12) & $p_2$ (H12) \\
\midrule
Pooled & AU & 1.92 & $[0.52, 3.32]$ & $0.0353$ & $0.0707$ \\
Pooled & NU & 4.51 & $[3.09, 5.94]$ & $1.83\!\times\!10^{-7}$ & $3.66\!\times\!10^{-7}$ \\
Pooled & Fresh-U & $-0.22$ & $[-0.72, 0.27]$ & $1$ & $1$ \\
Pooled & Fresh-E & $-0.07$ & $[-0.65, 0.51]$ & $1$ & $1$ \\
\addlinespace[2pt]
Qwen & AU & 0.73 & $[-0.60, 2.05]$ & $0.9660$ & $1$ \\
Qwen & NU & 4.11 & $[2.46, 5.76]$ & $6.38\!\times\!10^{-5}$ & $1.28\!\times\!10^{-4}$ \\
Qwen & Fresh-U & $-0.36$ & $[-1.16, 0.44]$ & $1$ & $1$ \\
Qwen & Fresh-E & $-0.12$ & $[-1.06, 0.83]$ & $1$ & $1$ \\
\addlinespace[2pt]
Mistral & AU & 3.92 & $[0.93, 6.91]$ & $0.0497$ & $0.0993$ \\
Mistral & NU & 5.18 & $[2.40, 7.96]$ & $0.0040$ & $0.0080$ \\
Mistral & Fresh-U & 0.00 & $[0.00, 0.00]$ & $1$ & $1$ \\
Mistral & Fresh-E & 0.00 & $[0.00, 0.00]$ & $1$ & $1$ \\
\bottomrule
\end{tabular}
\par\vspace{4pt}\noindent\begin{minipage}{\linewidth}\footnotesize H12 adjusts each direction over all twelve contrasts. AU/NU denote AlwaysUpdate/NoUpdate; Fresh-U/E denote FreshUnconditional/Exact. Pooled H$-$AU passes only the one-sided family. Table~\ref{tab:mixedprecision} gives sensitivity estimates.\end{minipage}
\end{table}
\begin{table}[!htbp]
\centering\small
\setlength{\tabcolsep}{5pt}
\captionsetup{position=top}
\caption{Banking77-Mixed deployment contrasts. Effects and ordinary cohort 95\% intervals are in pp.}\label{tab:bank}
\begin{tabular}{@{}rlrrrr@{}}
\toprule
$B$ & Comparator & $\widehat\Gamma$ & Cohort 95\% CI & $p_+$ (H4) & $p_2$ (H4) \\
\midrule
256 & AU & 16.80 & $[11.21, 22.40]$ & $3.52\!\times\!10^{-7}$ & $7.03\!\times\!10^{-7}$ \\
256 & NU & 5.37 & $[3.62, 7.13]$ & $3.10\!\times\!10^{-7}$ & $6.20\!\times\!10^{-7}$ \\
256 & Fresh-U & 0.23 & $[-0.90, 1.36]$ & $0.6877$ & $1$ \\
256 & Fresh-E & 0.23 & $[-0.90, 1.36]$ & $0.6877$ & $1$ \\
\addlinespace[2pt]
334 & AU & 17.11 & $[11.63, 22.59]$ & $1.53\!\times\!10^{-7}$ & $3.05\!\times\!10^{-7}$ \\
334 & NU & 5.68 & $[3.89, 7.46]$ & $1.28\!\times\!10^{-7}$ & $2.57\!\times\!10^{-7}$ \\
334 & Fresh-U & 0.05 & $[-0.58, 0.68]$ & $0.8689$ & $1$ \\
334 & Fresh-E & 0.05 & $[-0.58, 0.68]$ & $0.8689$ & $1$ \\
\bottomrule
\end{tabular}
\par\vspace{4pt}\noindent\begin{minipage}{\linewidth}\footnotesize $B=256$ and $B=334$ use separate Holm-four families; neither establishes Hybrid superiority over Fresh.\end{minipage}
\end{table}
\begin{table}[!htbp]
\centering
\captionsetup{position=top}
\caption{Higher-precision effects, ordinary cohort intervals, fixed-allocation Welch sensitivity, and simultaneous conditional Hoeffding intervals. Panel A reports SIGMORPHON-Mixed; Panel B reports Banking77-Mixed. Effects and intervals are in pp.}
\label{tab:precision-all}\label{tab:mixedprecision}\label{tab:bankprecision}
\begingroup
\centering\footnotesize
\setlength{\tabcolsep}{3pt}
\noindent\textbf{Panel A. SIGMORPHON-Mixed}\par\smallskip
\begin{tabular}{@{}llrrrr@{}}
\toprule
Population & Comparator & $\widehat\Gamma$ & Cohort 95\% CI & Welch $p_2$ H12 & Conditional 95\% CI \\
\midrule
Pooled & AU & 1.9241 & $[0.5240, 3.3243]$ & $0.0023$ & $[1.1861, 2.6621]$ \\
Pooled & NU & 4.5128 & $[3.0851, 5.9405]$ & $5.85\!\times\!10^{-10}$ & $[3.8824, 5.1432]$ \\
Pooled & Fresh-U & $-0.2243$ & $[-0.7167, 0.2681]$ & $1$ & $[-0.4956, 0.0470]$ \\
Pooled & Fresh-E & $-0.0732$ & $[-0.6545, 0.5080]$ & $1$ & $[-0.3704, 0.2239]$ \\
Qwen & AU & 0.7251 & $[-0.6024, 2.0526]$ & $1$ & $[-0.2059, 1.6561]$ \\
Qwen & NU & 4.1101 & $[2.4555, 5.7647]$ & $1.56\!\times\!10^{-6}$ & $[3.3097, 4.9105]$ \\
Qwen & Fresh-U & $-0.3589$ & $[-1.1570, 0.4392]$ & $1$ & $[-0.7929, 0.0752]$ \\
Qwen & Fresh-E & $-0.1172$ & $[-1.0628, 0.8284]$ & $1$ & $[-0.5927, 0.3583]$ \\
Mistral & AU & 3.9225 & $[0.9313, 6.9138]$ & $0.0096$ & $[2.7120, 5.1331]$ \\
Mistral & NU & 5.1839 & $[2.4039, 7.9640]$ & $5.45\!\times\!10^{-4}$ & $[4.1608, 6.2070]$ \\
Mistral & Fresh-U & 0.0000 & $[0.0000, 0.0000]$ & $1$ & $[0.0000, 0.0000]$ \\
Mistral & Fresh-E & 0.0000 & $[0.0000, 0.0000]$ & $1$ & $[0.0000, 0.0000]$ \\
\bottomrule
\end{tabular}
\par\vspace{4pt}\noindent\begin{minipage}{\linewidth}\footnotesize Welch is a sensitivity analysis; conditional intervals quantify independent test sampling given fitted models and frozen actions.\end{minipage}
\par
\endgroup
\par\medskip
\begingroup
\centering\footnotesize
\setlength{\tabcolsep}{3pt}
\noindent\textbf{Panel B. Banking77-Mixed}\par\smallskip
\begin{tabular}{@{}rlrrrr@{}}
\toprule
$B$ & Comparator & $\widehat\Gamma$ & Cohort 95\% CI & Welch $p_+$ H4 & Conditional 95\% CI \\
\midrule
256 & AU & 16.8020 & $[11.2055, 22.3984]$ & $1.08\!\times\!10^{-18}$ & $[16.0541, 17.5498]$ \\
256 & NU & 5.3711 & $[3.6166, 7.1256]$ & $1.76\!\times\!10^{-12}$ & $[4.6832, 6.0590]$ \\
256 & Fresh-U & 0.2268 & $[-0.9013, 1.3550]$ & $0.6922$ & $[-0.1011, 0.5548]$ \\
256 & Fresh-E & 0.2268 & $[-0.9013, 1.3550]$ & $0.6922$ & $[-0.1011, 0.5548]$ \\
334 & AU & 17.1071 & $[11.6258, 22.5884]$ & $2.70\!\times\!10^{-15}$ & $[16.3738, 17.8405]$ \\
334 & NU & 5.6763 & $[3.8950, 7.4576]$ & $8.08\!\times\!10^{-15}$ & $[4.9729, 6.3797]$ \\
334 & Fresh-U & 0.0519 & $[-0.5768, 0.6806]$ & $0.8709$ & $[-0.1555, 0.2593]$ \\
334 & Fresh-E & 0.0519 & $[-0.5768, 0.6806]$ & $0.8709$ & $[-0.1555, 0.2593]$ \\
\bottomrule
\end{tabular}
\par
\endgroup
\end{table}
\begin{table}[!htbp]
\centering\small
\setlength{\tabcolsep}{4.0pt}
\captionsetup{position=top}
\caption{SIGMORPHON-Mixed: accuracy (\%), gain (pp), acceptance, and mean audit calls.}\label{tab:mixedpolicies}
\begin{tabular}{@{}lrrrrrrrr@{}}
\toprule
Policy & Acc.\% & Gain & G1 & G2a & G2b & $\widehat\Delta_T\leq0$ & Used & Saved \\
\midrule
Hybrid & 62.04 & 4.51 & 26/32 & 0/16 & 1/16 & 0/29 & 288.5 & 45.5 \\
Fresh-U & 62.27 & 4.74 & 28/32 & 0/16 & 0/16 & 0/29 & 308.4 & 25.6 \\
Fresh-E & 62.12 & 4.59 & 27/32 & 0/16 & 0/16 & 0/29 & 307.9 & 26.1 \\
Naive-Mean & 62.76 & 5.23 & 32/32 & 1/16 & 3/16 & 3/29 & 334.0 & -- \\
Naive-Z & 62.29 & 4.76 & 28/32 & 0/16 & 1/16 & 0/29 & 334.0 & -- \\
AU & 60.12 & 2.59 & 32/32 & 16/16 & 16/16 & 29/29 & 0.0 & -- \\
NU & 57.53 & 0.00 & 0/32 & 0/16 & 0/16 & 0/29 & 0.0 & -- \\
Fresh-E$_{256}$ & 62.52 & 4.99 & 30/32 & 0/16 & 0/16 & 0/29 & 512.0 & -- \\
\bottomrule
\end{tabular}
\par\vspace{4pt}\noindent\begin{minipage}{\linewidth}\footnotesize Counts are accepted/total cohorts; observed nonpositive gain is a post-hoc diagnostic. Fresh-E$_{256}$ is an over-budget reference ($B=512$); AU/NU require no audit.\end{minipage}
\end{table}
\begin{table}[!htbp]
\centering\footnotesize
\setlength{\tabcolsep}{4pt}
\caption{Complete seven-policy Banking77-Mixed outcomes at the primary and secondary budgets.}\label{tab:bankpolicies}
\begin{tabular}{@{}rlrrrrr@{}}
\toprule
$B$ & Policy & Accepted & Acc.\% & Gain pp & Used & Saved \\
\midrule
256 & Hybrid & 22 & 79.0120 & 5.3711 & 159.8542 & 96.1458 \\
256 & Fresh-U & 21 & 78.7852 & 5.1442 & 207.8333 & 48.1667 \\
256 & Fresh-E & 21 & 78.7852 & 5.1442 & 207.3750 & 48.6250 \\
256 & Naive-Mean & 24 & 79.5085 & 5.8675 & 256.0000 & -- \\
256 & Naive-Z & 22 & 79.0120 & 5.3711 & 256.0000 & -- \\
256 & AU & 48 & 62.2101 & $-11.4309$ & 0.0000 & -- \\
256 & NU & 0 & 73.6410 & 0.0000 & 0.0000 & -- \\
\addlinespace[2pt]
334 & Hybrid & 23 & 79.3172 & 5.6763 & 221.5000 & 112.5000 \\
334 & Fresh-U & 23 & 79.2653 & 5.6244 & 271.0417 & 62.9583 \\
334 & Fresh-E & 23 & 79.2653 & 5.6244 & 270.7917 & 63.2083 \\
334 & Naive-Mean & 24 & 79.5085 & 5.8675 & 334.0000 & -- \\
334 & Naive-Z & 24 & 79.5085 & 5.8675 & 334.0000 & -- \\
334 & AU & 48 & 62.2101 & $-11.4309$ & 0.0000 & -- \\
334 & NU & 0 & 73.6410 & 0.0000 & 0.0000 & -- \\
\bottomrule
\end{tabular}
\par\vspace{4pt}\noindent\begin{minipage}{\linewidth}\footnotesize Used and saved columns count mean virtual incremental endpoint evaluations.\end{minipage}
\end{table}

\subsection{Nested budget sweeps and exact fresh calibration}
\label{app:budget-sweeps}
Tables~\ref{tab:budget-v2} and~\ref{tab:budget-banking} retain every budget and family/recipe subset. At budgets $256,300,334,400,512$, the retrospective SIGMORPHON-Mixed comparison uses Fresh cutoffs $0.0586$, $0.0577$, $0.0570$, $0.0518$, and $0.0553$. Banking77-Mixed retains its frozen cutoffs $0.057$, $0.057$, $0.057$, $0.051$, and $0.055$, certified in Table~\ref{tab:fresh-certificates}. Identical displayed cutoff values need not imply identical Bernstein partitions. These curves use nested prefixes of the same outcome streams.
\begin{table}[!htbp]
\centering\footnotesize
\setlength{\tabcolsep}{4.0pt}
\caption{Complete SIGMORPHON-Mixed budget sweep. H/FU lists accepted cohorts; utility contrasts and intervals are in pp.}\label{tab:budget-v2}
\begin{tabular}{@{}lrrlrrrr@{}}
\toprule
Subset & $B$ & $N$ & H/FU & $\widehat\Gamma$ & Cohort 95\% CI & $p_2$ & Saved H / FU \\
\midrule
All & 256 & 64 & 27/24 & 0.4158 & $[-0.3112, 1.1428]$ & $0.2574$ & 42.39 / 20.69 \\
G1 & 256 & 32 & 27/24 & 0.8316 & $[-0.6486, 2.3118]$ & $0.2606$ & 17.50 / 9.06 \\
Mistral & 256 & 24 & 10/10 & 0.0000 & $[0.0000, 0.0000]$ & $1$ & 53.88 / 28.58 \\
Mistral G1 & 256 & 12 & 10/10 & 0.0000 & $[0.0000, 0.0000]$ & $1$ & 24.00 / 12.00 \\
Qwen & 256 & 40 & 17/14 & 0.6653 & $[-0.5103, 1.8409]$ & $0.2593$ & 35.50 / 15.95 \\
Qwen G1 & 256 & 20 & 17/14 & 1.3306 & $[-1.0924, 3.7535]$ & $0.2647$ & 13.60 / 7.30 \\
All & 300 & 64 & 28/26 & 0.2594 & $[-0.3894, 0.9082]$ & $0.4273$ & 44.67 / 23.91 \\
G1 & 300 & 32 & 27/25 & 0.4898 & $[-0.8331, 1.8127]$ & $0.4559$ & 21.31 / 11.44 \\
Mistral & 300 & 24 & 10/10 & 0.0000 & $[0.0000, 0.0000]$ & $1$ & 57.83 / 33.58 \\
Mistral G1 & 300 & 12 & 10/10 & 0.0000 & $[0.0000, 0.0000]$ & $1$ & 27.17 / 15.67 \\
Qwen & 300 & 40 & 18/16 & 0.4150 & $[-0.6375, 1.4676]$ & $0.4300$ & 36.77 / 18.10 \\
Qwen G1 & 300 & 20 & 17/15 & 0.7837 & $[-1.3977, 2.9650]$ & $0.4613$ & 17.80 / 8.90 \\
All & 334 & 64 & 27/28 & $-0.2243$ & $[-0.7167, 0.2681]$ & $0.3661$ & 45.45 / 25.62 \\
G1 & 334 & 32 & 26/28 & $-0.5066$ & $[-1.5022, 0.4890]$ & $0.3074$ & 22.19 / 12.31 \\
Mistral & 334 & 24 & 10/10 & 0.0000 & $[0.0000, 0.0000]$ & $1$ & 60.83 / 37.17 \\
Mistral G1 & 334 & 12 & 10/10 & 0.0000 & $[0.0000, 0.0000]$ & $1$ & 29.50 / 17.33 \\
Qwen & 334 & 40 & 17/18 & $-0.3589$ & $[-1.1570, 0.4392]$ & $0.3686$ & 36.23 / 18.70 \\
Qwen G1 & 334 & 20 & 16/18 & $-0.8105$ & $[-2.4440, 0.8229]$ & $0.3120$ & 17.80 / 9.30 \\
All & 400 & 64 & 28/29 & $-0.2403$ & $[-0.7525, 0.2719]$ & $0.3520$ & 49.27 / 30.62 \\
G1 & 400 & 32 & 27/29 & $-0.5386$ & $[-1.5745, 0.4972]$ & $0.2971$ & 26.19 / 16.81 \\
Mistral & 400 & 24 & 10/10 & 0.0000 & $[0.0000, 0.0000]$ & $1$ & 66.21 / 44.83 \\
Mistral G1 & 400 & 12 & 10/10 & 0.0000 & $[0.0000, 0.0000]$ & $1$ & 34.00 / 23.00 \\
Qwen & 400 & 40 & 18/19 & $-0.3845$ & $[-1.2145, 0.4455]$ & $0.3545$ & 39.10 / 22.10 \\
Qwen G1 & 400 & 20 & 17/19 & $-0.8618$ & $[-2.5605, 0.8369]$ & $0.3016$ & 21.50 / 13.10 \\
All & 512 & 64 & 29/30 & $-0.0931$ & $[-0.4713, 0.2852]$ & $0.6246$ & 57.56 / 37.03 \\
G1 & 512 & 32 & 29/30 & $-0.1862$ & $[-0.9630, 0.5907]$ & $0.6285$ & 34.62 / 23.31 \\
Mistral & 512 & 24 & 10/11 & $-0.2380$ & $[-0.7305, 0.2544]$ & $0.3277$ & 77.42 / 51.58 \\
Mistral G1 & 512 & 12 & 10/11 & $-0.4761$ & $[-1.5239, 0.5718]$ & $0.3388$ & 44.17 / 30.00 \\
Qwen & 512 & 40 & 19/19 & $-0.0061$ & $[-0.5490, 0.5368]$ & $0.9820$ & 45.65 / 28.30 \\
Qwen G1 & 512 & 20 & 19/19 & $-0.0122$ & $[-1.1504, 1.1260]$ & $0.9823$ & 28.90 / 19.30 \\
\bottomrule
\end{tabular}
\par\vspace{4pt}\noindent\begin{minipage}{\linewidth}\footnotesize Budget-sweep $p_2$ values are unadjusted descriptive cohort tests. Saved counts use the original query order.\end{minipage}
\end{table}
\begin{table}[!htbp]
\centering\footnotesize
\setlength{\tabcolsep}{4.0pt}
\caption{Complete Banking77-Mixed budget sweep. H/FU lists accepted cohorts; utility contrasts and intervals are in pp.}\label{tab:budget-banking}
\begin{tabular}{@{}lrrlrrrr@{}}
\toprule
Subset & $B$ & $N$ & H/FU & $\widehat\Gamma$ & Cohort 95\% CI & $p_2$ & Saved H / FU \\
\midrule
All & 256 & 48 & 22/21 & 0.2268 & $[-0.9013, 1.3550]$ & $0.6877$ & 96.15 / 48.17 \\
G1 & 256 & 24 & 22/21 & 0.4537 & $[-1.8875, 2.7948]$ & $0.6922$ & 17.54 / 9.00 \\
All & 300 & 48 & 23/23 & $-0.1139$ & $[-0.8454, 0.6175]$ & $0.7554$ & 106.71 / 56.75 \\
G1 & 300 & 24 & 23/23 & $-0.2279$ & $[-1.7468, 1.2911]$ & $0.7591$ & 27.71 / 12.42 \\
All & 334 & 48 & 23/23 & 0.0519 & $[-0.5768, 0.6806]$ & $0.8689$ & 112.50 / 62.96 \\
G1 & 334 & 24 & 23/23 & 0.1038 & $[-1.2028, 1.4103]$ & $0.8709$ & 30.33 / 15.25 \\
All & 400 & 48 & 24/22 & 0.4344 & $[-0.1815, 1.0502]$ & $0.1625$ & 122.96 / 75.12 \\
G1 & 400 & 24 & 24/22 & 0.8687 & $[-0.3837, 2.1212]$ & $0.1648$ & 35.42 / 19.50 \\
All & 512 & 48 & 24/23 & 0.2431 & $[-0.2460, 0.7322]$ & $0.3224$ & 142.15 / 95.88 \\
G1 & 512 & 24 & 24/23 & 0.4862 & $[-0.5196, 1.4921]$ & $0.3277$ & 45.25 / 30.25 \\
\bottomrule
\end{tabular}
\par\vspace{4pt}\noindent\begin{minipage}{\linewidth}\footnotesize Budget-sweep $p_2$ values are unadjusted descriptive cohort tests. Saved counts use the original query order.\end{minipage}
\end{table}
\begin{table}[!htbp]
\centering\small
\setlength{\tabcolsep}{12pt}
\caption{All five frozen FreshUnconditional certificates. The bold entry is the largest exact Bernstein coefficient bound for a given horizon.}\label{tab:fresh-certificates}
\renewcommand{\arraystretch}{1.25}
\begin{tabular}{@{}rccr@{}}
\toprule
Pairs & Cutoff & Discordance interval & Certified upper bound \\
\midrule
128 & $57/1000$ & $[0, 1/2]$ & $0.043771457077$ \\
 &  & $[1/2, 3/4]$ & $0.046596654526$ \\
 &  & $[3/4, 7/8]$ & $0.047151289288$ \\
 &  & $[7/8, 1]$ & $\mathbf{0.049881285213}$ \\
\addlinespace[2pt]
150 & $57/1000$ & $[0, 1/2]$ & $0.044470707217$ \\
 &  & $[1/2, 3/4]$ & $0.046659133160$ \\
 &  & $[3/4, 7/8]$ & $0.048127390863$ \\
 &  & $[7/8, 1]$ & $\mathbf{0.049852380548}$ \\
\addlinespace[2pt]
167 & $57/1000$ & $[0, 1/2]$ & $0.045600222182$ \\
 &  & $[1/2, 3/4]$ & $0.047668736635$ \\
 &  & $[3/4, 1]$ & $\mathbf{0.049847044723}$ \\
\addlinespace[2pt]
200 & $51/1000$ & $[0, 1/2]$ & $0.041706242775$ \\
 &  & $[1/2, 1]$ & $\mathbf{0.046804741902}$ \\
\addlinespace[2pt]
256 & $11/200$ & $[0, 1/2]$ & $0.046286626222$ \\
 &  & $[1/2, 1]$ & $\mathbf{0.049799869001}$ \\
\bottomrule
\end{tabular}
\par\vspace{4pt}\noindent\begin{minipage}{\linewidth}\footnotesize Decimals display exact rational bounds. Each interval family partitions $[0,1]$, and every global bound is below $1/20$.\end{minipage}
\end{table}

\subsection{Which cohorts change the policy ranking?}
\label{app:cohort-diagnostics}
The small primary Banking77-Mixed utility contrast combines two opposing effects (Table~\ref{tab:cohort-diagnostics}). Hybrid recovers three beneficial cohorts rejected by Fresh-U$_{128}$, with observed test gains $13.67$, $8.98$, and $12.06$ pp. Their larger reused samples provide sufficiently unbalanced repair/regression counts even when Fresh sees only 24--28 discordances. Conversely, Hybrid misses two beneficial cohorts with conservative drag of $4.49$ and $10.35$ pp, despite test gains of $9.18$ and $14.65$ pp. At $B=334$, the latter becomes accepted; the former remains just above the exact cutoff at $p=0.0503$. Thus neither a positive test gain nor more nominal pairs determines an individual acceptance.
\begin{table}[!htbp]
\centering\footnotesize
\setlength{\tabcolsep}{5pt}
\caption{The five beneficial Banking77-Mixed cohorts on which Hybrid and Fresh-U disagree at $B=256$.}\label{tab:cohort-diagnostics}
\begin{tabular}{@{}rrrrrrr@{}}
\toprule
Case & $\widehat\Delta_T$ pp & Drag pp & H $(b,c)$ & $p_{H,256}$ & $p_{F,128}$ & $p_{H,334}$ \\
\midrule
1 & 13.67 & 5.47 & 36/15 & $0.0023$ & $0.0758$ & $7.74\!\times\!10^{-4}$ \\
2 & 8.98 & $-0.78$ & 32/7 & $3.51\!\times\!10^{-5}$ & $0.0758$ & $1.94\!\times\!10^{-4}$ \\
3 & 12.06 & 2.69 & 42/18 & $0.0013$ & $0.0925$ & $5.47\!\times\!10^{-4}$ \\
4 & 9.18 & 4.49 & 34/22 & $0.0704$ & $0.0168$ & $0.0503$ \\
5 & 14.65 & 10.35 & 37/26 & $0.1037$ & $0.0015$ & $0.0382$ \\
\bottomrule
\end{tabular}
\par\vspace{4pt}\noindent\begin{minipage}{\linewidth}\footnotesize Cases 1--3 are accepted only by Hybrid; cases 4--5 only by Fresh-U at the primary budget. Positive drag is $\widehat\Delta_T-\widehat\Delta_S$. Hybrid uses cutoff $0.05$; Fresh-U uses $0.057$.\end{minipage}
\end{table}

Family heterogeneity has a different explanation. In SIGMORPHON-Mixed, Mistral's G2a/G2b mean test gains are $-12.43/-5.50$ pp, versus Qwen's $-4.62/-1.50$ pp. Rejecting these updates therefore yields more avoided loss for Mistral. Its selection-panel discordance is also higher: $d_S=0.303$ versus $0.214$ for Qwen, and within G1 it is $0.291$ versus $0.218$. The larger Mistral G1 gain ($11.49$ versus $9.83$ pp) is therefore accompanied by more paired noise. All three exact policies make the same decisions in its 24 cohorts; the observed family contrast reflects these gains, losses, and realizations rather than a general dominance ordering.

\subsection{Correlated-candidate Monte Carlo stress tests}
\label{app:mc-grid}
\begin{figure}[!htbp]
\centering
\begin{subfigure}[t]{.32\linewidth}\centering\includegraphics[width=\linewidth]{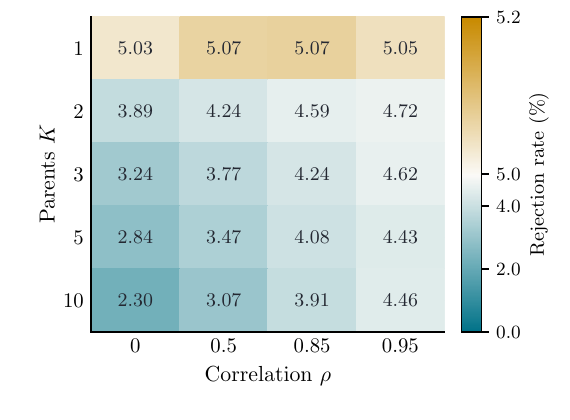}\caption{Null Naive-Z, $d=.05$}\label{fig:mc-null-small}\end{subfigure}\hfill
\begin{subfigure}[t]{.32\linewidth}\centering\includegraphics[width=\linewidth]{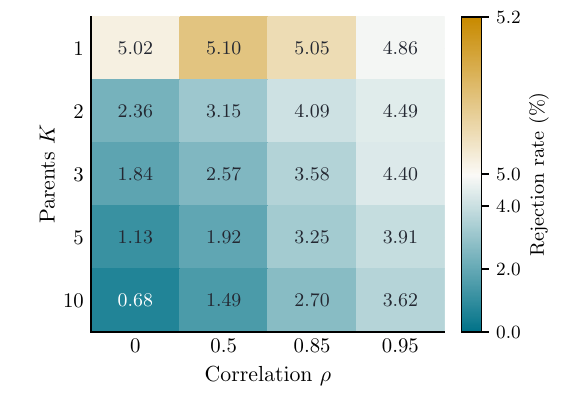}\caption{Null Naive-Z, $d=.22$}\label{fig:mc-null-large}\end{subfigure}\hfill
\begin{subfigure}[t]{.32\linewidth}\centering\includegraphics[width=\linewidth]{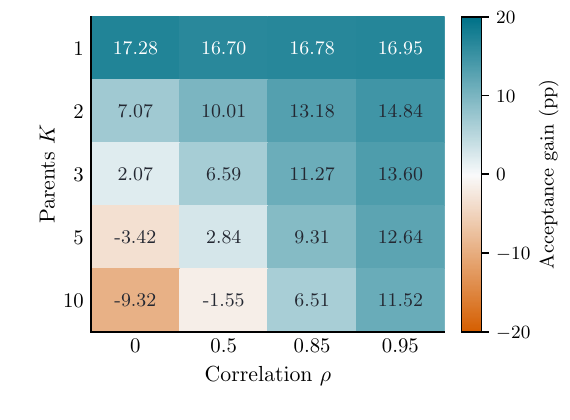}\caption{Power difference, $d=.22$}\label{fig:mc-power}\end{subfigure}
\caption{Monte Carlo dependence diagnostics: (a,b) Naive-Z null rejection versus $K$ and $\rho$ at fixed discordance; (c) Hybrid-minus-Fresh-U power at $d=.22$.}\label{fig:mc-diagnostics}
\end{figure}

The simulation crosses $K\in\{1,2,3,5,10\}$, $\rho\in\{0,.5,.85,.95\}$, and $d\in\{.01,.02,.05,.22\}$ under gains $0$, $-0.005$, and $\min(d/2,0.05)$. Every cell contains 50,000 independent replicates, totaling twelve million. With probability $\rho$, candidates share a common parent--child draw within a row; rows remain independent. Figure~\ref{fig:mc-diagnostics} shows how candidate dependence changes the normal-test rejection rate and beneficial-update power. The maximum normal-test null rate among $K\geq2$ cells is $4.864\%$. Table~\ref{tab:mc} shows the selected extrema, and Table~\ref{tab:mc-aggregate} summarizes all $(d,K)$ groups over correlations. The full 240-cell results remain in the data supplement.

\begin{table}[!htbp]
\centering\footnotesize
\setlength{\tabcolsep}{4pt}
\caption{Descriptive maximum observed acceptance across the eighty cells of each simulation role.}\label{tab:mc}
\begin{tabular}{@{}llrrrrr@{}}
\toprule
Role & Policy & $K$ & $\rho$ & $d$ & Acceptance \% & Pointwise 95\% CI \\
\midrule
Null & Hybrid & 1 & 0.85 & 0.22 & 3.912 & $[3.744, 4.086]$ \\
Null & Fresh-U & 1 & 0.95 & 0.22 & 4.240 & $[4.065, 4.420]$ \\
Null & Naive-Z & 1 & 0.50 & 0.22 & 5.096 & $[4.905, 5.292]$ \\
Null & Naive-Mean & 1 & 0.85 & 0.22 & 47.674 & $[47.235, 48.113]$ \\
Near null & Hybrid & 1 & 0.95 & 0.22 & 2.628 & $[2.490, 2.772]$ \\
Near null & Fresh-U & 1 & 0.85 & 0.22 & 3.186 & $[3.034, 3.344]$ \\
Near null & Naive-Z & 1 & 0.85 & 0.22 & 3.424 & $[3.266, 3.587]$ \\
Near null & Naive-Mean & 1 & 0.00 & 0.22 & 40.404 & $[39.973, 40.836]$ \\
Beneficial & Hybrid & 1 & 0.85 & 0.22 & 53.266 & $[52.827, 53.704]$ \\
Beneficial & Fresh-U & 1 & 0.85 & 0.22 & 36.482 & $[36.060, 36.906]$ \\
Beneficial & Naive-Z & 1 & 0.85 & 0.05 & 63.060 & $[62.635, 63.483]$ \\
Beneficial & Naive-Mean & 1 & 0.85 & 0.05 & 96.998 & $[96.845, 97.146]$ \\
\bottomrule
\end{tabular}
\par\vspace{4pt}\noindent\begin{minipage}{\linewidth}\footnotesize Each cell has 50,000 replicates. Pointwise \mbox{Clopper--Pearson} intervals are not simultaneous for selected maxima.\end{minipage}
\end{table}
\begin{table}[!htbp]
\centering\footnotesize
\setlength{\tabcolsep}{3.7pt}
\caption{Monte Carlo summary across the four candidate-correlation settings for each $(d,K)$. Acceptance and power entries are percentages.}\label{tab:mc-aggregate}
\begin{tabular}{@{}rrrrrrr@{}}
\toprule
$d$ & $K$ & Null H max & Null Z max & Power H range & Power FU range & Saved H/FU range \\
\midrule
0.01 & 1 & 0.46 & 4.57 & 3.66--3.86 & 0.54--0.62 & 0.928--1.264 \\
0.01 & 2 & 0.43 & 4.21 & 3.36--3.71 & 0.55--0.66 & 0.934--1.285 \\
0.01 & 3 & 0.42 & 4.12 & 3.16--3.67 & 0.58--0.62 & 0.935--1.300 \\
0.01 & 5 & 0.40 & 4.20 & 2.90--3.55 & 0.57--0.62 & 0.945--1.314 \\
0.01 & 10 & 0.35 & 4.10 & 2.64--3.51 & 0.54--0.63 & 0.946--1.328 \\
0.02 & 1 & 1.60 & 5.07 & 16.65--16.99 & 5.16--5.34 & 0.895--1.297 \\
0.02 & 2 & 1.51 & 4.86 & 15.22--16.43 & 5.28--5.48 & 0.909--1.333 \\
0.02 & 3 & 1.40 & 4.60 & 13.91--16.08 & 5.24--5.42 & 0.917--1.346 \\
0.02 & 5 & 1.42 & 4.61 & 13.25--15.95 & 5.22--5.39 & 0.927--1.372 \\
0.02 & 10 & 1.34 & 4.44 & 12.06--15.89 & 5.17--5.41 & 0.930--1.395 \\
0.05 & 1 & 3.03 & 5.07 & 52.85--53.02 & 30.59--31.02 & 1.022--1.313 \\
0.05 & 2 & 2.76 & 4.72 & 48.41--51.74 & 30.66--31.03 & 1.026--1.379 \\
0.05 & 3 & 2.64 & 4.62 & 45.88--51.07 & 30.53--30.97 & 1.042--1.410 \\
0.05 & 5 & 2.54 & 4.43 & 42.94--50.83 & 30.79--31.23 & 1.039--1.459 \\
0.05 & 10 & 2.55 & 4.46 & 40.20--50.27 & 30.62--30.92 & 1.033--1.509 \\
0.22 & 1 & 3.91 & 5.10 & 52.47--53.27 & 35.77--36.48 & 1.181--1.378 \\
0.22 & 2 & 3.48 & 4.49 & 42.78--50.74 & 35.71--35.94 & 1.185--1.546 \\
0.22 & 3 & 3.45 & 4.40 & 38.37--49.59 & 35.79--36.30 & 1.194--1.631 \\
0.22 & 5 & 3.00 & 3.91 & 32.47--48.45 & 35.78--35.91 & 1.194--1.727 \\
0.22 & 10 & 2.70 & 3.62 & 26.29--47.46 & 35.52--36.05 & 1.215--1.852 \\
\bottomrule
\end{tabular}
\par\vspace{4pt}\noindent\begin{minipage}{\linewidth}\footnotesize Power uses the simulated positive-gain role $\Delta=\min(d/2,0.05)$. Saved-query ranges cover all three simulation roles and all four correlations. The complete 240-cell results are included in the data supplement.\end{minipage}
\end{table}

\subsection{Historical budget frontier and synthetic decision trade-off}
\label{app:historical-figures}
\begin{figure}[!htbp]
\centering
\begin{subfigure}[t]{.48\linewidth}\centering\includegraphics[width=\linewidth]{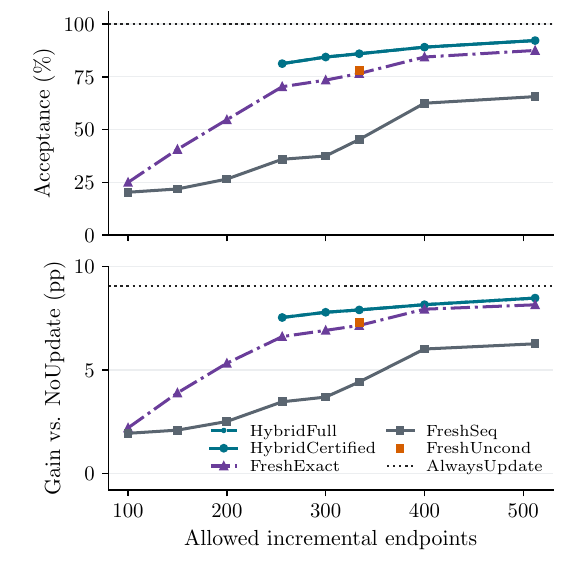}\caption{Allowed incremental endpoints}\label{fig:frontier-allowed}\end{subfigure}\hfill
\begin{subfigure}[t]{.48\linewidth}\centering\includegraphics[width=\linewidth]{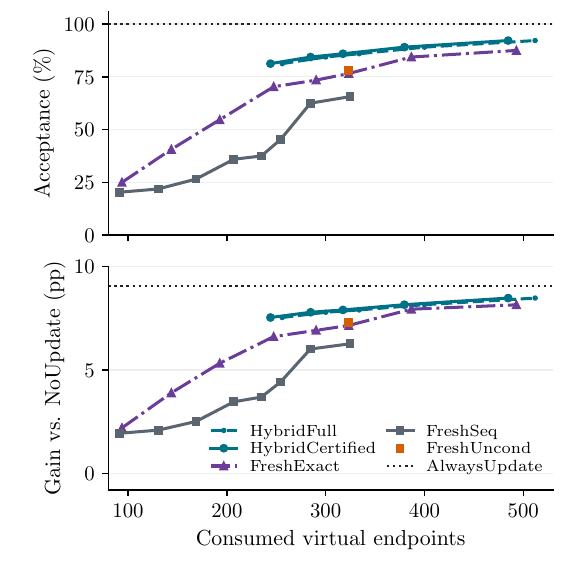}\caption{Consumed incremental endpoints}\label{fig:frontier-consumed}\end{subfigure}
\caption{Historical budget frontier for SIGMORPHON-II ($N=64$). Fresh-U$_{167}$ is a single certified operating point.}\label{fig:frontier}
\end{figure}

\begin{figure}[!htbp]
\centering
\includegraphics[width=.65\linewidth]{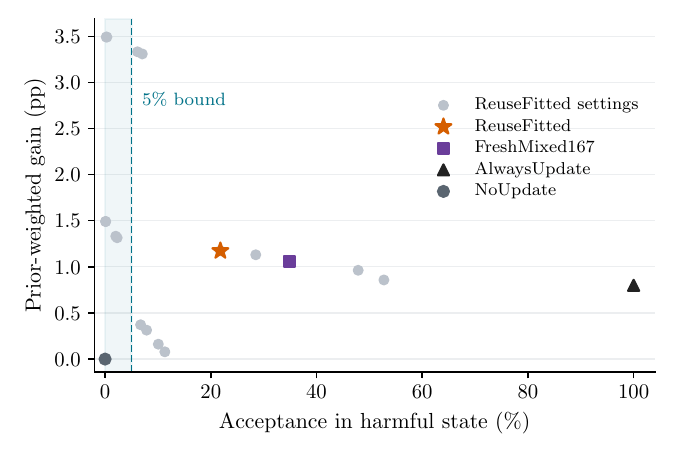}
\caption{Synthetic safety--utility trade-off across the representative fitted policy and sixteen fitted settings. The horizontal axis is harmful-update adoption and the vertical axis is net utility.}\label{fig:safety}
\end{figure}

Figure~\ref{fig:frontier} reports the 64-cohort SIGMORPHON-II frontier. Allowed budget and consumed budget are different horizontal axes because deterministic completion can stop before the endpoint; observed acceptance rates are cohort outcomes. Figure~\ref{fig:safety} reports the separate synthetic decision study, in which harmful states are known by construction. Its representative fitted rule accepts harmful states with probability $0.21814$, exceeding the 5\% harmful-adoption target. The vertical line marks Hybrid's theoretical error bound; that experiment does not evaluate its utility.

\FloatBarrier
\subsection{Complete policy outcomes within each model family}
\label{app:family-policies}
Tables~\ref{tab:family-qwen} and~\ref{tab:family-mistral} report all eight policies within each prespecified family. Accuracy averages $\widehat p_J(T)+A\widehat\Delta_T$ over cohorts, with frozen action $A$; gain subtracts that family's NoUpdate accuracy. Qwen's G1/G2a/G2b allocation is $20/10/10$, versus Mistral's $12/6/6$. The observed nonpositive-test-gain subsets instead contain 17 and 12 cohorts. Neither partition identifies known population nulls.

Qwen's Hybrid and Fresh-U accuracies are $63.87\%$ and $64.22\%$; Mistral's Hybrid, Fresh-U, and Fresh-E have identical actions and $59.01\%$ accuracy. These outcomes complement Table~\ref{tab:mixed}'s adjusted contrasts without establishing a general family-specific ranking. Matched-budget audits have ceiling 334; Fresh-E$_{256}$ spends 512 endpoints as a larger-budget reference, while AU and NU require no audit.

\begin{table}[!htbp]
\centering\footnotesize
\setlength{\tabcolsep}{3.1pt}
\renewcommand{\arraystretch}{1.04}
\caption{SIGMORPHON-Mixed, Qwen3-8B ($N=40$): complete policies at prospective $B=334$, with Fresh-E$_{256}$ as a larger-budget reference.}\label{tab:family-qwen}
\begin{tabular}{@{}lrrrrrrrr@{}}
\toprule
Policy & Acc.\% & Gain pp & G1 & G2a & G2b & $\widehat\Delta_T\leq0$ & Used & Saved \\
\midrule
Hybrid & 63.87 & 4.11 & 16/20 & 0/10 & 1/10 & 0/17 & 297.8 & 36.2 \\
Fresh-U$_{167}$ & 64.22 & 4.47 & 18/20 & 0/10 & 0/10 & 0/17 & 315.3 & 18.7 \\
Fresh-E$_{167}$ & 63.98 & 4.23 & 17/20 & 0/10 & 0/10 & 0/17 & 314.7 & 19.3 \\
Naive-Mean & 64.71 & 4.95 & 20/20 & 1/10 & 2/10 & 2/17 & 334.0 & -- \\
Naive-Z & 64.27 & 4.51 & 18/20 & 0/10 & 1/10 & 0/17 & 334.0 & -- \\
AU & 63.14 & 3.39 & 20/20 & 10/10 & 10/10 & 17/17 & 0.0 & -- \\
NU & 59.76 & 0.00 & 0/20 & 0/10 & 0/10 & 0/17 & 0.0 & -- \\
Fresh-E$_{256}$ & 64.49 & 4.73 & 19/20 & 0/10 & 0/10 & 0/17 & 512.0 & -- \\
\bottomrule
\end{tabular}
\par\vspace{4pt}\noindent\begin{minipage}{\linewidth}\footnotesize Hybrid: HybridCertified; Fresh-U/E: FreshUnconditional/Exact; AU/NU: AlwaysUpdate/NoUpdate. Counts are accepted/eligible cohorts; Used/Saved are mean virtual endpoints. Fresh-E$_{256}$ spends $512$ without curtailment; AU/NU use no audit. Dashes mark unreported completion savings. The nonpositive-test-gain subset is post-hoc.\end{minipage}
\end{table}
\begin{table}[!htbp]
\centering\footnotesize
\setlength{\tabcolsep}{3.1pt}
\renewcommand{\arraystretch}{1.04}
\caption{SIGMORPHON-Mixed, Mistral-7B ($N=24$): complete policies at prospective $B=334$, with Fresh-E$_{256}$ as a larger-budget reference.}\label{tab:family-mistral}
\begin{tabular}{@{}lrrrrrrrr@{}}
\toprule
Policy & Acc.\% & Gain pp & G1 & G2a & G2b & $\widehat\Delta_T\leq0$ & Used & Saved \\
\midrule
Hybrid & 59.01 & 5.18 & 10/12 & 0/6 & 0/6 & 0/12 & 273.2 & 60.8 \\
Fresh-U$_{167}$ & 59.01 & 5.18 & 10/12 & 0/6 & 0/6 & 0/12 & 296.8 & 37.2 \\
Fresh-E$_{167}$ & 59.01 & 5.18 & 10/12 & 0/6 & 0/6 & 0/12 & 296.7 & 37.3 \\
Naive-Mean & 59.52 & 5.69 & 12/12 & 0/6 & 1/6 & 1/12 & 334.0 & -- \\
Naive-Z & 59.01 & 5.18 & 10/12 & 0/6 & 0/6 & 0/12 & 334.0 & -- \\
AU & 55.09 & 1.26 & 12/12 & 6/6 & 6/6 & 12/12 & 0.0 & -- \\
NU & 53.82 & 0.00 & 0/12 & 0/6 & 0/6 & 0/12 & 0.0 & -- \\
Fresh-E$_{256}$ & 59.25 & 5.42 & 11/12 & 0/6 & 0/6 & 0/12 & 512.0 & -- \\
\bottomrule
\end{tabular}
\par\vspace{4pt}\noindent\begin{minipage}{\linewidth}\footnotesize Hybrid: HybridCertified; Fresh-U/E: FreshUnconditional/Exact; AU/NU: AlwaysUpdate/NoUpdate. Counts are accepted/eligible cohorts; Used/Saved are mean virtual endpoints. Fresh-E$_{256}$ spends $512$ without curtailment; AU/NU use no audit. Dashes mark unreported completion savings. The nonpositive-test-gain subset is post-hoc.\end{minipage}
\end{table}

\subsection{Where completion certificates save evaluations}
\label{app:completion-subgroups}
Table~\ref{tab:completion-subgroups} crosses family, recipe, and budget in the original query order. For virtual cost $q_i$, saving is the cohort mean of $B-q_i$. The reduction percentage divides Hybrid's mean saving by $B$; the final column divides it by Fresh-U's mean saving. Thus the columns measure the fraction of a full audit avoided and the relative saved calls, not physical training or inference costs.

At Banking77's primary $B=256$, G2a saves $166.83$ virtual endpoints under Hybrid versus $76.83$ under Fresh-U, giving a $65.2\%$ full-budget reduction and $2.17\times$ saved calls. G2b saves $182.67$ versus $97.83$, giving $71.4\%$ and $1.87\times$. Such rejection savings accord with Proposition~\ref{prop:completion}: known parent successes restrict attainable repairs on unread reused items. Stopping cost nevertheless depends on query order. SIGMORPHON-Mixed's $B=256$ replay is retrospective; its prospective endpoint is $B=334$, and Banking77 retains both frozen endpoints. At $B=256$, both fresh audits use 128 pairs, but inflection uses cutoff $0.0586$ and Banking77 uses $0.057$.

\begin{table}[!htbp]
\centering\footnotesize
\setlength{\tabcolsep}{2.8pt}
\renewcommand{\arraystretch}{1.04}
\caption{Completion savings by family, assigned regime, and budget, retaining the original query order. $N$ counts cohorts.}\label{tab:completion-subgroups}
\begin{tabular}{@{}lrrrrrrrr@{}}
\toprule
Population / regime & $B$ & $N$ & \shortstack{H\\used} & \shortstack{H\\saved} & \shortstack{FE\\saved} & \shortstack{FU\\saved} & \shortstack{H/$B$\\(\%)} & \shortstack{Saved\\H/FU} \\
\midrule
Mixed: pooled: All & 256 & 64 & 213.6 & 42.4 & 20.9 & 20.7 & 16.6 & 2.049 \\
Mixed: pooled: G1 & 256 & 32 & 238.5 & 17.5 & 8.8 & 9.1 & 6.8 & 1.931 \\
Mixed: pooled: G2a & 256 & 16 & 176.6 & 79.4 & 38.4 & 37.8 & 31.0 & 2.104 \\
Mixed: pooled: G2b & 256 & 16 & 200.9 & 55.1 & 27.8 & 26.9 & 21.5 & 2.051 \\
\addlinespace[2pt]
Mixed: Qwen: All & 256 & 40 & 220.5 & 35.5 & 16.4 & 15.9 & 13.9 & 2.226 \\
Mixed: Qwen: G1 & 256 & 20 & 242.4 & 13.6 & 7.3 & 7.3 & 5.3 & 1.863 \\
Mixed: Qwen: G2a & 256 & 10 & 189.9 & 66.1 & 27.8 & 27.2 & 25.8 & 2.430 \\
Mixed: Qwen: G2b & 256 & 10 & 207.3 & 48.7 & 23.0 & 22.0 & 19.0 & 2.214 \\
\addlinespace[2pt]
Mixed: Mistral: All & 256 & 24 & 202.1 & 53.9 & 28.6 & 28.6 & 21.0 & 1.885 \\
Mixed: Mistral: G1 & 256 & 12 & 232.0 & 24.0 & 11.3 & 12.0 & 9.4 & 2.000 \\
Mixed: Mistral: G2a & 256 & 6 & 154.3 & 101.7 & 56.0 & 55.3 & 39.7 & 1.837 \\
Mixed: Mistral: G2b & 256 & 6 & 190.2 & 65.8 & 35.7 & 35.0 & 25.7 & 1.881 \\
\addlinespace[2pt]
Mixed: pooled: All & 334 & 64 & 288.5 & 45.5 & 26.1 & 25.6 & 13.6 & 1.774 \\
Mixed: pooled: G1 & 334 & 32 & 311.8 & 22.2 & 12.2 & 12.3 & 6.6 & 1.802 \\
Mixed: pooled: G2a & 334 & 16 & 251.8 & 82.2 & 46.6 & 46.0 & 24.6 & 1.787 \\
Mixed: pooled: G2b & 334 & 16 & 278.8 & 55.2 & 33.1 & 31.9 & 16.5 & 1.733 \\
\addlinespace[2pt]
Mixed: Qwen: All & 334 & 40 & 297.8 & 36.2 & 19.3 & 18.7 & 10.8 & 1.937 \\
Mixed: Qwen: G1 & 334 & 20 & 316.2 & 17.8 & 9.5 & 9.3 & 5.3 & 1.914 \\
Mixed: Qwen: G2a & 334 & 10 & 270.3 & 63.7 & 33.0 & 32.2 & 19.1 & 1.978 \\
Mixed: Qwen: G2b & 334 & 10 & 288.4 & 45.6 & 25.2 & 24.0 & 13.7 & 1.900 \\
\addlinespace[2pt]
Mixed: Mistral: All & 334 & 24 & 273.2 & 60.8 & 37.3 & 37.2 & 18.2 & 1.637 \\
Mixed: Mistral: G1 & 334 & 12 & 304.5 & 29.5 & 16.8 & 17.3 & 8.8 & 1.702 \\
Mixed: Mistral: G2a & 334 & 6 & 221.0 & 113.0 & 69.3 & 69.0 & 33.8 & 1.638 \\
Mixed: Mistral: G2b & 334 & 6 & 262.7 & 71.3 & 46.3 & 45.0 & 21.4 & 1.585 \\
\addlinespace[2pt]
Banking77: All & 256 & 48 & 159.9 & 96.1 & 48.6 & 48.2 & 37.6 & 1.996 \\
Banking77: G1 & 256 & 24 & 238.5 & 17.5 & 8.9 & 9.0 & 6.9 & 1.949 \\
Banking77: G2a & 256 & 12 & 89.2 & 166.8 & 77.7 & 76.8 & 65.2 & 2.171 \\
Banking77: G2b & 256 & 12 & 73.3 & 182.7 & 99.0 & 97.8 & 71.4 & 1.867 \\
\addlinespace[2pt]
Banking77: All & 334 & 48 & 221.5 & 112.5 & 63.2 & 63.0 & 33.7 & 1.787 \\
Banking77: G1 & 334 & 24 & 303.7 & 30.3 & 14.9 & 15.2 & 9.1 & 1.989 \\
Banking77: G2a & 334 & 12 & 150.1 & 183.9 & 99.0 & 98.0 & 55.1 & 1.877 \\
Banking77: G2b & 334 & 12 & 128.6 & 205.4 & 124.0 & 123.3 & 61.5 & 1.666 \\
\bottomrule
\end{tabular}
\par\vspace{4pt}\noindent\begin{minipage}{\linewidth}\footnotesize H/FE/FU denote HybridCertified/FreshExact/FreshUnconditional. The last column compares saved calls. Inflection $B=256$ is retrospective; other rows retain frozen rules, with domain-specific Fresh cutoffs. Family and pooled rows overlap. Virtual endpoints do not measure runtime, token, or monetary savings.\end{minipage}
\end{table}

\subsection{Six study summaries and the scope of retrospective pooling}
\label{app:retrospective-synthesis}
Table~\ref{tab:retrospective-synthesis} preserves the distinct evidence status of six summaries. SIGMORPHON-Hist gives equal weight to each of its 48 SIGMORPHON-I and 64 SIGMORPHON-II cohorts. Beneficial-144 adds the 32 assigned SIGMORPHON-Mixed G1 cohorts; Beneficial-168 adds the 24 Banking77 G1 cohorts at $B=334$. Inclusion follows study membership and assigned recipe, without test-sign filtering; these names imply neither independently sampled tasks nor guaranteed positive effects.

Beneficial-168 accepts $148/168=88.10\%$ under Hybrid and $137/168=81.55\%$ under Fresh-U. Its deployment contrast is $+0.6041$ pp, with ordinary cohort interval $[+0.0575,+1.1507]$ pp and unadjusted $p_2=0.0305$. The mixed-regime $B=256$ combination accepts $49/112$ versus $45/112$, with contrast $+0.3348$ pp, interval $[-0.2920,+0.9617]$ pp, and $p_2=0.2922$, retaining fresh cutoffs $0.0586$ for inflection and $0.057$ for Banking77. Prospective-Mixed-112 describes cohort collection: the combination is retrospective, including a counterfactual inflection budget. Neither synthesis replaces the original multiplicity-controlled comparisons. Historical rows preserve token-budget and recovery constraints, unlike the pure endpoint frontier below.

\begin{table}[!htbp]
\centering\footnotesize
\setlength{\tabcolsep}{2.4pt}
\renewcommand{\arraystretch}{1.04}
\caption{Six study summaries against the strengthened fresh comparator. Acceptance counts share denominator $N$; $\widehat\Gamma$ is Hybrid-minus-Fresh accuracy in pp.}\label{tab:retrospective-synthesis}
\begin{tabular}{@{}lrrlrrr@{}}
\toprule
Study / synthesis & $N$ & $B$ & \shortstack{Accepted\\H / FU} & $\widehat\Gamma$ & Cohort 95\% CI & $p_2$ \\
\midrule
SIGMORPHON-I & 48 & 334 & 44 / 36 & $+1.5727$ & $[+0.3652,\,+2.7802]$ & 0.0118 \\
SIGMORPHON-II & 64 & 334 & 55 / 50 & $+0.6207$ & $[-0.2642,\,+1.5055]$ & 0.1659 \\
SIGMORPHON-Hist & 112 & 334 & 99 / 86 & $+1.0287$ & $[+0.3115,\,+1.7458]$ & 0.0053 \\
Beneficial-144 & 144 & 334 & 125 / 114 & $+0.6875$ & $[+0.0834,\,+1.2916]$ & 0.0260 \\
Beneficial-168 & 168 & 334 & 148 / 137 & $+0.6041$ & $[+0.0575,\,+1.1507]$ & 0.0305 \\
\shortstack[l]{Prospective-Mixed-112} & 112 & 256 & 49 / 45 & $+0.3348$ & $[-0.2920,\,+0.9617]$ & 0.2922 \\
\bottomrule
\end{tabular}
\par\vspace{4pt}\noindent\begin{minipage}{\linewidth}\footnotesize SIGMORPHON-I's comparator was strengthened after outcomes; SIGMORPHON-II reproduces its primary contrast. Historical $B=334$ is a nominal target retaining original token caps and recovery constraints. All pooling is retrospective: Beneficial rows append assigned G1 without test-sign filtering; Mixed-112 includes a counterfactual inflection budget. Intervals and $p_2$ are ordinary cohort-$t$ sensitivities, not a new confirmatory family.\end{minipage}
\end{table}

\subsection{Qwen disagreements and the distinction between recipes and test signs}
\label{app:qwen-disagreements}
Table~\ref{tab:qwen-disagreements} complements Banking77's Table~\ref{tab:cohort-diagnostics} with every Qwen disagreement at $B=334$, regardless of test gain. Hybrid alone accepts \texttt{c021} and \texttt{c030}; Fresh-U$_{167}$ alone accepts \texttt{c015}, \texttt{c026}, and \texttt{c037}. Frozen audit actions and costs are reported alongside independent-test accuracies.

For G2b cohort \texttt{c030}, parent and child scores are 1,205 and 1,243 correct out of 2,048 sampled test occurrences, giving $100(38/2048)=1.85546875$ pp. Its adoption contributes to G2b but not to the nonpositive-test-gain count: $1/10$ versus $0/17$ within Qwen, and $1/16$ versus $0/29$ after adding Mistral. The distinction reconciles recipe and sign summaries without turning a noisy positive estimate into proof of a positive population effect.

\begin{table}[!htbp]
\centering\footnotesize
\setlength{\tabcolsep}{2.8pt}
\renewcommand{\arraystretch}{1.04}
\caption{All five Qwen disagreements between HybridCertified and FreshUnconditional$_{167}$ at $B=334$, with paired independent-test accuracies.}\label{tab:qwen-disagreements}
\begin{tabular}{@{}llrrrllrr@{}}
\toprule
Cohort & Recipe & $\widehat p_J(T)$\% & $\widehat p_C(T)$\% & $\widehat\Delta_T$ pp & H action & FU action & H used & FU used \\
\midrule
\texttt{c015} & G1 & 58.74 & 68.26 & $+9.52$ & Retain & Accept & 334 & 314 \\
\texttt{c021} & G1 & 57.23 & 64.60 & $+7.37$ & Accept & Retain & 312 & 322 \\
\texttt{c026} & G1 & 61.23 & 67.68 & $+6.45$ & Retain & Accept & 334 & 334 \\
\texttt{c030} & G2b & 58.84 & 60.69 & $+1.86$ & Accept & Retain & 332 & 316 \\
\texttt{c037} & G1 & 62.50 & 70.12 & $+7.62$ & Retain & Accept & 320 & 326 \\
\bottomrule
\end{tabular}
\par\vspace{4pt}\noindent\begin{minipage}{\linewidth}\footnotesize Test streams contain 2,048 sampled occurrences, not distinct items. Actions precede test analysis; costs are virtual endpoints in the original order. G2b cohort \texttt{c030} has positive observed gain; recipe labels do not establish population nulls.\end{minipage}
\end{table}

\subsection{Milestones in the complete offline fresh calibration}
\label{app:offline-calibration-milestones}
Table~\ref{tab:calibration-milestones} samples eleven of the 129 calibrations for $n=128,\ldots,256$. Each provides a rational cutoff and exact Bernstein upper bound after three uniform dyadic subdivisions. Integer re-evaluation reproduces all 129 bounds below $1/20$; table decimals display those rational values. These are certified bounds, not estimated suprema or claims of maximal cutoffs.

Cutoffs need not be monotone: $0.0586$ at $n=128$ falls to $0.0514$ at $n=136$, then rises to $0.0565$ at $n=144$. Discrete binomial rejection counts alter the size polynomial and contribute to Figure~\ref{fig:crossover}'s sawtooth structure. This sequence is separate from the five frozen Banking77 certificates in Table~\ref{tab:fresh-certificates}; substitution would change the earlier policy. Proposition~\ref{prop:bernstein} certifies independent fresh audits, without transferring these cutoffs to selected-panel reuse.

\begin{table}[!htbp]
\centering\footnotesize
\setlength{\tabcolsep}{8pt}
\renewcommand{\arraystretch}{1.04}
\caption{Milestones from 129 offline Bernstein calibrations, $128\leq n\leq256$. Upper bounds display the exact rational certificates as decimals.}\label{tab:calibration-milestones}
\begin{tabular}{@{}rcrrr@{}}
\toprule
Pairs $n$ & Rational $c_n$ & Decimal $c_n$ & Bernstein upper bound & Depth \\
\midrule
128 & $293/5000$ & 0.0586 & 0.049881285213 & 3 \\
136 & $257/5000$ & 0.0514 & 0.044761725453 & 3 \\
144 & $113/2000$ & 0.0565 & 0.049591481988 & 3 \\
150 & $577/10000$ & 0.0577 & 0.049852380548 & 3 \\
160 & $281/5000$ & 0.0562 & 0.049407334787 & 3 \\
167 & $57/1000$ & 0.0570 & 0.049692685669 & 3 \\
180 & $113/2000$ & 0.0565 & 0.049645503222 & 3 \\
200 & $259/5000$ & 0.0518 & 0.046495199587 & 3 \\
224 & $539/10000$ & 0.0539 & 0.048538089825 & 3 \\
240 & $531/10000$ & 0.0531 & 0.048331102882 & 3 \\
256 & $553/10000$ & 0.0553 & 0.049379986505 & 3 \\
\bottomrule
\end{tabular}
\par\vspace{4pt}\noindent\begin{minipage}{\linewidth}\footnotesize Depth three uses eight uniform dyadic subintervals of $[0,1]$. Exact integer subdivision reproduces every bound below $1/20$. These offline calibrations differ from Table~\ref{tab:fresh-certificates}'s five frozen Banking77 certificates; equal cutoffs need not mean equal partitions or bounds. Maximal cutoffs are not claimed.\end{minipage}
\end{table}

\subsection{Numerical counterparts of the frontier and synthetic trade-off}
\label{app:frontier-safety-tables}
\paragraph{Historical endpoint frontier.}
Table~\ref{tab:frontier-numerical} completes Figure~\ref{fig:frontier}'s policy--budget grid. Budget $B$ is the ceiling; mean used endpoints measure stopping cost. Every budget reuses the same 64 SIGMORPHON-II cohorts, so the curves compare nested streams rather than independent replications. Gain intervals concern deployment improvement over NoUpdate, not Hybrid-minus-Fresh contrasts. Full-panel Hybrid is unavailable below $B=256$; Fresh-U$_{167}$ has only its certified $B=334$ point. These absent cells preserve the selection-panel and calibration requirements.

\paragraph{Synthetic state-conditional adoption.}
Tables~\ref{tab:safety-settings} and~\ref{tab:safety-methods} retain sixteen ReuseFitted settings and four methods from Figure~\ref{fig:safety}, with positive and negative states known by construction. Each row satisfies $g=d\{p a_+-(1-p)a_-\}$ for state-conditional adoption probabilities $a_+,a_-$, checking utility consistency without reconstructing the simulation. Coordinate $q$ retains its original parameterization; its meaning, replicate counts, and evaluation costs are not inferred.

ReuseFitted's representative gain and harmful-state adoption are $1.1735$ pp and $21.814\%$, versus $1.0581$ pp and $34.816\%$ for FreshMixed167, a utility-oriented comparator distinct from FreshUnconditional$_{167}$. Hybrid was not evaluated; its bound supplies no empirical estimates.

\begin{table}[!htbp]
\centering\footnotesize
\setlength{\tabcolsep}{5.5pt}
\renewcommand{\arraystretch}{1.04}
\caption{All sixteen ReuseFitted settings behind Figure~\ref{fig:safety}: known-state conditional adoption and prior-weighted net utility.}\label{tab:safety-settings}
\begin{tabular}{@{}rrrrrrr@{}}
\toprule
$q$ & $d$ & $p$ & History $h$ & Positive adoption \% & Negative adoption \% & Gain pp \\
\midrule
0.1 & 0.02 & 0.3 & 16 & 70.655 & 7.855 & 0.3140 \\
0.1 & 0.02 & 0.3 & 64 & 77.690 & 6.723 & 0.3720 \\
0.1 & 0.02 & 0.7 & 16 & 93.005 & 28.509 & 1.1310 \\
0.1 & 0.02 & 0.7 & 64 & 93.171 & 21.814 & 1.1735 \\
\addlinespace[2pt]
0.1 & 0.05 & 0.3 & 16 & 99.655 & 0.116 & 1.4908 \\
0.1 & 0.05 & 0.3 & 64 & 99.750 & 0.097 & 1.4929 \\
0.1 & 0.05 & 0.7 & 16 & 99.900 & 0.341 & 3.4914 \\
0.1 & 0.05 & 0.7 & 64 & 99.866 & 0.215 & 3.4921 \\
\addlinespace[2pt]
0.4 & 0.02 & 0.3 & 16 & 39.512 & 11.312 & 0.0787 \\
0.4 & 0.02 & 0.3 & 64 & 50.443 & 10.077 & 0.1616 \\
0.4 & 0.02 & 0.7 & 16 & 83.892 & 52.751 & 0.8580 \\
0.4 & 0.02 & 0.7 & 64 & 89.299 & 47.885 & 0.9629 \\
\addlinespace[2pt]
0.4 & 0.05 & 0.3 & 16 & 93.023 & 2.284 & 1.3154 \\
0.4 & 0.05 & 0.3 & 64 & 93.568 & 2.025 & 1.3326 \\
0.4 & 0.05 & 0.7 & 16 & 97.556 & 7.046 & 3.3088 \\
0.4 & 0.05 & 0.7 & 64 & 97.834 & 6.137 & 3.3321 \\
\bottomrule
\end{tabular}
\par\vspace{4pt}\noindent\begin{minipage}{\linewidth}\footnotesize Coordinates retain the original parameterization; $h$ counts history banks, not replicates. Gain satisfies $g=d\{p a_+-(1-p)a_-\}$ for the displayed conditional adoption probabilities. The numerical summary does not define $q$ further or supply replicate counts or query costs. These settings differ from prospective G1/G2 recipes.\end{minipage}
\end{table}
\begin{table}[!htbp]
\centering\footnotesize
\setlength{\tabcolsep}{3.6pt}
\renewcommand{\arraystretch}{1.04}
\caption{Figure~\ref{fig:frontier}'s eight-budget, four-policy frontier for SIGMORPHON-II ($N=64$). Gain is deployment improvement over NoUpdate.}\label{tab:frontier-numerical}
\begin{tabular}{@{}rlrrrrr@{}}
\toprule
$B$ & Policy & Accepted & Rate \% & Mean used & Gain pp & Cohort 95\% CI \\
\midrule
100 & HybridCertified & ---\textsuperscript{a} & ---\textsuperscript{a} & ---\textsuperscript{a} & ---\textsuperscript{a} & ---\textsuperscript{a} \\
100 & FreshExact & 16/64 & 25.00 & 93.78 & 2.2076 & $[+1.2225,\,+3.1926]$ \\
100 & FreshSeq & 13/64 & 20.31 & 91.06 & 1.9455 & $[+0.9517,\,+2.9393]$ \\
100 & Fresh-U$_{167}$ & ---\textsuperscript{b} & ---\textsuperscript{b} & ---\textsuperscript{b} & ---\textsuperscript{b} & ---\textsuperscript{b} \\
\addlinespace[2pt]
150 & HybridCertified & ---\textsuperscript{a} & ---\textsuperscript{a} & ---\textsuperscript{a} & ---\textsuperscript{a} & ---\textsuperscript{a} \\
150 & FreshExact & 26/64 & 40.62 & 143.75 & 3.8975 & $[+2.6848,\,+5.1101]$ \\
150 & FreshSeq & 14/64 & 21.88 & 130.53 & 2.0996 & $[+1.0776,\,+3.1216]$ \\
150 & Fresh-U$_{167}$ & ---\textsuperscript{b} & ---\textsuperscript{b} & ---\textsuperscript{b} & ---\textsuperscript{b} & ---\textsuperscript{b} \\
\addlinespace[2pt]
200 & HybridCertified & ---\textsuperscript{a} & ---\textsuperscript{a} & ---\textsuperscript{a} & ---\textsuperscript{a} & ---\textsuperscript{a} \\
200 & FreshExact & 35/64 & 54.69 & 192.78 & 5.3280 & $[+4.0649,\,+6.5911]$ \\
200 & FreshSeq & 17/64 & 26.56 & 168.75 & 2.5169 & $[+1.4402,\,+3.5937]$ \\
200 & Fresh-U$_{167}$ & ---\textsuperscript{b} & ---\textsuperscript{b} & ---\textsuperscript{b} & ---\textsuperscript{b} & ---\textsuperscript{b} \\
\addlinespace[2pt]
256 & HybridCertified & 52/64 & 81.25 & 244.11 & 7.5359 & $[+6.5207,\,+8.5512]$ \\
256 & FreshExact & 45/64 & 70.31 & 247.38 & 6.6135 & $[+5.4733,\,+7.7538]$ \\
256 & FreshSeq & 23/64 & 35.94 & 207.03 & 3.4676 & $[+2.2709,\,+4.6643]$ \\
256 & Fresh-U$_{167}$ & ---\textsuperscript{b} & ---\textsuperscript{b} & ---\textsuperscript{b} & ---\textsuperscript{b} & ---\textsuperscript{b} \\
\addlinespace[2pt]
300 & HybridCertified & 54/64 & 84.38 & 284.72 & 7.7888 & $[+6.8324,\,+8.7452]$ \\
300 & FreshExact & 47/64 & 73.44 & 290.34 & 6.9149 & $[+5.8061,\,+8.0238]$ \\
300 & FreshSeq & 24/64 & 37.50 & 234.91 & 3.6949 & $[+2.4545,\,+4.9353]$ \\
300 & Fresh-U$_{167}$ & ---\textsuperscript{b} & ---\textsuperscript{b} & ---\textsuperscript{b} & ---\textsuperscript{b} & ---\textsuperscript{b} \\
\addlinespace[2pt]
334 & HybridCertified & 55/64 & 85.94 & 317.47 & 7.9018 & $[+6.9765,\,+8.8271]$ \\
334 & FreshExact & 49/64 & 76.56 & 323.38 & 7.1533 & $[+6.0886,\,+8.2181]$ \\
334 & FreshSeq & 29/64 & 45.31 & 254.38 & 4.4247 & $[+3.1581,\,+5.6912]$ \\
334 & Fresh-U$_{167}$ & 50/64 & 78.12 & 323.31 & 7.2811 & $[+6.2405,\,+8.3218]$ \\
\addlinespace[2pt]
400 & HybridCertified & 57/64 & 89.06 & 379.69 & 8.1551 & $[+7.3013,\,+9.0088]$ \\
400 & FreshExact & 54/64 & 84.38 & 386.69 & 7.9372 & $[+6.9793,\,+8.8952]$ \\
400 & FreshSeq & 40/64 & 62.50 & 284.56 & 6.0150 & $[+4.7893,\,+7.2407]$ \\
400 & Fresh-U$_{167}$ & ---\textsuperscript{b} & ---\textsuperscript{b} & ---\textsuperscript{b} & ---\textsuperscript{b} & ---\textsuperscript{b} \\
\addlinespace[2pt]
512 & HybridCertified & 59/64 & 92.19 & 484.75 & 8.4743 & $[+7.6981,\,+9.2506]$ \\
512 & FreshExact & 56/64 & 87.50 & 493.03 & 8.1490 & $[+7.2594,\,+9.0385]$ \\
512 & FreshSeq & 42/64 & 65.62 & 324.69 & 6.2668 & $[+5.0688,\,+7.4648]$ \\
512 & Fresh-U$_{167}$ & ---\textsuperscript{b} & ---\textsuperscript{b} & ---\textsuperscript{b} & ---\textsuperscript{b} & ---\textsuperscript{b} \\
\bottomrule
\end{tabular}
\par\vspace{4pt}\noindent\begin{minipage}{\linewidth}\footnotesize \textsuperscript{a}Full-$S_{256}$ Hybrid requires 256 endpoints; no shorter selection prefix is substituted. \textsuperscript{b}Fresh-U$_{167}$ has only its certified 167-pair point at $B=334$. Dashes mean unavailable, not zero. FreshSeq is the historical sequential benchmark. Nested replays share 64 cohorts; intervals are descriptive and unadjusted, not independent power estimates.\end{minipage}
\end{table}
\begin{table}[!htbp]
\centering\footnotesize
\setlength{\tabcolsep}{9pt}
\renewcommand{\arraystretch}{1.04}
\caption{Synthetic method comparison at the representative setting $(q,d,p,h)=(0.1,0.02,0.7,64)$.}\label{tab:safety-methods}
\begin{tabular}{@{}lrrr@{}}
\toprule
Method & Positive adoption \% & Negative adoption \% & Gain pp \\
\midrule
ReuseFitted & 93.171 & 21.814 & 1.1735 \\
FreshMixed167 & 90.503 & 34.816 & 1.0581 \\
AlwaysUpdate & 100.000 & 100.000 & 0.8000 \\
NoUpdate & 0.000 & 0.000 & 0.0000 \\
\bottomrule
\end{tabular}
\par\vspace{4pt}\noindent\begin{minipage}{\linewidth}\footnotesize FreshMixed167 is utility-oriented, distinct from FreshUnconditional$_{167}$. ReuseFitted's $0.21814$ is harmful-state adoption, not a utility penalty. Hybrid was not evaluated here; its 5\% line is a theoretical bound under its assumptions, with no empirical utility point.\end{minipage}
\end{table}
\FloatBarrier
\end{document}